\documentclass{article} 
\usepackage{iclr2026_conference,times}

\usepackage{amsmath,amsfonts,bm}

\def\eqref#1{equation~\ref{#1}}

\def\1{\bm{1}}

\DeclareMathAlphabet{\mathsfit}{\encodingdefault}{\sfdefault}{m}{sl}
\SetMathAlphabet{\mathsfit}{bold}{\encodingdefault}{\sfdefault}{bx}{n}

\usepackage{hyperref}
\usepackage{url}

\usepackage{xspace}
\usepackage{graphicx}
\usepackage[table,dvipsnames]{xcolor}
\usepackage{wrapfig}
\usepackage{dsfont}     
\newcommand{\ind}{\mathds{1}}

\usepackage{listings}
\usepackage{amssymb}
\usepackage{soul}    
\DeclareRobustCommand{\hlc}[2][yellow!35]{%
  {\sethlcolor{#1}\hl{#2}}%
}

\lstdefinestyle{plaincode}{
  basicstyle=\ttfamily\small\color{black},
  backgroundcolor=\color{white},
  columns=fullflexible,
  keepspaces=true,
  showstringspaces=false,
  frame=single,
  framerule=0.4pt,
  rulecolor=\color{black!30},
  escapeinside={(*@}{@*)},
  breaklines=true,            
  breakatwhitespace=false,    
  breakindent=1em,            
  postbreak=\mbox{\textcolor{gray}{$\hookrightarrow$}\space} 
}

\definecolor{hlAmber}{HTML}{FFECB3}   
\definecolor{hlGreen}{HTML}{C8E6C9}   
\definecolor{hlBlue}{HTML}{BBDEFB}    
\definecolor{hlLavender}{HTML}{D1C4E9}
\definecolor{hlRed}{HTML}{FFCDD2}     

\newcommand{\emoji}[2][1em]{%
  \raisebox{-0.15\height}{\includegraphics[height=#1]{#2}}%
}

\newcommand{\spectrumemoji}{\emoji{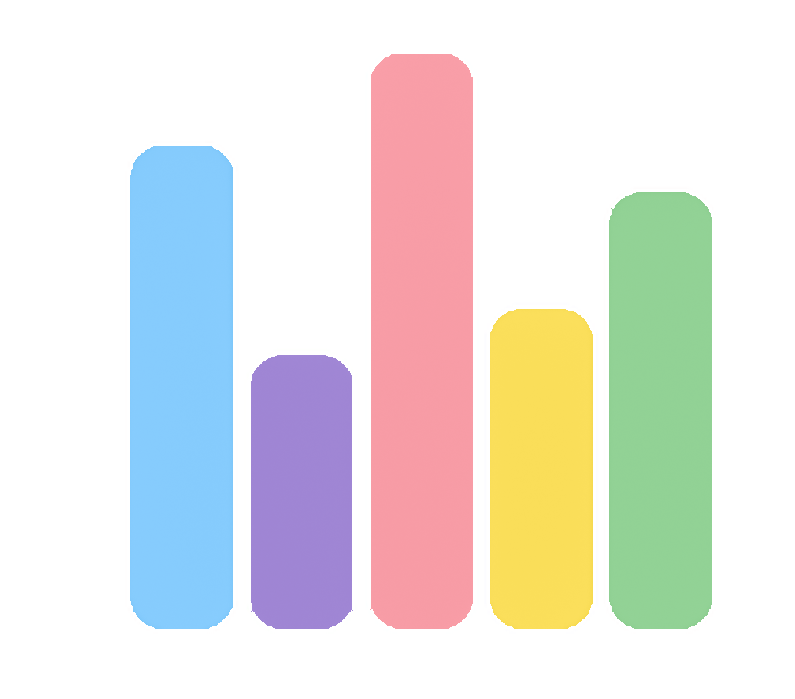}\xspace}

\DeclareRobustCommand{\spectrumsuiteemoji}{%
  \raisebox{-0.20\height}{%
    \hspace{-.1em}%
    \includegraphics[height=1.2em]{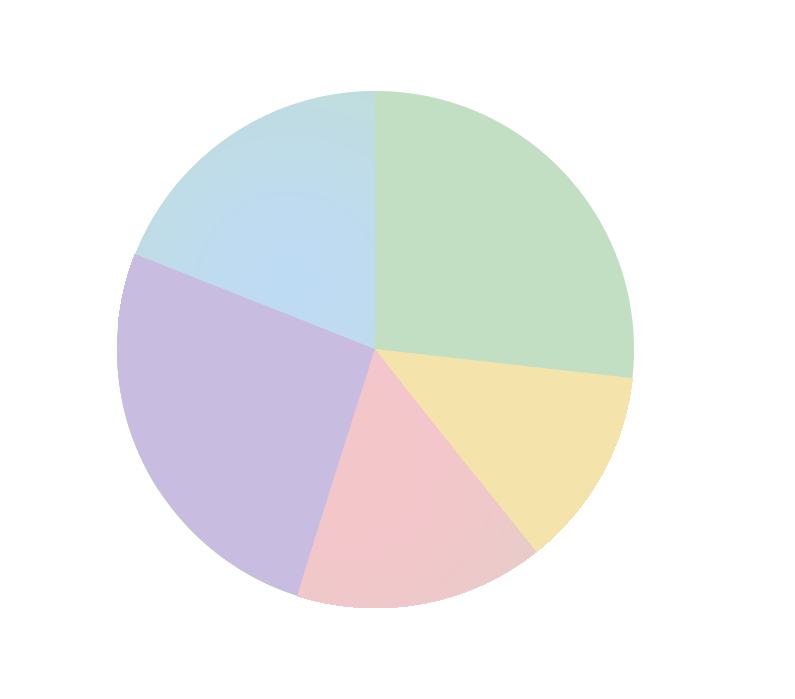}%
  }\xspace
}

\newcommand{\githubemoji}{%
  \raisebox{-0.20\height}{%
    \hspace{0.15em}
    \includegraphics[height=0.9em]{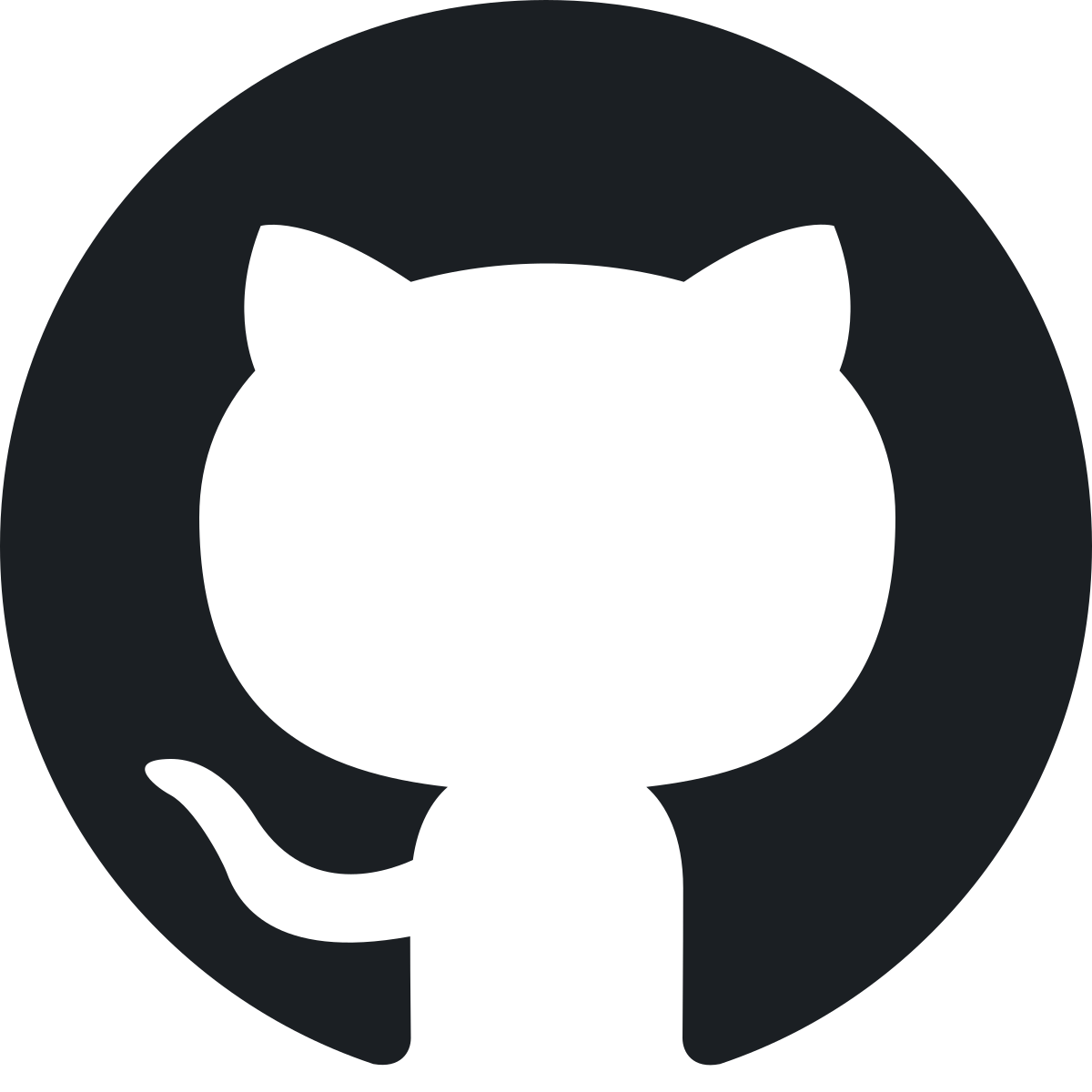}%
  }\xspace
}
\newcommand{\hfemoji}{%
  \raisebox{-0.23\height}{%
    \hspace{0.0em}
    \includegraphics[height=1.2em]{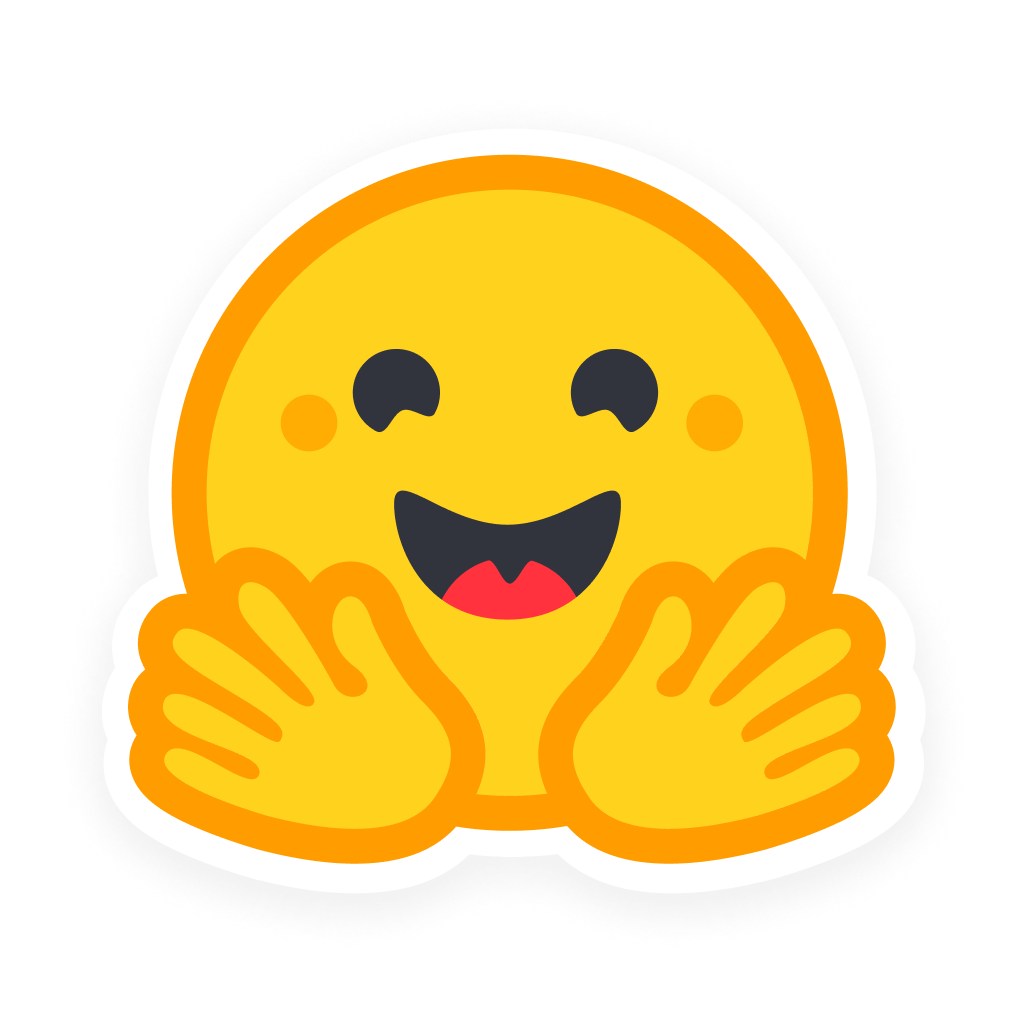}%
  }\xspace
}

\usepackage{algorithm}
\usepackage{algpseudocode}
\algrenewcommand\algorithmicrequire{\textbf{Inputs:}}
\algrenewcommand\algorithmicensure{\textbf{Output:}}
\algrenewcommand\algorithmiccomment[1]{\hfill\(\triangleright\) #1}

\usepackage{xparse}

\usepackage{adjustbox}

\DeclareRobustCommand{\method}{\textsc{Spectrum Tuning}\xspace}

\DeclareRobustCommand{\methodabb}{\textsc{ST}\xspace}
\DeclareRobustCommand{\modelshort}{\textsc{ST}\xspace}
\DeclareRobustCommand{\resource}{\textsc{Spectrum Suite}\xspace}

\usepackage{verbatim}
\usepackage{booktabs}

\title{\spectrumemoji Spectrum Tuning: \\ Post-Training for Distributional Coverage\\ and In-Context Steerability}

\iclrfinalcopy

\author{Taylor Sorensen$^1$, Benjamin Newman$^1$, Jared Moore$^2$, Chan Young Park$^3$, Jillian Fisher$^1$, \\
\textbf{Niloofar Mireshghallah$^{4}$, Liwei Jiang$^1$, Yejin Choi$^2$} \\
$^1$University of Washington, 
$^2$Stanford University,
$^3$Microsoft Research,
$^4$Carnegie Mellon University\\
Correspondence: \texttt{tsor13@cs.washington.edu, yejinc@stanford.edu} \\
\githubemoji Code and Dataset: \href{https://github.com/tsor13/spectrum}{\texttt{github.com/tsor13/spectrum}} \\
\hfemoji Models: \href{https://huggingface.co/collections/tsor13/spectrum-68dac670f618224845c0fb7d}{\texttt{huggingface.co/collections/tsor13/spectrum}}
}

\begin{document}

\maketitle

\begin{abstract}
\vspace{-10pt}
Language model post-training has enhanced instruction-following and performance on many downstream tasks, but also comes with an often-overlooked cost on tasks with many possible valid answers.
On many tasks such as creative writing, synthetic data generation, or steering to diverse preferences, models must cover an entire distribution of outputs, rather than a single correct answer.
We characterize three desiderata for conditional distributional modeling: in-context steerability, valid output space coverage, and distributional alignment, and document across three model families how current post-training can reduce these properties. In particular, we disambiguate between two kinds of in-context learning: ICL for eliciting existing underlying knowledge or capabilities, and \textit{in-context steerability}, where a model must use in-context information to override its priors and steer to a novel data generating distribution. To better evaluate and improve these desiderata, we introduce \resource, a large-scale resource compiled from $>$$40$ data sources and spanning $>$$90$ tasks requiring models to steer to and match diverse distributions ranging from varied human preferences to numerical distributions and more. We find that while current post-training techniques elicit underlying capabilities and knowledge, they hurt models' ability to flexibly steer in-context. To mitigate these issues, we propose \method, a post-training method using \resource to improve steerability and distributional coverage. We find that \method often improves over
pretrained and typical instruction-tuned models,
enhancing steerability, spanning more of the output space, and improving distributional alignment on held-out datasets.
\end{abstract}

\vspace{-5pt}
\section{Introduction}

Current post-training recipes \citep{rafailov2024directpreferenceoptimizationlanguage, tie2025surveyposttraininglargelanguage, wang2025reinforcementlearningreasoninglarge} have made language models (LLMs) easier to use via instruction-following \citep{ouyang2022training}, improved safety, and led to performance increases across many tasks, especially those with a single correct answer (e.g., mathematical reasoning, programming, chat preferences, etc.).
However, the effect of current post-training on tasks requiring steerability and distribution matching is less studied.
We show that current post-training can also negatively impact three related desiderata for conditional distributional modeling: in-context steerability, output coverage, and distributional alignment.

In this paper, we contribute: 1) an outline of these related desiderata, including the novel concept of \textit{in-context steerability}; 2) \resource, a dataset for evaluating and enhancing these desiderata; 3) a novel finding that while current post-training helps at many objective tasks, it can \textit{hurt} LLMs' in-context steerability; and 4) empirical evidence from our and related work that current post-training hurts output coverage and distributional alignment. To alleviate these weaknesses, we contribute 5) \method, a post-training technique utilizing \resource to improve these desiderata, and 6) show that our method enhances these properties compared to pretrained and current instruction-tuned models. To our knowledge, our method is the first to improve distributional alignment over pretrained models.

\section{Desiderata for Conditional Distributional Modeling}
\label{sec:desiderata}
\vspace{-5pt}

Before the age of post-training, in-context learning was necessary to reliably get pretrained language models to perform tasks such as  sentiment classification, translation, entailment, summarization, etc. \citep{brown2020languagemodelsfewshotlearners, dong2024surveyincontextlearning}. Let us call this use of in-context learning \textit{capability elicitation}, as its main purpose is to elicit some latent knowledge or capability of a language model \citep{min2022rethinkingroledemonstrationsmakes}. As post-training methods have increased LLMs' instruction-following capability, zero-shot instruct models have even surpassed their few-shot pretrained counterparts \citep{wei2022finetunedlanguagemodelszeroshot, sanh2022multitaskpromptedtrainingenables, ouyang2022training}, obviating the need for in-context capability elicitation.

\textbf{In-Context Steerability.}
In contrast to knowledge elicitation, many tasks require steering, or modifying output probabilities, based on novel information at inference time. For example, if a user wants an LLM to write an email in their style, it needs to either see examples of their writing or have an in-depth description of their style, and be able to effectively leverage this information to change its output distribution. This is distinct from pure capability/knowledge elicitation on unambiguous tasks, where the model can place a sharp prior on the ``correct" answer. Instead, the model must 1) maintain a prior over many possible generation functions and 2) maximally leverage in-context information in a well-calibrated way to form a posterior. Let us term this ability \textit{in-context steerability}. For example, this steerability is necessary for predicting a particular user's preferences or estimating an unknown numerical distribution from draws. In-context steerability can also be seen as implicit Bayesian reasoning \citep{qiu2025bayesianteachingenablesprobabilistic} or as a subset of in-context learning/instruction-following tasks where the model must utilize novel information in-context.

\textbf{Valid Output Coverage.} Many prompts entail multiple valid responses. For example, in creative story-writing, hypothesis proposal, and synthetic data generation, the number of possible valid outputs can be thousands or more. While in some cases it may be sufficient to produce one reasonable output, more value may lie in producing \textit{many} outputs so that a user can select the most interesting story, test all possible hypotheses, or otherwise span the entire task space. In the words of \citet{wilson2022bayesiandeeplearningprobabilistic},  ``we want the support of the model to be large so that we can represent any hypothesis we believe to be possible, even if it is unlikely."

\textbf{Distributional Alignment.} Sometimes, a user may not want a particular output, but rather a \textit{distribution} over outputs \citep{meister2024benchmarkingdistributionalalignmentlarge}. For example, \citet{sorensen2024roadmappluralisticalignment} propose \textit{distributional pluralism}  for modeling or representing a population by matching their opinion distribution. In addition, distributional alignment can simulate stochastic processes and estimate uncertainty. Distinct from valid output coverage, distributional alignment includes a target probability mass function.

\vspace{-5pt}
\section{Dataset and Method}
\label{sec:datasetandmethod}
\vspace{-5pt}
\subsection{\resource \texorpdfstring{\spectrumsuiteemoji}{Spectrum Suite}}
\vspace{-5pt}

\begin{wrapfigure}{r}{0.65\columnwidth}
\vspace{-40pt}
    \centering
    \includegraphics[width=0.7\textwidth]{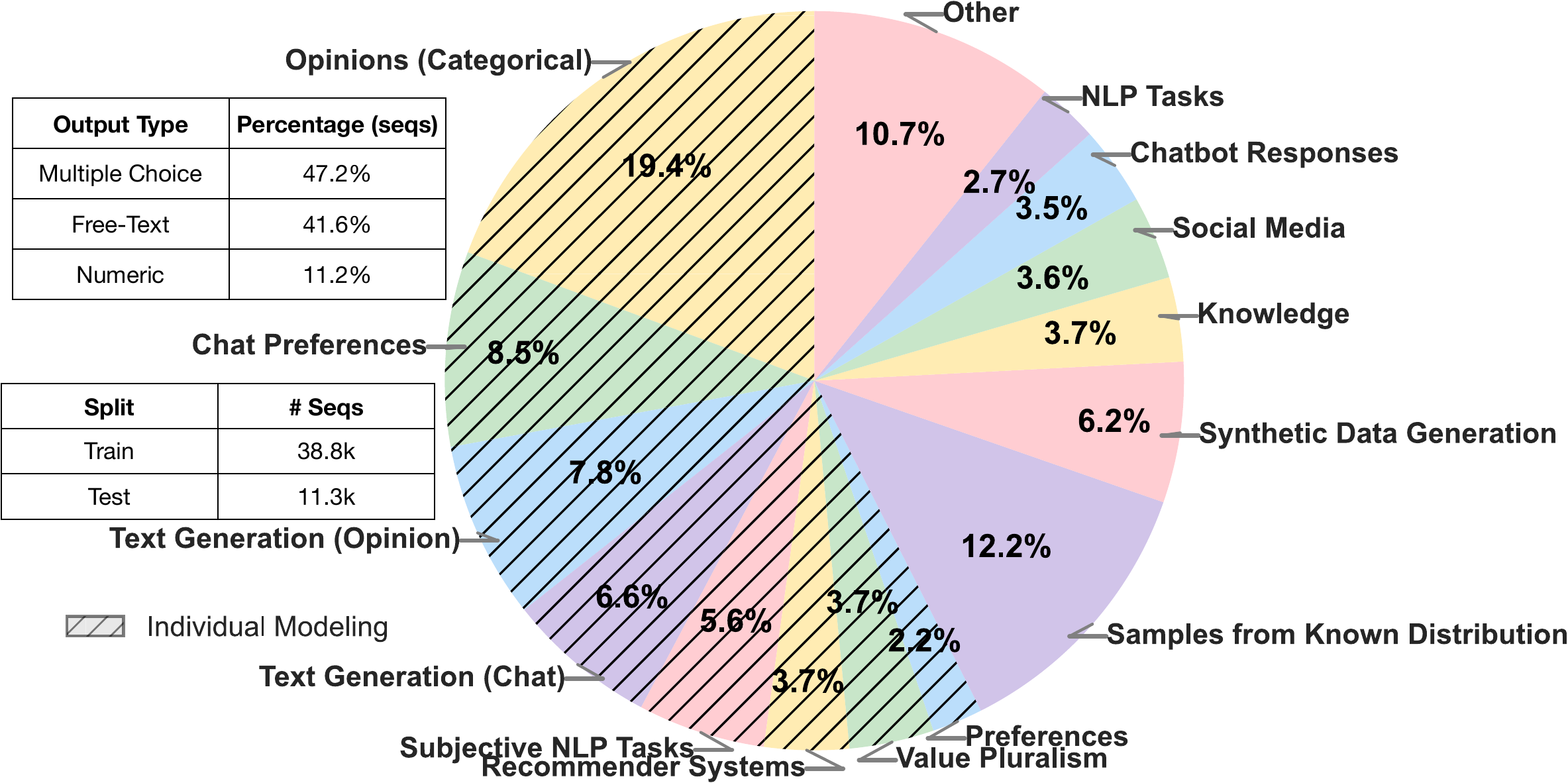}
    \vspace{-15pt}
    \caption{Task composition from \resource. Individual modeling tasks (data from the same person) are shaded.}
    \label{fig:spectrumsuitecomposition}
\vspace{-10pt}
\end{wrapfigure}

To measure and elicit these properties, we compile datasets that either 1) exhibit natural person-to-person variation (e.g.,  opinion modeling, chat preferences, subjective NLP tasks); 2) involve a large collection of interchangeable texts drawn from a particular distribution (e.g., synthetic data, poems in a particular format); 3) are i.i.d. draws from a random distribution (e.g., draws from a normal distribution); or 4) involve reasoning under uncertainty. We draw from $>\!\!40$ data sources in order to make $>\!\!90$ separate tasks. We unify each task data into a common format including: \texttt{description}: a natural language description of the task, \texttt{input}: any given information for a particular data instance, and \texttt{output}: the output sequence which we would like the model to learn. Some tasks require an \texttt{input} associated with each \texttt{output} (e.g., the question asked in a survey is needed to contextualize the answer), while other tasks consist of only \texttt{outputs}.
In particular, we focus on individual modeling data on tasks with human variation. We do so for a couple of reasons: many use cases involve steering to a particular individual at inference time; and these data sources are very rich as modeling each person involves a different data generation task. These data comprise 50.1k distinct sequences consisting of a \texttt{description} followed by multiple \texttt{inputs/outputs}. For summary statistics and task breakdown of \resource, see Figure \ref{fig:spectrumsuitecomposition}. For information on all data sources, see App. \ref{app:datasources}.
We split \resource into non-overlapping train and test tasks, with held-out test tasks drawn from separate data sources to ensure generality.

\vspace{-5pt}
\subsection{\method \spectrumemoji}
\vspace{-5pt}

Let $T_i \in \mathcal{T}$ be some task (or, data generation process) that we want to model. Let $Y_i$ be the \texttt{output} space to approximate, $X_i$ be any known covariates (optional \texttt{input}), and $Z_i$ be a latent context for the task (optional \texttt{description}). $T_i: X_i, Z_i \to P(Y_i)$ maps to a probability distribution over potential outputs. This is the classic meta-learning formulation \citep{hospedales2020metalearningneuralnetworkssurvey}, except that the target is a distribution over $P(Y^i)$ instead of a single $y_i$. Because the task $T_i$ may be difficult to directly observe, we may instead wish to learn it from data (e.g., Monte Carlo samples).

\begin{wrapfigure}{r}{0.6\textwidth}
\vspace{-\baselineskip}
\vspace{-\baselineskip}
\begin{minipage}[t]{\linewidth}
\small
\begin{algorithm}[H]
\caption{\method}
\label{alg:spectrum_tuning}
\begin{algorithmic}[1]
\Require Pretrained LM $m_\theta$; train task distribution $\mathcal{T}^{\text{train}}$; tokenizer $t(\cdot)$ with template for description/input/output; terminal token $\langle\text{END}\rangle$; loss ignore index $i_\textrm{drop}$; description drop probability $p_{\text{drop}}$ (default $0.2$).
\Ensure Finetuned parameters $\theta'$
\For{each task $T \sim \mathcal{T}^{\text{train}}$}
    \Comment{Sample a task}
    \State Sample description $z$ and support set $S=\{(x_j,y_j)\}_{j=1}^{n}$.
    \State Randomly permute indices $\pi$ of $\{1,\dots,n\}$.
    \If{$\mathrm{Uniform}(0,1)>p_{\text{drop}}$ } \Comment{Keep description}
    \State $\textit{seq} \gets t(z) \Vert t(x_{\pi[0]}) \Vert t(y_{\pi[0]}) \Vert \langle\text{END}\rangle$
    \State $\textit{labels} \gets i_\textrm{drop}(t(z) \Vert t(x_{\pi[0]})) \Vert t(y_{\pi[0]})\Vert \langle\text{END}\rangle$
    \Comment{Loss on first output, no loss on description/output}
\Else \Comment{Description dropout w/ prob. $p_\textrm{drop}$}
    \State $\textit{seq} \gets t(x_{\pi[0]}) \Vert t(y_{\pi[0]})\Vert \langle\text{END}\rangle$
    \State $\textit{labels} \gets  i_\textrm{drop}(t(x_{\pi[0]}) \Vert t(y_{\pi[0]}))\Vert \langle\text{END}\rangle)$
    \Comment{No loss on first output if description is missing}
\EndIf
    \For{$j$ in $\pi[1:]$} \Comment{Add remaining}
        \State $\textit{seq} \gets \textit{seq} \,\Vert\, t(x_j) \,\Vert\, t(y_j) \,\Vert\, \langle\text{END}\rangle$
        \State $\textit{labels} \gets \textit{labels} \,\Vert\, i_\textrm{drop}(t(x_j)) \,\Vert\, t(y_j) \,\Vert\, \langle\text{END}\rangle$ \Comment{Loss on output, no loss on input}
    \EndFor
    \State $L \gets \mathrm{CrossEntropy}\!\left(m_\theta(\textit{seq}),\ \textit{labels}\right)$
    \State $\theta \gets \theta - \eta \, \nabla_\theta L$
\EndFor \Comment{Train for one epoch}
\State \textbf{return} $\theta' \gets \theta$
\end{algorithmic}
\end{algorithm}
\end{minipage}
\vspace{-10pt}
\end{wrapfigure}

The method (Algorithm \ref{alg:spectrum_tuning}) is simple: for a collection of tasks, tokenize the task context/description $z_i$ followed by (randomly ordered) in-context examples $x_{ij}, y_{ij}$, then perform supervised finetuning calculating cross-entropy loss \textit{only} on the output tokens. Because cross-entropy loss on Monte Carlo samples from a distribution encourages a well-calibrated estimate of the underlying distribution in the underfit regime ($\leq$$1$ epoch, \citealt{ji2021earlystoppedneuralnetworksconsistent}) the optimal model solution is to approximate the true underlying distribution $P(Y_i)$.

To build intuition on how \method supports the desiderata, let us consider a few cases. When a model predicts the first output, it must rely only on the description, and shift its probabilities to outputs fitting the description. Because there can be many possible valid outputs and the model has no information about which output to expect, it is incentivized to \textit{cover} the entire possible distribution of outputs. Additionally, if the distribution over valid outputs is skewed in some predictable way (e.g., an opinion distribution), the model is further incentivized to \textit{match} said distribution. On subsequent outputs, the model must \textit{steer} its output distribution, utilizing in-context examples to update its beliefs in a well-calibrated way. Additionally, \resource tasks allow the model to utilize assumptions which don't apply to the pretraining distribution: predictions are invariant to output ordering,\footnote{i.e. ``exchangeable" in Bayesian analysis \citep{bayesianstatisticalanalysis}, as the posterior is invariant to sample order.} the underlying generative process remains constant, and the model can concentrate all probability mass on valid outputs instead of on other possible text continuations.
In many ways, \method is similar to supervised fine-tuning on instruction data \citep{zhang2025instructiontuninglargelanguage}, as loss is calculated only on an output. However, it differs in several important respects: 1) many identically-distributed outputs are included in-context, encouraging meta-learning; 2) training on data that is distributional in nature; 3) sole focus on distribution fitting instead of chat-style data; and 4) inputs are optional, unlike chat user messages which are always required.

\vspace{-5pt}
\subsection{Implementation Details}
We train models from three families using \method on the train tasks from \resource: \texttt{gemma-3-12b} \citep{gemmateam2025gemma3technicalreport}, \texttt{Llama-3.1-8B} \citep{grattafiori2024llama}, and \texttt{Qwen3-14B} \citep{yang2025qwen3technicalreport}. We refer to pretrained or base models as PT models and instruction-tuned post-trained models as IT models, and utilize each family's provided PT/IT model as baselines.
To match our meta-learning task setup (as opposed to chat), we adapt each model's chat template to use the \texttt{description/input/output} roles instead of \texttt{system/user/assistant} (cf. Fig. \ref{fig:spectrumsuite}).
For \method, we initialize with the PT model weights, except for the uninitialized (un/)embedding weights for the two or three special format tokens which we initialize from the IT model.
See App. \ref{app:trainingdetails} for more training details.


\vspace{-5pt}
\section{In-Context Steerability}
\label{sec:incontextsteerability}

\begin{wrapfigure}{r}{0.54\columnwidth}
\vspace{-40pt}
    \includegraphics[width=0.54\textwidth]{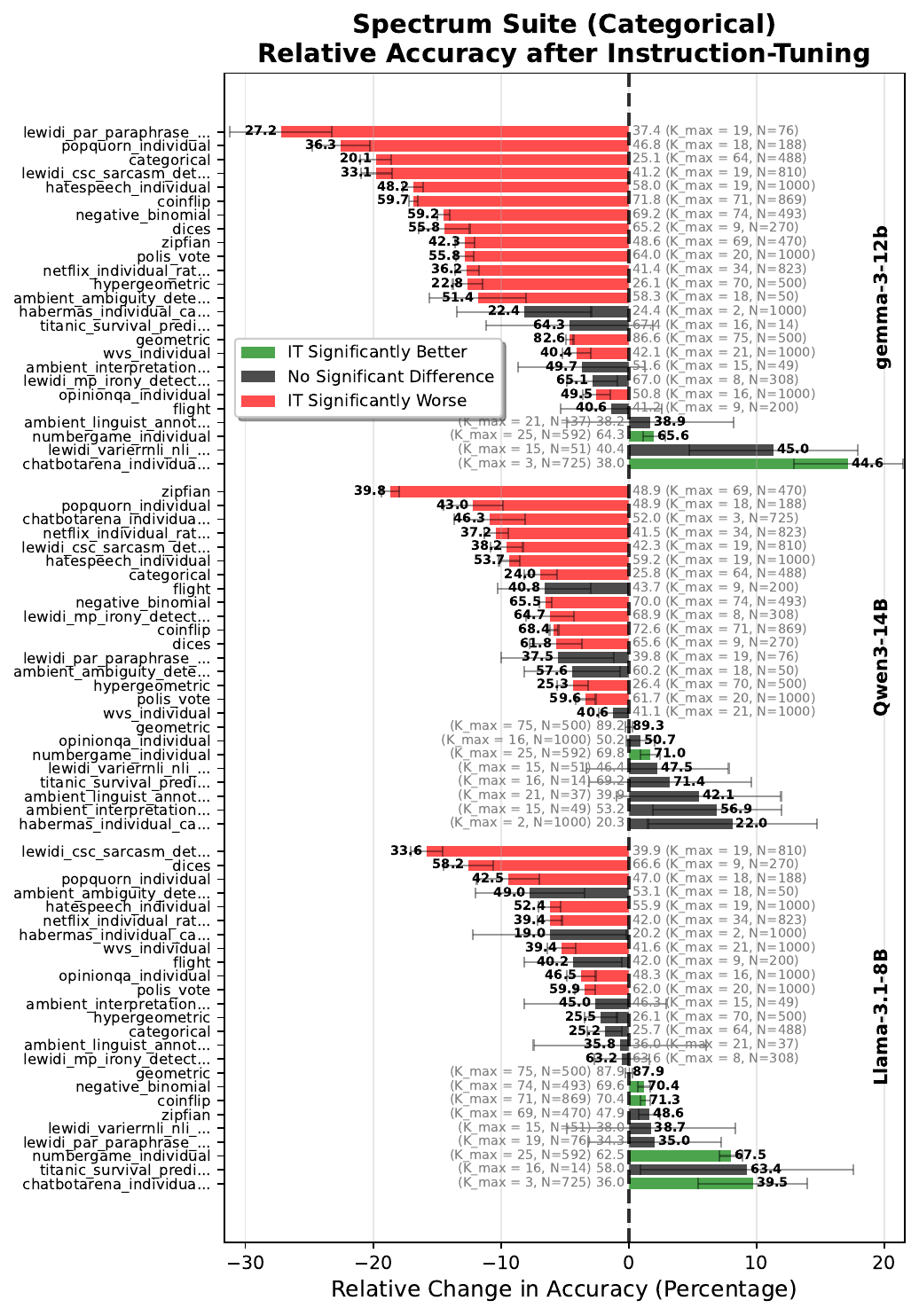}
    \vspace{-15pt}
    \caption{
    Change in accuracy on \resource from the pretrained to instruction-tuned model. Current instruction-tuning hurts in-context steerability.
    }
    \label{fig:mc_acc}
\vspace{-10pt}
\end{wrapfigure}

\vspace{-10pt}
We use \resource to evaluate models' ability to steer to varied generation tasks. We measure $k$-shot learning by 1) fitting the description and examples from a single task into context, 2) measuring the loss (negative log-likelihood) of each \texttt{output} conditioned on the prior examples under the model $m_\theta$: $\textrm{NLL}_{m_\theta}(y_k) = -\log p_{m_\theta} (y_k | z, y_0, \cdots, y_{k-1})$. Additionally, for multiple-choice datasets, we calculate the accuracy of the output: whether the greedily-decoded model response results in the correct answer. For each task, we choose $K_\textrm{max}$ such that it maximizes the total number of examples that we can evaluate when we restrict to only sequences with at least $K_\textrm{max}$ examples that fit into a 1024-token context-window. In order to maximize sample efficiency and evaluate a model's ability to steer for varied $k$, we report the average loss and accuracy for $k$-shot learning for $k \in \{1,\cdots,K_\textrm{max}\}$.

First, we ask: how does current instruction-tuning impact in-context steerability? For the PT models, we use the same prompt template for all models, with \texttt{Description:}/\texttt{Input:}/\texttt{Output:} delineated by newlines. To ensure we are leveraging maximum performance from the IT models, we test each IT model's performance on both the PT prompt and two chat-style ICL prompts, and report results for the best performing prompt template (see App. \ref{app:prompttemplates}). We evaluate in-context steerability on all of \resource for the PT/IT models. We include the entire suite of results in Appendix \ref{app:supplementaryfigures}, and highlight the principal results below.

\textbf{Current instruction-tuning hurts in-context steerability.} First, let's examine the change in accuracy for the IT models. We report accuracy for all categorical data (multiple-choice + small support numeric distributions) in Figure \ref{fig:mc_acc}. Out of 76 model family/task comparisons, instruction-tuning \textit{significantly decreases} accuracy in 35 cases, doesn't significantly affect accuracy in 33 cases, and significantly increases accuracy in only 7 cases.  Additionally, two of the seven comparisons where instruction-tuning helped were on predicting an individual's chatbot preferences---which is adjacent to precisely what instruct models are optimized for (chat). The performance drop is even more stark on loss: for Gemma and Qwen, loss is higher on 50/50 comparisons, while on Llama loss is worse in 11 cases, the same in 11 cases, and better in 3 cases. Loss results are similar on the free-text \resource datasets: out of 144 comparisons, IT loss is worse than PT loss in 117 cases, tied in 25 cases, and better only in 2 cases.

\begin{wrapfigure}{rt}{0.45\columnwidth}
\vspace{-15pt}
    \includegraphics[width=0.45\textwidth]{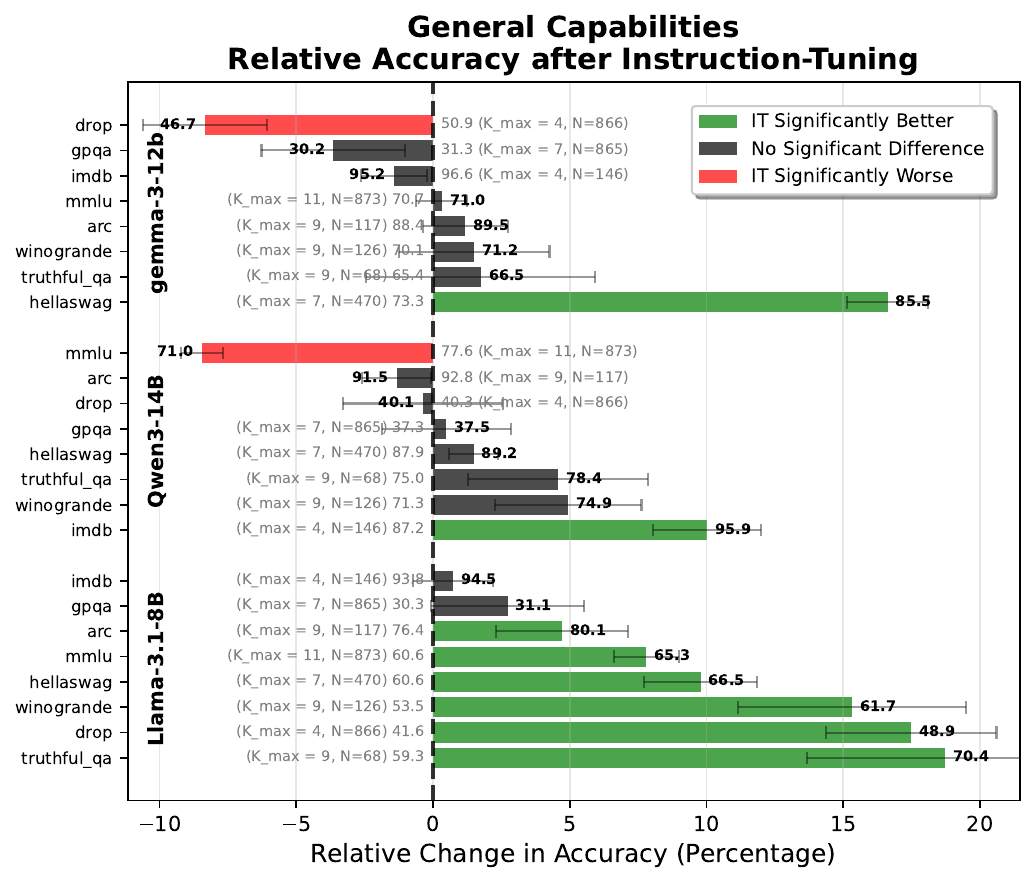}
    \vspace{-15pt}
    \caption{
    Current instruction-tuning generally helps on capability benchmarks.
    }
    \label{fig:mc_acc_gen}
\vspace{-15pt}
\end{wrapfigure}

\textbf{ICL for general capability elicitation is not degraded by instruction-tuning.} To disambiguate in-context steerability from general capability elicitation, we also run the exact same experiment with eight general capability task datasets (Fig. \ref{fig:mc_acc_gen}). In contrast with the \resource datasets, accuracy \textit{increases} in 8 of 24 cases, is the same in 13 cases, and decreases in 2 cases.

All in all, we believe that this characterizes a difference in behavior for IT models---while they maintain the ability to utilize in-context demonstrations for general capability elicitation, they seem to struggle to adapt at tasks that require heavy in-context steerability. Limited prior work has suggested that instruction-tuned models sometimes perform better without in-context examples \citep{asai-etal-2024-buffet,lambert2025tulu}; however, to our knowledge, ours is the first work to empirically characterize this in-context learning performance degradation for in-context steerability tasks.

What explains this difference? While we leave an in-depth exploration of this phenomenon to future work, we hypothesize that it could be due to some combination of 1) instruction-tuning inducing very strong priors that are difficult to override even with in-context demonstrations,  2) over-optimization on tasks with a single ground truth, or 3) overfitting to particular benchmarks.

\vspace{-5pt}
\subsection{\method and In-Context Steerability on Held-out Tasks} We have characterized that current instruction-tuned models struggle at in-context steerability, but how does our method compare? We evaluate Spectrum-Tuned (\methodabb) models on \resource test tasks and compare them to their PT and IT counterparts (Table \ref{tab:methodsteerability}). Note that the test task data sources have no overlap with the train split, requiring generalization.

\begin{table}[h]
\centering
\tiny
\setlength{\tabcolsep}{6pt}
\renewcommand{\arraystretch}{1.15}
\begin{tabular}{ll*{9}{c}}
& & \multicolumn{3}{c}{\textbf{gemma-3-12b}} & \multicolumn{3}{c}{\textbf{Qwen3-14B}} & \multicolumn{3}{c}{\textbf{Llama-3.1-8B}} \\
\cmidrule(lr){3-5} \cmidrule(lr){6-8} \cmidrule(lr){9-11}
\textbf{Multiple-Choice Datasets} & \textbf{Metric} & \textbf{\methodabb (ours)} & \textbf{PT} & \textbf{IT} & \textbf{\methodabb} & \textbf{PT} & \textbf{IT} & \textbf{\methodabb} & \textbf{PT} & \textbf{IT} \\
\midrule
\rowcolor{gray!20} \textbf{habermas\_individual\_categorical} ($K_\textrm{max}$=2, N=1000) & \textbf{Loss} & \textbf{2.47} & \textbf{2.50} & 10.5 & \textbf{1.97} & 2.62 & 9.10 & \textbf{1.99} & 2.58 & 2.74 \\
 & \textbf{Acc} & \textbf{23.8} & \textbf{24.4} & \textbf{22.4} & \textbf{23.5} & 20.3 & \textbf{22.0} & \textbf{20.8} & \textbf{20.2} & \textbf{19.0} \\
\rowcolor{gray!20} \textbf{wvs\_individual} ($K_\textrm{max}$=21, N=1000) & \textbf{Loss} & \textbf{1.36} & 1.50 & 4.10 & \textbf{1.48} & 1.74 & 4.35 & \textbf{1.42} & 1.57 & 1.76 \\
 & \textbf{Acc} & \textbf{42.6} & \textbf{42.1} & 40.4 & \textbf{44.3} & 41.1 & 40.6 & \textbf{41.7} & \textbf{41.6} & 39.4 \\
\rowcolor{gray!20} \textbf{numbergame\_individual} ($K_\textrm{max}$=25, N=592) & \textbf{Loss} & \textbf{.639} & .705 & 1.80 & \textbf{.621} & .697 & 1.28 & \textbf{.618} & .864 & .770 \\
 & \textbf{Acc} & \textbf{70.2} & 64.3 & 65.6 & \textbf{70.6} & 69.8 & \textbf{71.0} & \textbf{69.1} & 62.5 & 67.5 \\
\rowcolor{gray!20} \textbf{chatbotarena\_individual\_prefs} ($K_\textrm{max}$=3, N=725) & \textbf{Loss} & \textbf{1.43} & 1.62 & 4.94 & \textbf{1.34} & 1.47 & 4.39 & \textbf{1.39} & 1.76 & 1.77 \\
 & \textbf{Acc} & 38.6 & 38.0 & \textbf{44.6} & \textbf{51.4} & \textbf{52.0} & 46.3 & \textbf{38.9} & 36.0 & \textbf{39.5} \\
\rowcolor{gray!20} \textbf{flight} ($K_\textrm{max}$=9, N=200) & \textbf{Loss} & \textbf{1.09} & 1.32 & 4.06 & \textbf{1.08} & 1.29 & 2.92 & \textbf{1.12} & 1.45 & 1.41 \\
 & \textbf{Acc} & \textbf{39.8} & \textbf{41.2} & \textbf{40.6} & \textbf{43.7} & \textbf{43.7} & \textbf{40.8} & 33.4 & \textbf{42.0} & \textbf{40.2} \\
\midrule
\textbf{Free-Text Datasets} & \textbf{Metric} & \textbf{\methodabb (ours)} & \textbf{PT} & \textbf{IT} & \textbf{\methodabb} & \textbf{PT} & \textbf{IT} & \textbf{\methodabb} & \textbf{PT} & \textbf{IT} \\
\midrule
 \textbf{novacomet\_hypothesis} ($K_\textrm{max}$=11, N=155) & \textbf{Loss} & \textbf{104} & \textbf{104} & 135 & \textbf{106} & \textbf{106} & 129 & \textbf{107} & \textbf{106} & 112 \\
 \textbf{novacomet\_premise} ($K_\textrm{max}$=55, N=51) & \textbf{Loss} & \textbf{27.7} & \textbf{28.0} & 35.5 & \textbf{28.1} & \textbf{27.5} & 38.0 & \textbf{27.8} & \textbf{27.7} & 28.6 \\
 \textbf{habermas\_question} ($K_\textrm{max}$=29, N=30) & \textbf{Loss} & \textbf{23.8} & \textbf{23.1} & 41.4 & \textbf{23.8} & \textbf{24.0} & 31.8 & \textbf{23.8} & \textbf{23.8} & 24.8 \\
 \textbf{habermas\_opinions} ($K_\textrm{max}$=2, N=186) & \textbf{Loss} & \textbf{930} & \textbf{928} & 1070 & \textbf{948} & \textbf{949} & 1070 & \textbf{943} & \textbf{944} & \textbf{991} \\
 \textbf{habermas\_individual} ($K_\textrm{max}$=2, N=1000) & \textbf{Loss} & \textbf{164} & \textbf{164} & 203 & \textbf{168} & \textbf{168} & 210 & \textbf{166} & \textbf{167} & 176 \\
 \textbf{numbergame\_perc} ($K_\textrm{max}$=24, N=182) & \textbf{Loss} & \textbf{4.23} & \textbf{4.22} & 6.68 & \textbf{4.22} & \textbf{4.24} & 5.61 & \textbf{4.24} & 4.43 & 4.41 \\
 \textbf{globaloqa} ($K_\textrm{max}$=8, N=231) & \textbf{Loss} & \textbf{14.0} & \textbf{14.4} & 21.5 & \textbf{14.0} & \textbf{14.4} & 20.9 & \textbf{14.2} & 14.7 & 15.6 \\
 \textbf{chatbotarena\_prompts} ($K_\textrm{max}$=3, N=988) & \textbf{Loss} & \textbf{70.2} & \textbf{69.4} & 117 & \textbf{69.1} & \textbf{68.2} & 97.8 & \textbf{72.0} & \textbf{72.0} & \textbf{77.6} \\
 \textbf{chatbotarena\_assistant} ($K_\textrm{max}$=5, N=716) & \textbf{Loss} & \textbf{127} & \textbf{125} & 259 & \textbf{124} & \textbf{124} & 169 & \textbf{134} & \textbf{133} & 149 \\
 \textbf{chemistry\_esol} ($K_\textrm{max}$=8, N=59) & \textbf{Loss} & 8.94 & \textbf{8.37} & 12.9 & \textbf{8.07} & 8.47 & 11.8 & \textbf{8.28} & \textbf{8.51} & \textbf{8.55} \\
 \textbf{chemistry\_oxidative} ($K_\textrm{max}$=9, N=101) & \textbf{Loss} & \textbf{7.57} & \textbf{7.58} & 11.6 & \textbf{7.64} & 7.84 & 10.2 & \textbf{7.64} & \textbf{7.72} & 7.84 \\
\midrule
\end{tabular}
\caption{
In-context steerability on held-out \resource-Test. \method generally matches or improves upon the pretrained model performance.
Best values (and ties, failing to find a significant difference at $\alpha=.05$) are bolded.
}
\label{tab:methodsteerability}
\end{table}

\textbf{\method usually matches, and sometimes improves upon, PT steerability.} Out of 15 multiple-choice (MC) loss comparisons, \modelshort ties with PT models in one case and achieves lower loss compared to PT models in 14 cases. On MC accuracy, ST matches/improves/worsens on 10/3/2 comparisons. On the free-text datasets, \modelshort matches PT in 28 cases, is worse in 1 case and is better in 4 cases. In most cases, \method matches (but does not beat) the very strong baseline of a pretrained model at in-context steerability, but does improve performance more often than it hurts performance.

\textbf{Models trained with \method most often have the best calibration.} We report calibration in Table \ref{tab:calibration}. In 9/15 cases, the \modelshort models have the best calibration. Additionally, the Gemma and Qwen IT models have worse calibration in 10/10 cases than their pretrained counterparts, showing another side effect of heavy instruction-tuning (cf. \citealt{tian2023justaskcalibrationstrategies, openai2024gpt4technicalreport}).

\begin{table}[t]
\centering
\tiny
\setlength{\tabcolsep}{6pt}
\renewcommand{\arraystretch}{1.15}
\begin{tabular}{l*{9}{c}}
\toprule
\textbf{Expected Calibration Error (ECE, $\downarrow$)} & \multicolumn{3}{c}{\textbf{gemma-3-12b}} & \multicolumn{3}{c}{\textbf{Qwen3-14B}} & \multicolumn{3}{c}{\textbf{Llama-3.1-8B}} \\
\cmidrule(lr){2-4} \cmidrule(lr){5-7} \cmidrule(lr){8-10}
\textbf{Multiple-Choice Dataset} & \textbf{ST (ours)} & \textbf{PT} & \textbf{IT} & \textbf{ST (ours)} & \textbf{PT} & \textbf{IT} & \textbf{ST (ours)} & \textbf{PT} & \textbf{IT} \\
\midrule
\textbf{habermas\_individual\_categorical} & 0.116 & \textbf{0.069} & 0.239 & \textbf{0.032} & 0.05 & 0.198 & \textbf{0.037} & 0.084 & 0.055 \\
\textbf{wvs\_individual} & \textbf{0.006} & 0.015 & 0.223 & \textbf{0.017} & 0.02 & 0.191 & \textbf{0.005} & 0.012 & 0.024 \\
\textbf{numbergame\_individual} & \textbf{0.015} & 0.029 & 0.163 & 0.027 & \textbf{0.026} & 0.108 & 0.028 & 0.024 & \textbf{0.017} \\
\textbf{chatbotarena\_individual\_prefs} & \textbf{0.020} & 0.041 & 0.194 & 0.048 & \textbf{0.046} & 0.189 & \textbf{0.046} & 0.075 & 0.049 \\
\textbf{flight} & \textbf{0.011} & 0.040 & 0.271 & 0.038 & \textbf{0.035} & 0.228 & 0.046 & 0.070 & \textbf{0.038} \\
\bottomrule
\end{tabular}
\caption{
Calibration on \resource-Test, binning label token probabilities every decile for expected calibration error ($\text{ECE} = \sum_{b=1}^{B} \frac{n_b}{N} |\text{acc}(b) - \text{conf}(b)|$, where $B=10$ bins, $n_b$ is the number of samples in bin $b$, $\text{acc}(b)$ is the accuracy in bin $b$, and $\text{conf}(b)$ is the average confidence in bin $b$). \method (\methodabb) usually results in the best calibration (9/15 cases).
}
\vspace{-5pt}
\label{tab:calibration}
\end{table}

\vspace{-5pt}
\section{Spanning the output space (or; Diversity vs. Validity)}
\label{sec:spanningoutputspace}
\vspace{-3pt}

\begin{figure}[t]
    \vspace{-10pt}
    \centering
    \includegraphics[width=\textwidth]{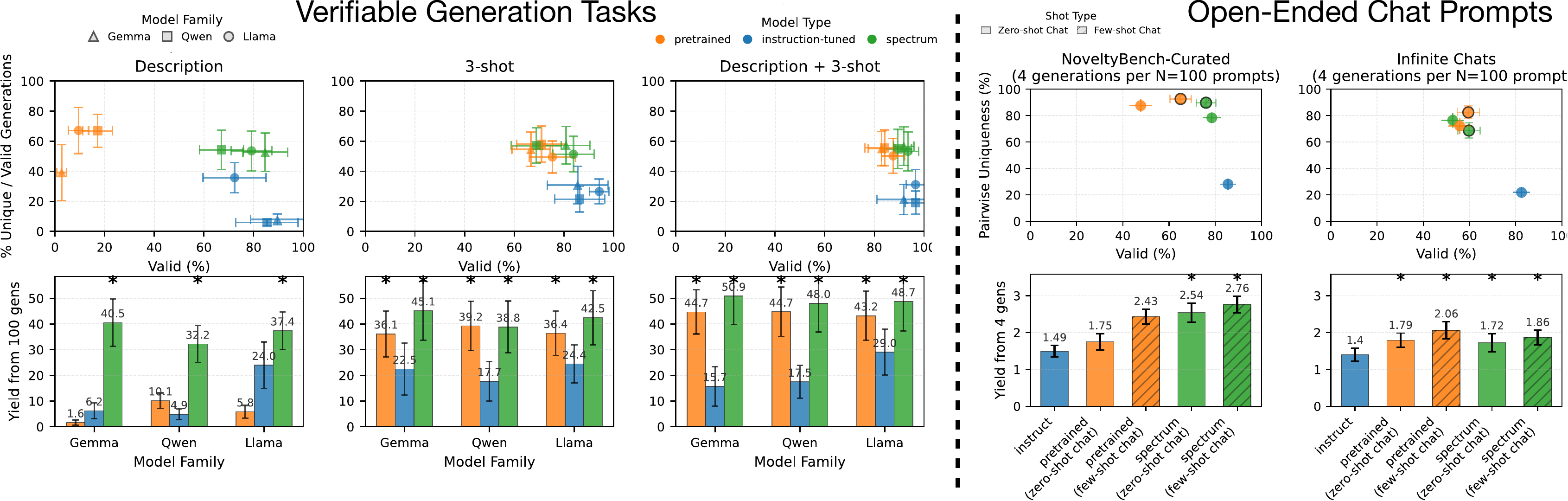}
    \vspace{-15pt}
    \caption{Diversity vs. Validity. Left: Results on 22 verifiable tasks across 100 generations. Right: Human-annotated validity results on two sets of 100 open-ended prompt sets (Gemma).
    \method generally offers a Pareto improvement on diversity-validity over PT/IT models. In particular, \method increases the yield (\# of unique usable generations) in the zero-shot case and on NoveltyBench-Curated.
    Error bars are 95\% confidence intervals over the SEM, and asterisks ($*$) show the best in family performance (within 95\% confidence).
    }
    \vspace{-10pt}
    \label{fig:diverse_verify_bar}
\end{figure}

To measure how each model trades off validity and diversity, we create 22 generation tasks for which there can be many valid values and we can programmatically verify correctness ($\ind_{\textrm{correct}}$).
Given a prompt, we generate 100 completions $o_1,\cdots,o_{100}$ ($\textrm{temperature}=1$ here and throughout) from each model, and report the following statistics: the percentage of outputs which are valid ($\sum_{i=1}^{100} \mathds{1}_{\textrm{correct}}(o_i)$), the percentage of valid generations that are unique
($\frac{|\textrm{dedup}(\{o_i : \mathds{1}_{\textrm{correct}}(o_i) = 1\})|}{\sum_{i=1}^{100} \mathds{1}_{\textrm{correct}}(o_i)}$),
and the number of distinct valid generations (or, \textit{yield}:
${|\textrm{dedup}(\{o_i : \mathds{1}_{\textrm{correct}}(o_i) = 1\})|}$).
We perform deduplication with exact string matching.
Yield is a particularly important metric for settings such as synthetic data generation, ideation, or creative writing where you want to cover a space as much as possible within some requirements. Additionally, we evaluate each model under three settings: zero-shot with a task description, three-shot with no task description, and three-shot with a task description (also see App. \ref{app:diversityvalidity}). Results can be found in Fig \ref{fig:diverse_verify_bar}. Tasks are the same across models.

\textbf{Instruction-tuned models have high validity but low diversity.} IT models produce valid outputs $>70\%$ of the time, even in the zero-shot setting. However, this comes at the price of diversity, resulting in fewer than $30$ unique valid generations in few-shot settings. Yield is even lower in the zero-shot setting---Qwen and Gemma average yields of only 5--6, while Llama averages only 24.

\textbf{Pretrained models are more diverse, but rely on few-shot examples for validity.} Pretrained models do not suffer from the same mode collapse, and consistently have higher diversity ($>40\%$ of valid generations are unique). However, this comes at a trade-off with validity, where their generations are universally less valid than the IT models'. The pretrained models also rely heavily on the few-shot examples to elicit valid generations, achieving a validity of $<20\%$ in the zero-shot case. However, in the few-shot cases, they have a significantly higher yield than the instruction-tuned models due to their higher coverage of the space.

\textbf{\method offers a Pareto improvement on diversity and validity, matching or exceeding pretraining yield.} In eight of nine model/setting comparisons, \method offers either a Pareto or strict improvement over the PT/IT models on validity/diversity. In all eight settings with a Pareto improvement, this also leads to a higher yield---i.e., \textbf{for a fixed generation budget, \method generates the most unique valid generations.}

\textbf{\method achieves much higher yield in the zero-shot setting.} Focusing in on the zero-shot setting, \method particularly shines. The IT models are able to follow the description and produce a valid output, but have very low diversity ($\sim\!\!30\%$ for Llama, $\sim\!\!5\%$ for Qwen and Gemma). Meanwhile, the pretrained models are unable to consistently generate valid outputs ($<20\%$ validity). \methodabb models, however, are able to follow the instructions and produce valid outputs $>60\%$ of the time while maintaining ~$50\%$ diversity. This leads to much higher yields compared to PT and IT models (Gemma: 40.5 vs. 6.2; Qwen: 32.2 vs. 10.1, Llama: 37.4 vs. 24.0).

\textbf{\method's gains hold across temperature values.}
One way to trade-off validity for diversity for a given model is sweeping temperature.
To ensure that our results hold across temperatures, we ran the same experiment with $T=[10, 5, 2, 1.5, 1, .9, .7, .5]$. We found that \method A) still expanded the Pareto frontier and B) gave the highest possible yield when choosing an optimal temperature (see App. \ref{app:temp} for more details).

\vspace{-5pt}
\subsection{Human Eval}
\vspace{-3pt}
We extend the verifiable task experiments with a human evaluation on open-ended chat prompts: NoveltyBench-Curated (100 prompts, \citealt{zhang2025noveltybenchevaluatinglanguagemodels}) and Infinite-Chats-Eval (100 prompts, yet to be published, obtained from the authors). However, \method does not optimize for chat capabilities, but rather for fitting to \texttt{description/input/output}. In order to elicit chat capabilities in-context, we try two approaches: zero-shot chat, where we prompt with \texttt{description: You are a helpful AI assistant}, \texttt{input: <prompt>}; and few-shot chat, where we utilize the same description and four examples of prompt inputs and chat responses as outputs. Additionally, we use a similar prompt for the pretrained model as a baseline, with the description, a prefix for the prompt of \texttt{User:}, and an output prefix of \texttt{Assistant:}, zero-shot and with the same four few-shot examples (similar to URIAL, \citealt{lin2023unlockingspellbasellms}). More details in App. \ref{app:diversityvalidity}.

For each prompt, we generate four completions from the model. We recruit annotators to judge whether a given generation is a valid response to the prompt. Each generation is annotated by four annotators, and we count the generation as valid if three of four annotators marked it as valid. Overall, annotators had a 73\% pairwise agreement rate. Due to the cost of the evaluation, we only annotate generations for one model family, \texttt{gemma-3-12b}. For additional evaluation details, see App. \ref{app:human_annotation}. For calculating diversity, we follow NoveltyBench's approach and utilize their \texttt{deberta-v3-large}-based model for assigning two generations as duplicates. We report the Pairwise Uniqueness \%, or the probability that any two valid generations are not considered duplicates, along with yield. Results are in Tab.~\ref{fig:diverse_verify_bar}.

\textbf{Few-shot pretrained models improve yield over instruct models.} While lagging in validity, pretrained models produce much more diverse responses than their instruct counterparts, and are able to achieve $>$$40$\% validity from few-shot chat examples, improving yield and offering a strong baseline.

\textbf{\method offers a Pareto improvement on diversity/validity and improves yield over baselines on NoveltyBench-Curated.} On NoveltyBench-Curated, our method offers higher validity than the pretrained model, while offering substantially higher diversity than the instruct model. This improvement results in a statistically significant increase in yield over the baselines. On Infinite-Chats, the pretrained models and our models do not perform significantly differently, covering roughly the same space on the Pareto frontier and on yield. While disambiguating the reason for the differing performance may require further investigation, we do note that many of the Infinite-Chat eval prompts have specific requirements, such as ``In five words", ``In a couple of paragraphs," etc., which our models often fail to adhere to. In contrast, the NoveltyBench-Curated prompts are far more open-ended. It may be that our model performs best at generating shorter outputs, and future work may be needed to enhance precise instruction-following while maintaining diversity. However, on both datasets, the instruct model has significantly lower yield and diversity.

\vspace{-5pt}
\section{Distributional Alignment and Pluralism}
\label{sec:distributionalalignment}
\vspace{-3pt}

Next, we evaluate our system's ability to steer to match a target distribution. We utilize seven held-out datasets \footnote{Machine Personality Inventory \citep{jiang2023evaluatinginducingpersonalitypretrained}, Rotten Tomatoes \citep{RT_1970_2024_Community_2024}, NYTimes Books \citep{meister2024benchmarkingdistributionalalignmentlarge}, GlobalOQA \citep{durmus2023measuring}, Urn (ours, new contribution), Habermas \citep{doi:10.1126/science.adq2852}, Number Game \citep{BigelowPiantadosi2016_NumberGame, Tenenbaum1999ConceptLearning}.} mainly focusing on human response distributions and a synthetic random draws task. We prompt models zero-shot with a description of the setting and a target question. We then calculate the probability of each possible valid output, normalize, and calculate Jensen-Shannon divergence from the target distribution. We also measure coverage, or the total probability mass on the set of valid answers. Results are in Table \ref{tab:distributionalalignment}, and takeaways are as follows. (More details in App. \ref{app:distributionalalignment}.)

\begin{table}[h]
\centering
\tiny
\setlength{\tabcolsep}{6pt}
\renewcommand{\arraystretch}{1.15}
\begin{tabular}{l*{9}{c}}
\toprule
\textit{\textbf{Distributional Alignment: JS-Divergence $\downarrow$}} & \multicolumn{3}{c}{\textbf{gemma-3-12b}} & \multicolumn{3}{c}{\textbf{Qwen3-14B}} & \multicolumn{3}{c}{\textbf{Llama-3.1-8B}} \\
\cmidrule(lr){2-4} \cmidrule(lr){5-7} \cmidrule(lr){8-10}
\textbf{Dataset} & \textbf{\methodabb (ours)} & \textbf{PT} & \textbf{IT} & \textbf{\methodabb (ours)} & \textbf{PT} & \textbf{IT} & \textbf{\methodabb (ours)} & \textbf{PT} & \textbf{IT} \\
\midrule
\textbf{Machine Personality Inventory} (N=120, $|Y|$=6) & \textbf{0.083} & 0.126 & 0.347 & \textbf{0.100} & \textbf{0.093} & 0.405 & \textbf{0.063} & 0.087 & 0.131 \\
\textbf{Rotten Tomatoes} (N=1000, $|Y|$=2) & \textbf{0.032} & \textbf{0.032} & 0.134 & \textbf{0.028} & \textbf{0.028} & 0.122 & \textbf{0.035} & \textbf{0.035} & 0.086 \\
\textbf{NYTimes Books} (N=940, $|Y|$=4) & \textbf{0.051} & 0.063 & 0.328 & \textbf{0.070} & 0.088 & 0.344 & \textbf{0.046} & 0.061 & 0.247 \\
\textbf{GlobalOQA} (N=1000, $|Y|$$\leq$6) & \textbf{0.077} & 0.094 & 0.270 & \textbf{0.090} & \textbf{0.088} & 0.274 & \textbf{0.091} & 0.108 & 0.163 \\
\textbf{Urn} (N=1000, $|Y|$$\leq$6) & \textbf{0.021} & 0.071 & 0.185 & \textbf{0.051} & 0.059 & 0.198 & \textbf{0.032} & 0.124 & 0.086 \\
\textbf{Habermas} (N=658, $|Y|$=7) & \textbf{0.149} & \textbf{0.147} & 0.436 & \textbf{0.123} & \textbf{0.127} & 0.434 & \textbf{0.151} & \textbf{0.155} & 0.242 \\
\textbf{Number Game} (N=1000, $|Y|$=2) & \textbf{0.051} & \textbf{0.049} & 0.138 & 0.052 & \textbf{0.043} & 0.131 & \textbf{0.055} & \textbf{0.060} & 0.094 \\
\bottomrule
\end{tabular}
\small
\caption{
Distributional alignment results.
Instruction-tuning drastically hurts distributional alignment.
\method generalizes to unseen tasks and improves or matches distributional alignment compared to the pretrained model.
Best result (within 95\% statistical significance) in bold.
$N$ is the number of distinct instances, $|Y|$ is the number of possible outputs.
}
\label{tab:distributionalalignment}
\end{table}

\textbf{Instruction-tuned models have higher distributional divergence than pretrained models.} In line with prior work \citep{sorensen2024roadmappluralisticalignment}, we find that instruction-tuned models show higher distributional divergence than pretrained models on all tasks. We believe that this is in large part due to their low-entropy, spiky distributions. In other words, for distribution matching, current instruction-tuning categorically hurts performance compared to the pretrained model.

\textbf{\method generally improves distributional alignment over pretrained models.} Out of 21 model/dataset comparisons, \method improves distributional alignment in 10 cases, matches PT models in 10 cases, and degrades performance in 1 case. Pretrained models are a strong baseline---the pretraining objective entirely consists of trying to estimate a well-calibrated distribution over the next token.  To our knowledge, ours is the \textit{first method to improve distributional alignment on unseen datasets} over pretrained models.

\begin{wrapfigure}{r}{0.4\columnwidth}
\centering
\vspace{-15pt}
\includegraphics[width=0.4\textwidth]{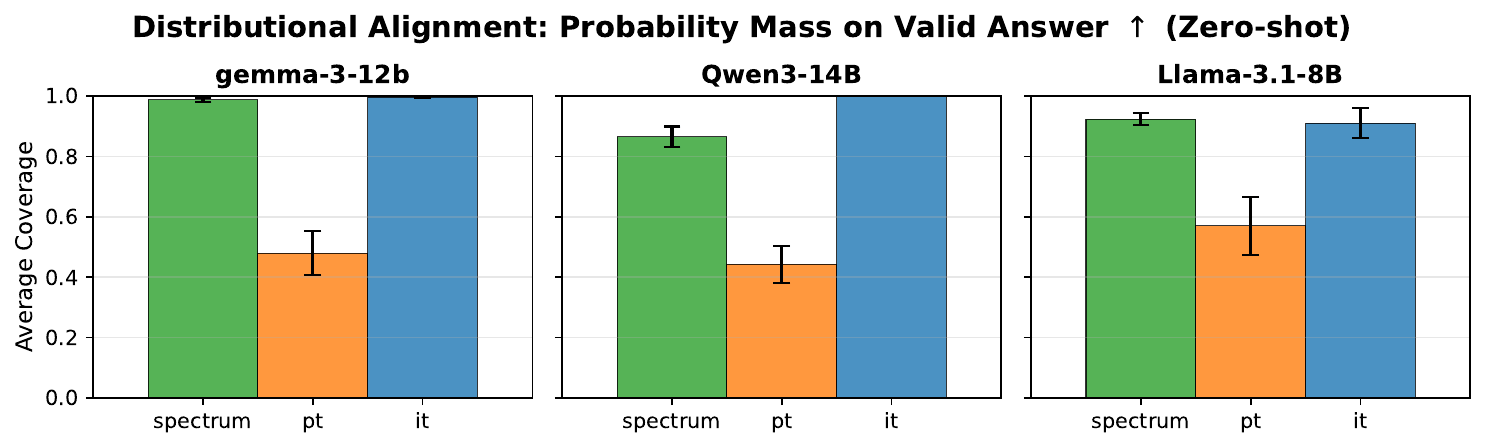}
\small
\vspace{-20pt}
\caption{
Valid answer coverage ($\uparrow$).
}
\label{fig:coverage}
\end{wrapfigure}

\textbf{\method improves coverage of valid answers over pretrained models and roughly matches instruction-tuned models.} For each of the datasets, there is a limited set of valid answers. Pretrained models often struggle to shift their probability mass based on instructions in a zero-shot manner to only cover the valid output distribution, achieving $\sim50\%$ coverage in our evaluation. In contrast, \method achieves $>90\%$ coverage, nearly matching the instruction-tuned model coverage (Fig \ref{fig:coverage}).

\vspace{-5pt}
\section{Ablations and General Capabilities}
\label{sec:additionalexperiments}
\vspace{-3pt}

In Table \ref{tab:ablations}, we ablate parts of \method in order to further disentangle the effect of each component. We report averaged results for all three desiderata across all models and tasks. In A), we see the normalized data from the prior sections, illustrating Spectrum-Tuned models improvements over base and default instruct models.

\textbf{\resource's selective loss is important for performance on all desiderata.} In B), we hold the Spectrum Tuning data constant, and ablate the training method. We compare against training on the first output only (similar to Instruct-SFT),\footnote{However, we also consider this distinct from traditional instruction-tuning, as the focus is on fitting the data generation task of the description as opposed to generating a helpful chat assistant response.} training on the last output only (similar to MetaICL, \citealt{min2022metaicllearninglearncontext}), and calculating loss on all tokens, including \texttt{description/inputs}. We find that training on the first output only causes a degradation in few-shot learning capabilities (ICL loss, few-shot yield), and training on the last output only causes across the board degradation, especially on zero-shot tasks (distributional alignment, description yield). Training on all tokens (including \texttt{description/input}) leads to slight degradations across the board.

\textbf{Training on capability-focused data only underperforms training on \resource.} We train on a subset of data in the same format as \resource, but focused on capability data instead of data requiring steerability (Table \ref{tab:ablations}, C). We find that including the \resource data is important for eliciting the desiderata.
Finally, we find that D) the default weight initialization (PT model weights, IT special token embeddings) overall elicits the best performance, although initializing the special tokens with bracket token embeddings seems to improve the multiple-choice accuracy and distributional alignment.

While the default recipe offers strong performance, future work could i) further optimize hyperparameters (as we have done limited optimization),\footnote{In fact, after running the main suite of experiments, we suspected that our models were somewhat underfit. We found that simply reducing the batch size resulted in significant gains in distributional alignment and yield (see App. \ref{app:betterhparams} for more details). We believe that this illustrates exciting opportunities for further optimization and improvements to improve performance---the performance ceiling has not been hit.} ii) reduce reliance on initializing the special tokens from IT models, and iii) probe which data is most important in eliciting gains.

\begin{table}[t]
\vspace{-7pt}
\centering
\tiny
\setlength{\tabcolsep}{2pt}
\renewcommand{\arraystretch}{1.1}
\adjustbox{width=\linewidth}{
\begin{tabular}{ccccccccccccc}
\multicolumn{6}{c}{\textbf{Ablation Components}} & \multicolumn{3}{c}{\textbf{ICL Steerability}} & \textbf{Dist. Align.} & \multicolumn{3}{c}{\textbf{Valid Output Coverage}} \\
\cmidrule(lr){1-6} \cmidrule(lr){7-9} \cmidrule(lr){10-10} \cmidrule(lr){11-13}
\textbf{Abl.} & \textbf{Weight} & \textbf{Special Tokens} & \textbf{Train on} & \textbf{\# Train} & \textbf{Loss only} & \textbf{MC Loss} & \textbf{MC Acc} & \textbf{Free-text} & \textbf{Dist. Align.} & \textbf{Yield -} & \textbf{Yield -} & \textbf{Yield - 3-shot} \\
\textbf{\#} & \textbf{Init} & \textbf{Embedding Init} & \textsc{Spectrum Suite} & \textbf{Seqs} & \textbf{Outputs} & \textbf{(Norm.)} & \textbf{(Norm.)} & \textbf{Loss (Norm.)} & \textbf{JS-Div.} & \textbf{Description} & \textbf{3-shot} & \textbf{+ Description} \\
\midrule
\multicolumn{5}{l}{\textit{\textbf{A - Default:} 1) Spectrum Tuning, 2) Pretrained, and 3) Instruction-Tuned}} \\
\rowcolor{gray!15} 1 & PT & IT & \checkmark & 38.8k & \checkmark & \textbf{1.00} & \textbf{1.00} & \textbf{1.00} & \textbf{.069} & \textbf{36.7} & \textbf{42.1} & \textbf{49.2} \\
2 & PT & - & $\times$ (PT prompt) & - & - & \underline{1.19} & \underline{0.99} & \textbf{1.00} & \underline{.083} & 5.8 & \underline{37.2} & \underline{44.2} \\
3 & IT & IT & $\times$ (IT prompt) & - & - & 2.62 & 0.98 & 1.30 & .228 & \underline{11.7} & 21.5 & 20.7 \\
\midrule
\multicolumn{13}{l}{\textit{\textbf{B - Training method ablations}: 1) Default; 4) Loss only first output (Instruct-SFT on S-Suite); 5) Loss only last output (Meta-ICL on S-Suite); 6) Loss on all tokens (S-Suite)}} \\
\rowcolor{gray!15} 1 & PT & IT & \checkmark & 38.8k & \checkmark & \textbf{1.00} & \textbf{1.00} & \textbf{1.00} & \underline{.069} & \underline{36.7} & \textbf{42.1} & \textbf{49.2} \\
4 & PT & IT & \checkmark & 38.8k & first only & 1.03 & \textbf{1.00} & 1.01 & \textbf{.067} & \textbf{37.9} & 33.0 & 44.0 \\
5 & PT & IT & \checkmark & 38.8k & last only & 1.02 & 0.99 & \textbf{1.00} & .103 & 17.1 & 35.4 & 39.6 \\
6 & PT & IT & \checkmark & 38.8k & $\times$ & \underline{1.01} & 0.98 & \textbf{1.00} & .075 & 33.0 & \underline{40.6} & \underline{47.1} \\
\midrule
\multicolumn{13}{l}{\textit{\textbf{C - Data ablation}: 7) Train only on capability / knowledge elicitation data, 8) Train on Spectrum Suite, data size matched to capability data}} \\
7 & PT & IT & $\times$ (capability data) & 3.9k & \checkmark & \textbf{1.03} & 0.99 & 1.02 & .111 & 12.7 & 21.2 & 39.5 \\
8 & PT & IT & \checkmark & 3.9k & \checkmark & \textbf{1.03} & \textbf{1.00} & \textbf{1.01} & \textbf{.086} & \textbf{21.8} & \textbf{35.5} & \textbf{40.8} \\
\midrule
\multicolumn{13}{l}{\textit{\textbf{D - Weight Init Ablation}: Spectrum Tuning with 1) Default weight init; 9) PT init, bracket as special token embed, 10) PT init, random special token embed, 11) IT init}} \\
\rowcolor{gray!15} 1 & PT & IT & \checkmark & 38.8k & \checkmark & \textbf{1.00} & 1.00 & \textbf{1.00} & \underline{.069} & \textbf{36.7} & \textbf{42.1} & \textbf{49.2} \\
9 & PT & \texttt{<</>>} (PT) & \checkmark & 38.8k & \checkmark & 1.43 & \textbf{1.03} & \underline{1.02} & \textbf{.063} & 28.0 & 30.0 & 33.0 \\
10 & PT & Random & \checkmark & 38.8k & \checkmark & 1.44 & 0.87 & 1.25 & .079 & 21.0 & 21.0 & 26.4 \\
11 & IT & IT & \checkmark & 38.8k & \checkmark & 1.08 & \underline{1.02} & 1.05 & \underline{.069} & \underline{33.4} & \underline{42.0} & \underline{45.2} \\
\bottomrule
\end{tabular}
}
\caption{Ablations, averaged across models and tasks. Shaded rows are default Spectrum-Tuned results. We show averaged results for A) the default setup, B) training on \resource with different methods, C) training on capability-focused data in place of \resource,
and D) different model weight initializations.
Best result within each ablation is bolded, and second best is underlined. ICL Steerability results are normalized to the default configuration.}
\label{tab:ablations}
\vspace{-10pt}
\end{table}

\textbf{\method does not harm general model capabilities.} Lastly, we evaluate whether our method affects general model capabilities. While we do not necessarily expect our method to improve upon standard evaluations where there is a single correct answer, we want to understand if it degrades performance compared to pretrained models.
While we find that Spectrum-Tuned models generally perform worse than instruction-tuned variants (as expected), we find that Spectrum-Tuned models perform similarly to the pretrained models on which they are based.
(c.f. App. \ref{app:capabilitydatasets})

\vspace{-5pt}
\section{Related Work}
\label{sec:relatedwork}
\vspace{-3pt}

\textbf{Diversity, distributional alignment, and steerability.} Several other works have documented diversity collapse in LLMs \citep{shumailov2023curse, dohmatob2024strong, yang2024model, zhang2024forcing, li2024predicting, west2025basemodelsbeataligned}, often linking it to alignment \citep{murthy2024one, kirk2024benefits, kirk2023understanding} or insufficient training data diversity \citep{chen2024diversitysyntheticdataimpact}. Potential consequences of diversity collapse include reduced creativity, loss of minority perspectives, spread of bias, and overall decline in model utility and trustworthiness \citep{anderson2024homogenization, kapania2024simulacrum}. Distributional alignment has been explored by a few prior works \citep{meister2024benchmarkingdistributionalalignmentlarge, durmus2023measuring, sorensen2024roadmappluralisticalignment}, but literature here is far less developed. Additionally, other works have focused on measuring steerability to system messages \citep{lee2024aligningthousandspreferencesmessage}, persona descriptions \citep{miehling2025evaluatingpromptsteerabilitylarge, castricato2024personareproducibletestbedpluralistic}, and values or attributes \citep{sorensen2024roadmappluralisticalignment, sorensen2025valueprofilesencodinghuman}. Our work builds on these directions by generalizing steerability to include any in-context information, including examples, and evaluating on a broader swath of distributions.

\textbf{Pluralistic alignment and integrating disagreement into LLMs.}
Many have recently challenged the idea of a single ground truth \citep{aroyo2023dicesdatasetdiversityconversational, basile-etal-2021-need, Gordon_2022}. Pluralistic alignment \citep{sorensen2024roadmappluralisticalignment, kirk2024prismalignmentdatasetparticipatory} is concerned with integrating diverse values and perspectives directly into the alignment process. Steerability in particular is related to user fairness and ensuring that AI systems are usable by diverse stakeholders \citep{alamdari2024consideratepathwaypluralisticalignment}.

\textbf{Related Methods}
\citet{zhang2024forcing} found that training on samples from diffuse distributions helps LLMs to avoid mode collapse, and served as inspiration for some experiments. \method is similar in spirit, but also includes in-context samples and leverages orders of magnitude more data.
Entropy maximization in finetuning can help increase diversity \citep{li2025jointlyreinforcingdiversityquality}.
MetaICL \citep{min2022metaicllearninglearncontext} uses in-context examples as in our method, but focuses on NLP datasets with a single ground truth and only trains on the last example. Centaur \citep{binz2024centaurfoundationmodelhuman} similarly modifies cross-entropy loss to only focus on tokens of interest, but focuses on a different data distribution (cognitive-science human experiments). Some very recent works have somewhat improved the diversity/validity Pareto frontier by adding some sort of diversity regularization to preference optimization or RL reward \citep{lanchantin2025diversepreferenceoptimization, chung2025modifyinglargelanguagemodel, li2025jointlyreinforcingdiversityquality}. Finally, several recent papers have found that prompting instruct models for multiple samples in-context can help to mitigate mode collapse  \citep{zhang2025verbalizedsamplingmitigatemode, zhang2025cultivating, zhang2025noveltybenchevaluatinglanguagemodels}.

\vspace{-7pt}
\section{Discussion and Conclusion}
\label{sec:discussionandconclusion}
\vspace{-3pt}
We have outlined three desiderata for conditional distributional modeling with LLMs: in-context steerability, output space coverage, and distributional alignment, and shown across three model families that current post-training can systematically hurt these properties. These results have implications for user steerability---e.g., when possible, pretrained models may be preferred over instruction-tuned models when steering to a particular user in a well-calibrated way is important.\footnote{However, access to the pretrained model is restricted in many proprietary cases. This illustrates a gap: Can companies offer very steerable and distributionally-aligned models, while maintaining safety constraints?} In addition, we have introduced \resource and \method, a resource and post-training method for enhancing these desiderata. Models trained with \method usually match or exceed their pretrained counterparts at these properties---to our knowledge, ours is the first method to improve upon pretrained models at distributional alignment or in-context steerability. However, much work remains. Promising directions for future work include
1) exploring which data is most important for eliciting the desiderata;
2) further characterizing why and how instruction-tuning hurts
in-context steerability;
3) more work to combine the strengths of instruction-tuned models and \method models (e.g., \citealt{zhu2025bareleveragingbaselanguage});\footnote{On the other hand, it is possible that top-1 chat performance and our desiderata are so fundamentally in tension, that we may need to specialize models to either top-1 chat performance or our desiderata, and select the appropriate model for each use case or combine strengths at inference (e.g., \citealt{zhu2025bareleveragingbaselanguage})}
and 4) scaling \method to larger models and more data.

\section*{Ethics Statement}
In this paper, we seek to enable AI systems that can work for a variety of perspectives and estimate human preferences and opinions in a well-calibrated manner. We believe that these are net positive developments, allowing AI systems to work properly for more people. Additionally, well-calibrated human preferences may be especially important as AI systems are used agentically - it will be important that an agent have a good model of what the user wants, as opposed to a modal preference. Calibration, where current instruction-tuned systems really struggle, can be especially important for agents to safely act autonomously when they are (properly) very confident about a users' preference, and ask for direction when they are less confident.

With \resource, we perform experiments on several datasets which may include personal information such as demographics. However, all included datasets are anonymized, we attempt to use the data only in line with their intended use, and we do not distribute the underlying datasets in \resource directly. Instead, we refer people interested in extending our work to the original data sources, and provide only the code to unify the data into the \texttt{description/input/output} format. Because of this, we believe that our compilation of \resource does not pose an additional privacy risk.

\section*{Reproducibility Statement}
We have attempted to ensure that every portion of the paper is reproducible, and
release code\footnote{\url{https://github.com/tsor13/spectrum}}
containing: \resource construction, including processing and pointers to hydrate each dataset; \method training code; and code for running all evaluations. We also release the weights for all trained \method models.\footnote{\url{https://huggingface.co/collections/tsor13/spectrum-68dac670f618224845c0fb7d}}
We include additional training details on hardware and hyperparameters used in App. \ref{app:trainingdetails} and additional experimental details in App. \ref{app:prompttemplates}, \ref{app:diversityvalidity}, \ref{app:distributionalalignment}. In App. \ref{app:exampletaskpromptstest}, we show demonstrative example prompts for each test task in \resource and include example prompts for remaining tasks in supplementary
materials.\footnote{\url{https://tsor13.github.io/files/spectrumprompts.pdf}}

\section*{Acknowledgments}
We would like to thank Hannaneh Hajishirzi, Sewon Min, Luke Zettlemoyer, Peter West, and Kshitish Ghate for draft feedback and Christopher Sorensen for help designing the \method logo. This research was supported in part by DARPA under the ITM program (FA8650-23-C-7316) and by the AI 2050 Schmidt Sciences Senior Fellowship.

\newpage

\bibliography{iclr2026_conference}

@misc{zhang2024forcingdiffusedistributionslanguage,
      title={Forcing Diffuse Distributions out of Language Models}, 
      author={Yiming Zhang and Avi Schwarzschild and Nicholas Carlini and Zico Kolter and Daphne Ippolito},
      year={2024},
      eprint={2404.10859},
      archivePrefix={arXiv},
      primaryClass={cs.CL},
      url={https://arxiv.org/abs/2404.10859}, 
}

@misc{binz2024centaurfoundationmodelhuman,
      title={Centaur: a foundation model of human cognition}, 
      author={Marcel Binz and Elif Akata and Matthias Bethge and Franziska Brändle and Fred Callaway and Julian Coda-Forno and Peter Dayan and Can Demircan and Maria K. Eckstein and Noémi Éltető and Thomas L. Griffiths and Susanne Haridi and Akshay K. Jagadish and Li Ji-An and Alexander Kipnis and Sreejan Kumar and Tobias Ludwig and Marvin Mathony and Marcelo Mattar and Alireza Modirshanechi and Surabhi S. Nath and Joshua C. Peterson and Milena Rmus and Evan M. Russek and Tankred Saanum and Natalia Scharfenberg and Johannes A. Schubert and Luca M. Schulze Buschoff and Nishad Singhi and Xin Sui and Mirko Thalmann and Fabian Theis and Vuong Truong and Vishaal Udandarao and Konstantinos Voudouris and Robert Wilson and Kristin Witte and Shuchen Wu and Dirk Wulff and Huadong Xiong and Eric Schulz},
      year={2024},
      eprint={2410.20268},
      archivePrefix={arXiv},
      primaryClass={cs.LG},
      url={https://arxiv.org/abs/2410.20268}, 
}

@misc{zhu2025bareleveragingbaselanguage,
      title={BARE: Leveraging Base Language Models for Few-Shot Synthetic Data Generation}, 
      author={Alan Zhu and Parth Asawa and Jared Quincy Davis and Lingjiao Chen and Boris Hanin and Ion Stoica and Joseph E. Gonzalez and Matei Zaharia},
      year={2025},
      eprint={2502.01697},
      archivePrefix={arXiv},
      primaryClass={cs.CL},
      url={https://arxiv.org/abs/2502.01697}, 
}

@misc{brown2020languagemodelsfewshotlearners,
      title={Language Models are Few-Shot Learners}, 
      author={Tom B. Brown and Benjamin Mann and Nick Ryder and Melanie Subbiah and Jared Kaplan and Prafulla Dhariwal and Arvind Neelakantan and Pranav Shyam and Girish Sastry and Amanda Askell and Sandhini Agarwal and Ariel Herbert-Voss and Gretchen Krueger and Tom Henighan and Rewon Child and Aditya Ramesh and Daniel M. Ziegler and Jeffrey Wu and Clemens Winter and Christopher Hesse and Mark Chen and Eric Sigler and Mateusz Litwin and Scott Gray and Benjamin Chess and Jack Clark and Christopher Berner and Sam McCandlish and Alec Radford and Ilya Sutskever and Dario Amodei},
      year={2020},
      eprint={2005.14165},
      archivePrefix={arXiv},
      primaryClas={cs.CL},
      url={https://arxiv.org/abs/2005.14165}, 
}

@misc{sorensen2024roadmappluralisticalignment,
      title={A Roadmap to Pluralistic Alignment}, 
      author={Taylor Sorensen and Jared Moore and Jillian Fisher and Mitchell Gordon and Niloofar Mireshghallah and Christopher Michael Rytting and Andre Ye and Liwei Jiang and Ximing Lu and Nouha Dziri and Tim Althoff and Yejin Choi},
      year={2024},
      eprint={2402.05070},
      archivePrefix={arXiv},
      primaryClass={cs.AI},
      url={https://arxiv.org/abs/2402.05070}, 
}

@misc{meister2024benchmarkingdistributionalalignmentlarge,
      title={Benchmarking Distributional Alignment of Large Language Models}, 
      author={Nicole Meister and Carlos Guestrin and Tatsunori Hashimoto},
      year={2024},
      eprint={2411.05403},
      archivePrefix={arXiv},
      primaryClass={cs.CL},
      url={https://arxiv.org/abs/2411.05403}, 
}

@inproceedings{liu-etal-2023-afraid,
    title = "We{'}re Afraid Language Models Aren{'}t Modeling Ambiguity",
    author = "Alisa Liu and Zhaofeng Wu and Julian Michael and Alane Suhr and Peter West and Alexander Koller and Swabha Swayamdipta and Noah A. Smith and Yejin Choi",
    booktitle = "Proceedings of the 2023 Conference on Empirical Methods in Natural Language Processing",
    month = dec,
    year = "2023",
    address = "Singapore",
    publisher = "Association for Computational Linguistics",
    url = "https://aclanthology.org/2023.emnlp-main.51",
    doi = "10.18653/v1/2023.emnlp-main.51",
    pages = "790--807",
}

@misc{SSA_BabyNames_National_2025,
  author       = {{Social Security Administration}},
  title        = {Baby Names from Social Security Card Applications — National Data},
  year         = {2025},
  howpublished = {\url{https://www.ssa.gov/oact/babynames/limits.html}},
  note         = {Data are from a 100\% sample of Social Security card applications; names with <5 occurrences are suppressed. Accessed 2025-09-21.},
}

@misc{Kolyada2020WebisCMV20,
  author       = {Kolyada, Nikolay and Al-Khatib, Khalid and V{\"o}lske, Michael and Syed, Shahbaz and Stein, Benno},
  title        = {Webis ChangeMyView Corpus 2020 (Webis-CMV-20)},
  year         = {2020},
  publisher    = {Zenodo},
  doi          = {10.5281/zenodo.3778298},
  url          = {https://doi.org/10.5281/zenodo.3778298},
  note         = {Version v1; file used: \texttt{threads.jsonl}. Accessed 2025-09-21.}
}

@misc{OpenAI_CA1_2025,
  author       = {{OpenAI}},
  title        = {Collective Alignment 1: Public Input on Model Defaults (Version 1.0)},
  year         = {2025},
  howpublished = {\url{https://huggingface.co/datasets/openai/collective-alignment-1}},
  note         = {Dataset; accessed 2025-09-21}
}

@article{zhang2025cultivating,
  title   = {Cultivating Pluralism In Algorithmic Monoculture: The Community Alignment Dataset},
  author  = {Lily Hong Zhang and Smitha Milli and Karen Jusko and Jonathan Smith and Brandon Amos and Wassim and Bouaziz and Manon Revel and Jack Kussman and Lisa Titus and Bhaktipriya Radharapu and Jane Yu and Vidya Sarma and Kris Rose and Maximilian Nickel},
  year    = {2025},
  journal = {arXiv preprint arXiv: 2507.09650}
}

@misc{aroyo2023dicesdatasetdiversityconversational,
      title={DICES Dataset: Diversity in Conversational AI Evaluation for Safety}, 
      author={Lora Aroyo and Alex S. Taylor and Mark Diaz and Christopher M. Homan and Alicia Parrish and Greg Serapio-Garcia and Vinodkumar Prabhakaran and Ding Wang},
      year={2023},
      eprint={2306.11247},
      archivePrefix={arXiv},
      primaryClass={cs.HC},
      url={https://arxiv.org/abs/2306.11247}, 
}

@misc{fish2025generativesocialchoice,
      title={Generative Social Choice}, 
      author={Sara Fish and Paul Gölz and David C. Parkes and Ariel D. Procaccia and Gili Rusak and Itai Shapira and Manuel Wüthrich},
      year={2025},
      eprint={2309.01291},
      archivePrefix={arXiv},
      primaryClass={cs.GT},
      url={https://arxiv.org/abs/2309.01291}, 
}

@article{cobbe2021gsm8k,
  title={Training Verifiers to Solve Math Word Problems},
  author={Cobbe, Karl and Kosaraju, Vineet and Bavarian, Mohammad and Chen, Mark and Jun, Heewoo and Kaiser, Lukasz and Plappert, Matthias and Tworek, Jerry and Hilton, Jacob and Nakano, Reiichiro and Hesse, Christopher and Schulman, John},
  journal={arXiv preprint arXiv:2110.14168},
  year={2021}
}

@misc{Neiman2018_HaikuDL,
  author       = {Neiman, Jeremy},
  title        = {Generating Haiku with Deep Learning},
  year         = {2018},
  month        = dec,
  howpublished = {\emph{Towards Data Science}},
  url          = {https://towardsdatascience.com/generating-haiku-with-deep-learning-dbf5d18b4246/},
  note         = {Accessed 2025-09-21}
}

@misc{kumar2021designingtoxiccontentclassification,
      title={Designing Toxic Content Classification for a Diversity of Perspectives}, 
      author={Deepak Kumar and Patrick Gage Kelley and Sunny Consolvo and Joshua Mason and Elie Bursztein and Zakir Durumeric and Kurt Thomas and Michael Bailey},
      year={2021},
      eprint={2106.04511},
      archivePrefix={arXiv},
      primaryClass={cs.SI},
      url={https://arxiv.org/abs/2106.04511}, 
}

@misc{wang2024helpsteer2preferencecomplementingratingspreferences,
      title={HelpSteer2-Preference: Complementing Ratings with Preferences}, 
      author={Zhilin Wang and Alexander Bukharin and Olivier Delalleau and Daniel Egert and Gerald Shen and Jiaqi Zeng and Oleksii Kuchaiev and Yi Dong},
      year={2024},
      eprint={2410.01257},
      archivePrefix={arXiv},
      primaryClass={cs.LG},
      url={https://arxiv.org/abs/2410.01257}, 
}

@InProceedings{maas-EtAl:2011:ACL-HLT2011,
  author    = {Maas, Andrew L.  and  Daly, Raymond E.  and  Pham, Peter T.  and  Huang, Dan  and  Ng, Andrew Y.  and  Potts, Christopher},
  title     = {Learning Word Vectors for Sentiment Analysis},
  booktitle = {Proceedings of the 49th Annual Meeting of the Association for Computational Linguistics: Human Language Technologies},
  month     = {June},
  year      = {2011},
  address   = {Portland, Oregon, USA},
  publisher = {Association for Computational Linguistics},
  pages     = {142--150},
  url       = {http://www.aclweb.org/anthology/P11-1015}
}

@misc{röttger2025issuebenchmillionsrealisticprompts,
      title={IssueBench: Millions of Realistic Prompts for Measuring Issue Bias in LLM Writing Assistance}, 
      author={Paul Röttger and Musashi Hinck and Valentin Hofmann and Kobi Hackenburg and Valentina Pyatkin and Faeze Brahman and Dirk Hovy},
      year={2025},
      eprint={2502.08395},
      archivePrefix={arXiv},
      primaryClass={cs.CL},
      url={https://arxiv.org/abs/2502.08395}, 
}

@misc{trexmatt_jeopardy_csv_2014,
  author       = {trexmatt},
  title        = {200{,}000+ Jeopardy! Questions (CSV dump from J-Archive)},
  year         = {2014},
  howpublished = {\url{https://drive.google.com/file/d/0BwT5wj_P7BKXUl9tOUJWYzVvUjA/view?resourcekey=0-uFrn8bQkUfSCvJlmtKGCdQ}},
  note         = {Original announcement on r/datasets; accessed 2025-09-21.}
}

@inproceedings{jang-frassinelli-2024-generalizable,
    title = "Generalizable Sarcasm Detection is Just Around the Corner, of Course!",
    author = "Jang, Hyewon  and
      Frassinelli, Diego",
    editor = "Duh, Kevin  and
      Gomez, Helena  and
      Bethard, Steven",
    booktitle = "Proceedings of the 2024 Conference of the North American Chapter of the Association for Computational Linguistics: Human Language Technologies (Volume 1: Long Papers)",
    month = jun,
    year = "2024",
    address = "Mexico City, Mexico",
    publisher = "Association for Computational Linguistics",
    url = "https://aclanthology.org/2024.naacl-long.238/",
    doi = "10.18653/v1/2024.naacl-long.238",
    pages = "4238--4249"
}

@inproceedings{casola-etal-2024-multipico,
    title = "{M}ulti{PIC}o: Multilingual Perspectivist Irony Corpus",
    author = "Casola, Silvia  and
      Frenda, Simona  and
      Lo, Soda Marem  and
      Sezerer, Erhan  and
      Uva, Antonio  and
      Basile, Valerio  and
      Bosco, Cristina  and
      Pedrani, Alessandro  and
      Rubagotti, Chiara  and
      Patti, Viviana  and
      Bernardi, Davide",
    editor = "Ku, Lun-Wei  and
      Martins, Andre  and
      Srikumar, Vivek",
    booktitle = "Proceedings of the 62nd Annual Meeting of the Association for Computational Linguistics (Volume 1: Long Papers)",
    month = aug,
    year = "2024",
    address = "Bangkok, Thailand",
    publisher = "Association for Computational Linguistics",
    url = "https://aclanthology.org/2024.acl-long.849/",
    doi = "10.18653/v1/2024.acl-long.849",
    pages = "16008--16021"
}

@inproceedings{lewidi,
title={LeWiDi-2025 at NLPerspectives: third Edition of the Learning with Disagreements Shared Task
},
author={Leonardelli,Elisa and Casola,Silvia and Peng,Siyao and Rizzi,Giulia and Basile,Valerio and Fersini,Elisabetta and Frassinelli,Diego and Jang,Hyewon and Pavlovic,Maja and Plank,Barbara and Poesio,Massimo},
booktitle={Proceedings of the 4th Workshop on Perspectivist Approaches to NLP (NLPerspectives)},
month = {nov},
year={2025},
publisher = {Association for Computational Linguistics}
}

@inproceedings{weber-genzel-etal-2024-varierr,
    title = "{V}ari{E}rr {NLI}: Separating Annotation Error from Human Label Variation",
    author = "Weber-Genzel, Leon  and
      Peng, Siyao  and
      De Marneffe, Marie-Catherine  and
      Plank, Barbara",
    editor = "Ku, Lun-Wei  and
      Martins, Andre  and
      Srikumar, Vivek",
    booktitle = "Proceedings of the 62nd Annual Meeting of the Association for Computational Linguistics (Volume 1: Long Papers)",
    month = aug,
    year = "2024",
    address = "Bangkok, Thailand",
    publisher = "Association for Computational Linguistics",
    url = "https://aclanthology.org/2024.acl-long.123/",
    doi = "10.18653/v1/2024.acl-long.123",
    pages = "2256--2269"
}

@misc{NetflixPrize_Kaggle_2009,
  author       = {{Netflix, Inc.}},
  title        = {Netflix Prize data},
  year         = {2009},
  publisher    = {Kaggle},
  url          = {https://www.kaggle.com/datasets/netflix-inc/netflix-prize-data},
  note         = {Dataset from the Netflix Prize competition; accessed 2025-09-21}
}

@Article{santurkar2023whose,
        title={Whose Opinions Do Language Models Reflect?},
        author={Shibani Santurkar and Esin Durmus and Faisal Ladhak and Cinoo Lee and Percy Liang and Tatsunori Hashimoto},
        year={2023},
        journal={arXiv preprint arXiv:2303.17548},
    }

@misc{CompDem_OpenData_2025,
  author       = {{The Computational Democracy Project}},
  title        = {Open Polis Data},
  year         = {2025},
  howpublished = {\url{https://github.com/compdemocracy/openData}},
  note         = {GitHub repository; data exports from select public Polis conversations; accessed 2025-09-21}
}

@misc{pei2023annotatordemographicsmattermeasuring,
      title={When Do Annotator Demographics Matter? Measuring the Influence of Annotator Demographics with the POPQUORN Dataset}, 
      author={Jiaxin Pei and David Jurgens},
      year={2023},
      eprint={2306.06826},
      archivePrefix={arXiv},
      primaryClass={cs.CL},
      url={https://arxiv.org/abs/2306.06826}, 
}

@misc{kirk2024prismalignmentdatasetparticipatory,
      title={The PRISM Alignment Dataset: What Participatory, Representative and Individualised Human Feedback Reveals About the Subjective and Multicultural Alignment of Large Language Models}, 
      author={Hannah Rose Kirk and Alexander Whitefield and Paul Röttger and Andrew Bean and Katerina Margatina and Juan Ciro and Rafael Mosquera and Max Bartolo and Adina Williams and He He and Bertie Vidgen and Scott A. Hale},
      year={2024},
      eprint={2404.16019},
      archivePrefix={arXiv},
      primaryClass={cs.CL},
      url={https://arxiv.org/abs/2404.16019}, 
}

@misc{mstz_titanic_2023,
  author    = {{mstz}},
  title     = {Titanic (survival) — Hugging Face Dataset},
  year      = {2023},
  publisher = {Hugging Face Datasets},
  url       = {https://huggingface.co/datasets/mstz/titanic},
  note      = {Subset: survival; 891 rows; accessed 2025-09-21}
}

@misc{moore2024largelanguagemodelsconsistent,
      title={Are Large Language Models Consistent over Value-laden Questions?}, 
      author={Jared Moore and Tanvi Deshpande and Diyi Yang},
      year={2024},
      eprint={2407.02996},
      archivePrefix={arXiv},
      primaryClass={cs.CL},
      url={https://arxiv.org/abs/2407.02996}, 
}

@article{Sorensen_2024,
   title={Value Kaleidoscope: Engaging AI with Pluralistic Human Values, Rights, and Duties},
   volume={38},
   ISSN={2159-5399},
   url={http://dx.doi.org/10.1609/aaai.v38i18.29970},
   DOI={10.1609/aaai.v38i18.29970},
   number={18},
   journal={Proceedings of the AAAI Conference on Artificial Intelligence},
   publisher={Association for the Advancement of Artificial Intelligence (AAAI)},
   author={Sorensen, Taylor and Jiang, Liwei and Hwang, Jena D. and Levine, Sydney and Pyatkin, Valentina and West, Peter and Dziri, Nouha and Lu, Ximing and Rao, Kavel and Bhagavatula, Chandra and Sap, Maarten and Tasioulas, John and Choi, Yejin},
   year={2024},
   month=mar, pages={19937–19947} }

@misc{clark2018thinksolvedquestionanswering,
      title={Think you have Solved Question Answering? Try ARC, the AI2 Reasoning Challenge}, 
      author={Peter Clark and Isaac Cowhey and Oren Etzioni and Tushar Khot and Ashish Sabharwal and Carissa Schoenick and Oyvind Tafjord},
      year={2018},
      eprint={1803.05457},
      archivePrefix={arXiv},
      primaryClass={cs.AI},
      url={https://arxiv.org/abs/1803.05457}, 
}

@misc{zheng2023judging,
      title={Judging LLM-as-a-judge with MT-Bench and Chatbot Arena}, 
      author={Lianmin Zheng and Wei-Lin Chiang and Ying Sheng and Siyuan Zhuang and Zhanghao Wu and Yonghao Zhuang and Zi Lin and Zhuohan Li and Dacheng Li and Eric. P Xing and Hao Zhang and Joseph E. Gonzalez and Ion Stoica},
      year={2023},
      eprint={2306.05685},
      archivePrefix={arXiv},
      primaryClass={cs.CL}
}

@misc{ramos2023bayesian,
      title={Bayesian Optimization of Catalysts With In-context Learning},
      author={Mayk Caldas Ramos and Shane S. Michtavy and Marc D. Porosoff and Andrew D. White},
      year={2023},
      eprint={2304.05341},
      archivePrefix={arXiv},
      primaryClass={physics.chem-ph}
}

@misc{dua2019dropreadingcomprehensionbenchmark,
      title={DROP: A Reading Comprehension Benchmark Requiring Discrete Reasoning Over Paragraphs}, 
      author={Dheeru Dua and Yizhong Wang and Pradeep Dasigi and Gabriel Stanovsky and Sameer Singh and Matt Gardner},
      year={2019},
      eprint={1903.00161},
      archivePrefix={arXiv},
      primaryClass={cs.CL},
      url={https://arxiv.org/abs/1903.00161}, 
}

@misc{qiu2025bayesianteachingenablesprobabilistic,
      title={Bayesian Teaching Enables Probabilistic Reasoning in Large Language Models}, 
      author={Linlu Qiu and Fei Sha and Kelsey Allen and Yoon Kim and Tal Linzen and Sjoerd van Steenkiste},
      year={2025},
      eprint={2503.17523},
      archivePrefix={arXiv},
      primaryClass={cs.CL},
      url={https://arxiv.org/abs/2503.17523}, 
}

@misc{durmus2023measuring,
      title={Towards Measuring the Representation of Subjective Global Opinions in Language Models}, 
      author={Esin Durmus and Karina Nyugen and Thomas I. Liao and Nicholas Schiefer and Amanda Askell and Anton Bakhtin and Carol Chen and Zac Hatfield-Dodds and Danny Hernandez and Nicholas Joseph and Liane Lovitt and Sam McCandlish and Orowa Sikder and Alex Tamkin and Janel Thamkul and Jared Kaplan and Jack Clark and Deep Ganguli},
      year={2023},
      eprint={2306.16388},
      archivePrefix={arXiv},
      primaryClass={cs.CL}
}

@misc{rein2023gpqagraduatelevelgoogleproofqa,
      title={GPQA: A Graduate-Level Google-Proof Q\&A Benchmark}, 
      author={David Rein and Betty Li Hou and Asa Cooper Stickland and Jackson Petty and Richard Yuanzhe Pang and Julien Dirani and Julian Michael and Samuel R. Bowman},
      year={2023},
      eprint={2311.12022},
      archivePrefix={arXiv},
      primaryClass={cs.AI},
      url={https://arxiv.org/abs/2311.12022}, 
}

@article{
doi:10.1126/science.adq2852,
author = {Michael Henry Tessler  and Michiel A. Bakker  and Daniel Jarrett  and Hannah Sheahan  and Martin J. Chadwick  and Raphael Koster  and Georgina Evans  and Lucy Campbell-Gillingham  and Tantum Collins  and David C. Parkes  and Matthew Botvinick  and Christopher Summerfield },
title = {AI can help humans find common ground in democratic deliberation},
journal = {Science},
volume = {386},
number = {6719},
pages = {eadq2852},
year = {2024},
doi = {10.1126/science.adq2852},
URL = {https://www.science.org/doi/abs/10.1126/science.adq2852},
eprint = {https://www.science.org/doi/pdf/10.1126/science.adq2852}}

@misc{zellers2019hellaswagmachinereallyfinish,
      title={HellaSwag: Can a Machine Really Finish Your Sentence?}, 
      author={Rowan Zellers and Ari Holtzman and Yonatan Bisk and Ali Farhadi and Yejin Choi},
      year={2019},
      eprint={1905.07830},
      archivePrefix={arXiv},
      primaryClass={cs.CL},
      url={https://arxiv.org/abs/1905.07830}, 
}

@misc{hendrycks2021measuringmassivemultitasklanguage,
      title={Measuring Massive Multitask Language Understanding}, 
      author={Dan Hendrycks and Collin Burns and Steven Basart and Andy Zou and Mantas Mazeika and Dawn Song and Jacob Steinhardt},
      year={2021},
      eprint={2009.03300},
      archivePrefix={arXiv},
      primaryClass={cs.CY},
      url={https://arxiv.org/abs/2009.03300}, 
}

@misc{west2023novacometopencommonsensefoundation,
      title={NovaCOMET: Open Commonsense Foundation Models with Symbolic Knowledge Distillation}, 
      author={Peter West and Ronan Le Bras and Taylor Sorensen and Bill Yuchen Lin and Liwei Jiang and Ximing Lu and Khyathi Chandu and Jack Hessel and Ashutosh Baheti and Chandra Bhagavatula and Yejin Choi},
      year={2023},
      eprint={2312.05979},
      archivePrefix={arXiv},
      primaryClass={cs.CL},
      url={https://arxiv.org/abs/2312.05979}, 
}

@article{BigelowPiantadosi2016_NumberGame,
  author  = {Bigelow, Eric and Piantadosi, Steven T.},
  title   = {A Large Dataset of Generalization Patterns in the Number Game},
  journal = {Journal of Open Psychology Data},
  year    = {2016},
  volume  = {4},
  number  = {1},
  pages   = {e4},
  doi     = {10.5334/jopd.19},
  url     = {https://openpsychologydata.metajnl.com/articles/10.5334/jopd.19/},
  note    = {Published 2016-03-18; accessed 2025-09-21}
}

@misc{lin2022truthfulqameasuringmodelsmimic,
      title={TruthfulQA: Measuring How Models Mimic Human Falsehoods}, 
      author={Stephanie Lin and Jacob Hilton and Owain Evans},
      year={2022},
      eprint={2109.07958},
      archivePrefix={arXiv},
      primaryClass={cs.CL},
      url={https://arxiv.org/abs/2109.07958}, 
}

@misc{EVS_WVS_Joint_2017_2022_v5_2024,
  author    = {{EVS/WVS}},
  title     = {European Values Study and World Values Survey: Joint EVS/WVS 2017–2022 Dataset},
  year      = {2024},
  version   = {5.0.0},
  publisher = {GESIS Data Archive, Cologne},
  doi       = {10.4232/1.14320},
  url       = {https://www.gesis.org/en/european-values-study/data-and-documentation/joint-evs/wvs-2017-2022-dataset},
  note      = {Identical version also via WVS site with DOI 10.14281/18241.26; accessed 2025-09-21}
}

@article{ai2:winogrande,
author = {Sakaguchi, Keisuke and Bras, Ronan Le and Bhagavatula, Chandra and Choi, Yejin},
title = {WinoGrande: an adversarial winograd schema challenge at scale},
year = {2021},
issue_date = {September 2021},
publisher = {Association for Computing Machinery},
address = {New York, NY, USA},
volume = {64},
number = {9},
issn = {0001-0782},
url = {https://doi.org/10.1145/3474381},
doi = {10.1145/3474381},
journal = {Commun. ACM},
month = aug,
pages = {99–106},
numpages = {8}
}

@article{zhang2024forcing,
  title={Forcing Diffuse Distributions out of Language Models},
  author={Zhang, Yiming and Schwarzschild, Avi and Carlini, Nicholas and Kolter, Zico and Ippolito, Daphne},
  journal={arXiv preprint arXiv:2404.10859},
  year={2024}
}

@article{li2024predicting,
  title={Predicting vs. acting: A trade-off between world modeling \& agent modeling},
  author={Li, Margaret and Shi, Weijia and Pagnoni, Artidoro and West, Peter and Holtzman, Ari},
  journal={arXiv preprint arXiv:2407.02446},
  year={2024}
}

@article{shumailov2023curse,
  title={The curse of recursion: Training on generated data makes models forget},
  author={Shumailov, Ilia and Shumaylov, Zakhar and Zhao, Yiren and Gal, Yarin and Papernot, Nicolas and Anderson, Ross},
  journal={arXiv preprint arXiv:2305.17493},
  year={2023}
}

@article{murthy2024one,
  title={One fish, two fish, but not the whole sea: Alignment reduces language models' conceptual diversity},
  author={Murthy, Sonia K and Ullman, Tomer and Hu, Jennifer},
  journal={arXiv preprint arXiv:2411.04427},
  year={2024}
}

@article{kirk2024benefits,
  title={The benefits, risks and bounds of personalizing the alignment of large language models to individuals},
  author={Kirk, Hannah Rose and Vidgen, Bertie and R{\"o}ttger, Paul and Hale, Scott A},
  journal={Nature Machine Intelligence},
  volume={6},
  number={4},
  pages={383--392},
  year={2024},
  publisher={Nature Publishing Group UK London}
}

@inproceedings{
yang2024model,
title={Model Autophagy Analysis to Explicate Self-consumption within Human-{AI} Interactions},
author={Shu Yang and Muhammad Asif Ali and Lu Yu and Lijie Hu and Di Wang},
booktitle={First Conference on Language Modeling},
year={2024},
url={https://openreview.net/forum?id=FX4fUThO9H}
}

@article{dohmatob2024strong,
  title={Strong model collapse},
  author={Dohmatob, Elvis and Feng, Yunzhen and Subramonian, Arjun and Kempe, Julia},
  journal={arXiv preprint arXiv:2410.04840},
  year={2024}
}

@inproceedings{anderson2024homogenization,
  title={Homogenization effects of large language models on human creative ideation},
  author={Anderson, Barrett R and Shah, Jash Hemant and Kreminski, Max},
  booktitle={Proceedings of the 16th conference on creativity \& cognition},
  pages={413--425},
  year={2024}
}

@article{kirk2023understanding,
  title={Understanding the effects of rlhf on llm generalisation and diversity},
  author={Kirk, Robert and Mediratta, Ishita and Nalmpantis, Christoforos and Luketina, Jelena and Hambro, Eric and Grefenstette, Edward and Raileanu, Roberta},
  journal={arXiv preprint arXiv:2310.06452},
  year={2023}
}

@article{kapania2024simulacrum,
  title={'Simulacrum of Stories': Examining Large Language Models as Qualitative Research Participants},
  author={Kapania, Shivani and Agnew, William and Eslami, Motahhare and Heidari, Hoda and Fox, Sarah},
  journal={arXiv preprint arXiv:2409.19430},
  year={2024}
}

@misc{zhang2025noveltybenchevaluatinglanguagemodels,
      title={NoveltyBench: Evaluating Language Models for Humanlike Diversity}, 
      author={Yiming Zhang and Harshita Diddee and Susan Holm and Hanchen Liu and Xinyue Liu and Vinay Samuel and Barry Wang and Daphne Ippolito},
      year={2025},
      eprint={2504.05228},
      archivePrefix={arXiv},
      primaryClass={cs.CL},
      url={https://arxiv.org/abs/2504.05228}, 
}

@misc{chen2024diversitysyntheticdataimpact,
      title={On the Diversity of Synthetic Data and its Impact on Training Large Language Models}, 
      author={Hao Chen and Abdul Waheed and Xiang Li and Yidong Wang and Jindong Wang and Bhiksha Raj and Marah I. Abdin},
      year={2024},
      eprint={2410.15226},
      archivePrefix={arXiv},
      primaryClass={cs.CL},
      url={https://arxiv.org/abs/2410.15226}, 
}

@article{ouyang2022training,
  title={Training language models to follow instructions with human feedback},
  author={Ouyang, Long and Wu, Jeffrey and Jiang, Xu and Almeida, Diogo and Wainwright, Carroll and Mishkin, Pamela and Zhang, Chong and Agarwal, Sandhini and Slama, Katarina and Ray, Alex and others},
  journal={Advances in neural information processing systems},
  volume={35},
  pages={27730--27744},
  year={2022}
}

@misc{alamdari2024consideratepathwaypluralisticalignment,
      title={Being Considerate as a Pathway Towards Pluralistic Alignment for Agentic AI}, 
      author={Parand A. Alamdari and Toryn Q. Klassen and Rodrigo Toro Icarte and Sheila A. McIlraith},
      year={2024},
      eprint={2411.10613},
      archivePrefix={arXiv},
      primaryClass={cs.AI},
      url={https://arxiv.org/abs/2411.10613}, 
}

@misc{castricato2024personareproducibletestbedpluralistic,
      title={PERSONA: A Reproducible Testbed for Pluralistic Alignment}, 
      author={Louis Castricato and Nathan Lile and Rafael Rafailov and Jan-Philipp Fränken and Chelsea Finn},
      year={2024},
      eprint={2407.17387},
      archivePrefix={arXiv},
      primaryClass={cs.CL},
      url={https://arxiv.org/abs/2407.17387}, 
}

@misc{jiang2023evaluatinginducingpersonalitypretrained,
      title={Evaluating and Inducing Personality in Pre-trained Language Models}, 
      author={Guangyuan Jiang and Manjie Xu and Song-Chun Zhu and Wenjuan Han and Chi Zhang and Yixin Zhu},
      year={2023},
      eprint={2206.07550},
      archivePrefix={arXiv},
      primaryClass={cs.CL},
      url={https://arxiv.org/abs/2206.07550}, 
}

@misc{chung2025modifyinglargelanguagemodel,
      title={Modifying Large Language Model Post-Training for Diverse Creative Writing}, 
      author={John Joon Young Chung and Vishakh Padmakumar and Melissa Roemmele and Yuqian Sun and Max Kreminski},
      year={2025},
      eprint={2503.17126},
      archivePrefix={arXiv},
      primaryClass={cs.CL},
      url={https://arxiv.org/abs/2503.17126}, 
}

@misc{west2025basemodelsbeataligned,
      title={Base Models Beat Aligned Models at Randomness and Creativity}, 
      author={Peter West and Christopher Potts},
      year={2025},
      eprint={2505.00047},
      archivePrefix={arXiv},
      primaryClass={cs.CL},
      url={https://arxiv.org/abs/2505.00047}, 
}

@article{grattafiori2024llama,
  title={The llama 3 herd of models},
  author={Grattafiori, Aaron and Dubey, Abhimanyu and Jauhri, Abhinav and Pandey, Abhinav and Kadian, Abhishek and Al-Dahle, Ahmad and Letman, Aiesha and Mathur, Akhil and Schelten, Alan and Vaughan, Alex and others},
  journal={arXiv preprint arXiv:2407.21783},
  year={2024}
}

@inproceedings{gu-etal-2025-olmes,
    title = "{OLMES}: A Standard for Language Model Evaluations",
    author = "Gu, Yuling  and
      Tafjord, Oyvind  and
      Kuehl, Bailey  and
      Haddad, Dany  and
      Dodge, Jesse  and
      Hajishirzi, Hannaneh",
    editor = "Chiruzzo, Luis  and
      Ritter, Alan  and
      Wang, Lu",
    booktitle = "Findings of the Association for Computational Linguistics: NAACL 2025",
    month = apr,
    year = "2025",
    address = "Albuquerque, New Mexico",
    publisher = "Association for Computational Linguistics",
    url = "https://aclanthology.org/2025.findings-naacl.282/",
    doi = "10.18653/v1/2025.findings-naacl.282",
    pages = "5005--5033",
    ISBN = "979-8-89176-195-7"
}

@inproceedings{
lambert2025tulu,
title={Tulu 3: Pushing Frontiers in Open Language Model Post-Training},
author={Nathan Lambert and Jacob Morrison and Valentina Pyatkin and Shengyi Huang and Hamish Ivison and Faeze Brahman and Lester James Validad Miranda and Alisa Liu and Nouha Dziri and Xinxi Lyu and Yuling Gu and Saumya Malik and Victoria Graf and Jena D. Hwang and Jiangjiang Yang and Ronan Le Bras and Oyvind Tafjord and Christopher Wilhelm and Luca Soldaini and Noah A. Smith and Yizhong Wang and Pradeep Dasigi and Hannaneh Hajishirzi},
booktitle={Second Conference on Language Modeling},
year={2025},
url={https://openreview.net/forum?id=i1uGbfHHpH}
}

@inproceedings{
dubois2024lengthcontrolled,
title={Length-Controlled AlpacaEval: A Simple Debiasing of Automatic Evaluators},
author={Yann Dubois and Percy Liang and Tatsunori Hashimoto},
booktitle={First Conference on Language Modeling},
year={2024},
url={https://openreview.net/forum?id=CybBmzWBX0}
}

@inproceedings{suzgun-etal-2023-challenging,
    title = "Challenging {BIG}-Bench Tasks and Whether Chain-of-Thought Can Solve Them",
    author = {Suzgun, Mirac  and
      Scales, Nathan  and
      Sch{\"a}rli, Nathanael  and
      Gehrmann, Sebastian  and
      Tay, Yi  and
      Chung, Hyung Won  and
      Chowdhery, Aakanksha  and
      Le, Quoc  and
      Chi, Ed  and
      Zhou, Denny  and
      Wei, Jason},
    editor = "Rogers, Anna  and
      Boyd-Graber, Jordan  and
      Okazaki, Naoaki",
    booktitle = "Findings of the Association for Computational Linguistics: ACL 2023",
    month = jul,
    year = "2023",
    address = "Toronto, Canada",
    publisher = "Association for Computational Linguistics",
    url = "https://aclanthology.org/2023.findings-acl.824/",
    doi = "10.18653/v1/2023.findings-acl.824",
    pages = "13003--13051",
}

@inproceedings{
rein2024gpqa,
title={{GPQA}: A Graduate-Level Google-Proof Q\&A Benchmark},
author={David Rein and Betty Li Hou and Asa Cooper Stickland and Jackson Petty and Richard Yuanzhe Pang and Julien Dirani and Julian Michael and Samuel R. Bowman},
booktitle={First Conference on Language Modeling},
year={2024},
url={https://openreview.net/forum?id=Ti67584b98}
}

@inproceedings{
zeng2024evaluating,
title={Evaluating Large Language Models at Evaluating Instruction Following},
author={Zhiyuan Zeng and Jiatong Yu and Tianyu Gao and Yu Meng and Tanya Goyal and Danqi Chen},
booktitle={The Twelfth International Conference on Learning Representations},
year={2024},
url={https://openreview.net/forum?id=tr0KidwPLc}
}

@inproceedings{
wang2024mmlupro,
title={{MMLU}-Pro: A More Robust and Challenging Multi-Task Language Understanding Benchmark},
author={Yubo Wang and Xueguang Ma and Ge Zhang and Yuansheng Ni and Abhranil Chandra and Shiguang Guo and Weiming Ren and Aaran Arulraj and Xuan He and Ziyan Jiang and Tianle Li and Max Ku and Kai Wang and Alex Zhuang and Rongqi Fan and Xiang Yue and Wenhu Chen},
booktitle={The Thirty-eight Conference on Neural Information Processing Systems Datasets and Benchmarks Track},
year={2024},
url={https://openreview.net/forum?id=y10DM6R2r3}
}

@inproceedings{lin-etal-2022-truthfulqa,
    title = "{T}ruthful{QA}: Measuring How Models Mimic Human Falsehoods",
    author = "Lin, Stephanie  and
      Hilton, Jacob  and
      Evans, Owain",
    editor = "Muresan, Smaranda  and
      Nakov, Preslav  and
      Villavicencio, Aline",
    booktitle = "Proceedings of the 60th Annual Meeting of the Association for Computational Linguistics (Volume 1: Long Papers)",
    month = may,
    year = "2022",
    address = "Dublin, Ireland",
    publisher = "Association for Computational Linguistics",
    url = "https://aclanthology.org/2022.acl-long.229/",
    doi = "10.18653/v1/2022.acl-long.229",
    pages = "3214--3252",
}

@misc{hospedales2020metalearningneuralnetworkssurvey,
      title={Meta-Learning in Neural Networks: A Survey}, 
      author={Timothy Hospedales and Antreas Antoniou and Paul Micaelli and Amos Storkey},
      year={2020},
      eprint={2004.05439},
      archivePrefix={arXiv},
      primaryClass={cs.LG},
      url={https://arxiv.org/abs/2004.05439}, 
}

@misc{gemmateam2025gemma3technicalreport,
      title={Gemma 3 Technical Report}, 
      author={Gemma Team and Aishwarya Kamath and Johan Ferret and Shreya Pathak and Nino Vieillard and Ramona Merhej and Sarah Perrin and Tatiana Matejovicova and Alexandre Ramé and Morgane Rivière and Louis Rouillard and Thomas Mesnard and Geoffrey Cideron and Jean-bastien Grill and Sabela Ramos and Edouard Yvinec and Michelle Casbon and Etienne Pot and Ivo Penchev and Gaël Liu and Francesco Visin and Kathleen Kenealy and Lucas Beyer and Xiaohai Zhai and Anton Tsitsulin and Robert Busa-Fekete and Alex Feng and Noveen Sachdeva and Benjamin Coleman and Yi Gao and Basil Mustafa and Iain Barr and Emilio Parisotto and David Tian and Matan Eyal and Colin Cherry and Jan-Thorsten Peter and Danila Sinopalnikov and Surya Bhupatiraju and Rishabh Agarwal and Mehran Kazemi and Dan Malkin and Ravin Kumar and David Vilar and Idan Brusilovsky and Jiaming Luo and Andreas Steiner and Abe Friesen and Abhanshu Sharma and Abheesht Sharma and Adi Mayrav Gilady and Adrian Goedeckemeyer and Alaa Saade and Alex Feng and Alexander Kolesnikov and Alexei Bendebury and Alvin Abdagic and Amit Vadi and András György and André Susano Pinto and Anil Das and Ankur Bapna and Antoine Miech and Antoine Yang and Antonia Paterson and Ashish Shenoy and Ayan Chakrabarti and Bilal Piot and Bo Wu and Bobak Shahriari and Bryce Petrini and Charlie Chen and Charline Le Lan and Christopher A. Choquette-Choo and CJ Carey and Cormac Brick and Daniel Deutsch and Danielle Eisenbud and Dee Cattle and Derek Cheng and Dimitris Paparas and Divyashree Shivakumar Sreepathihalli and Doug Reid and Dustin Tran and Dustin Zelle and Eric Noland and Erwin Huizenga and Eugene Kharitonov and Frederick Liu and Gagik Amirkhanyan and Glenn Cameron and Hadi Hashemi and Hanna Klimczak-Plucińska and Harman Singh and Harsh Mehta and Harshal Tushar Lehri and Hussein Hazimeh and Ian Ballantyne and Idan Szpektor and Ivan Nardini and Jean Pouget-Abadie and Jetha Chan and Joe Stanton and John Wieting and Jonathan Lai and Jordi Orbay and Joseph Fernandez and Josh Newlan and Ju-yeong Ji and Jyotinder Singh and Kat Black and Kathy Yu and Kevin Hui and Kiran Vodrahalli and Klaus Greff and Linhai Qiu and Marcella Valentine and Marina Coelho and Marvin Ritter and Matt Hoffman and Matthew Watson and Mayank Chaturvedi and Michael Moynihan and Min Ma and Nabila Babar and Natasha Noy and Nathan Byrd and Nick Roy and Nikola Momchev and Nilay Chauhan and Noveen Sachdeva and Oskar Bunyan and Pankil Botarda and Paul Caron and Paul Kishan Rubenstein and Phil Culliton and Philipp Schmid and Pier Giuseppe Sessa and Pingmei Xu and Piotr Stanczyk and Pouya Tafti and Rakesh Shivanna and Renjie Wu and Renke Pan and Reza Rokni and Rob Willoughby and Rohith Vallu and Ryan Mullins and Sammy Jerome and Sara Smoot and Sertan Girgin and Shariq Iqbal and Shashir Reddy and Shruti Sheth and Siim Põder and Sijal Bhatnagar and Sindhu Raghuram Panyam and Sivan Eiger and Susan Zhang and Tianqi Liu and Trevor Yacovone and Tyler Liechty and Uday Kalra and Utku Evci and Vedant Misra and Vincent Roseberry and Vlad Feinberg and Vlad Kolesnikov and Woohyun Han and Woosuk Kwon and Xi Chen and Yinlam Chow and Yuvein Zhu and Zichuan Wei and Zoltan Egyed and Victor Cotruta and Minh Giang and Phoebe Kirk and Anand Rao and Kat Black and Nabila Babar and Jessica Lo and Erica Moreira and Luiz Gustavo Martins and Omar Sanseviero and Lucas Gonzalez and Zach Gleicher and Tris Warkentin and Vahab Mirrokni and Evan Senter and Eli Collins and Joelle Barral and Zoubin Ghahramani and Raia Hadsell and Yossi Matias and D. Sculley and Slav Petrov and Noah Fiedel and Noam Shazeer and Oriol Vinyals and Jeff Dean and Demis Hassabis and Koray Kavukcuoglu and Clement Farabet and Elena Buchatskaya and Jean-Baptiste Alayrac and Rohan Anil and Dmitry and Lepikhin and Sebastian Borgeaud and Olivier Bachem and Armand Joulin and Alek Andreev and Cassidy Hardin and Robert Dadashi and Léonard Hussenot},
      year={2025},
      shortauthor={Gemma Team},
      eprint={2503.19786},
      archivePrefix={arXiv},
      primaryClass={cs.CL},
      url={https://arxiv.org/abs/2503.19786}, 
}

@misc{yang2025qwen3technicalreport,
      title={Qwen3 Technical Report}, 
      author={An Yang and Anfeng Li and Baosong Yang and Beichen Zhang and Binyuan Hui and Bo Zheng and Bowen Yu and Chang Gao and Chengen Huang and Chenxu Lv and Chujie Zheng and Dayiheng Liu and Fan Zhou and Fei Huang and Feng Hu and Hao Ge and Haoran Wei and Huan Lin and Jialong Tang and Jian Yang and Jianhong Tu and Jianwei Zhang and Jianxin Yang and Jiaxi Yang and Jing Zhou and Jingren Zhou and Junyang Lin and Kai Dang and Keqin Bao and Kexin Yang and Le Yu and Lianghao Deng and Mei Li and Mingfeng Xue and Mingze Li and Pei Zhang and Peng Wang and Qin Zhu and Rui Men and Ruize Gao and Shixuan Liu and Shuang Luo and Tianhao Li and Tianyi Tang and Wenbiao Yin and Xingzhang Ren and Xinyu Wang and Xinyu Zhang and Xuancheng Ren and Yang Fan and Yang Su and Yichang Zhang and Yinger Zhang and Yu Wan and Yuqiong Liu and Zekun Wang and Zeyu Cui and Zhenru Zhang and Zhipeng Zhou and Zihan Qiu},
      year={2025},
      eprint={2505.09388},
      archivePrefix={arXiv},
      primaryClass={cs.CL},
      url={https://arxiv.org/abs/2505.09388}, 
}

@phdthesis{Tenenbaum1999ConceptLearning,
  author  = {Joshua Tenenbaum},
  title   = {A {Bayesian} Framework for Concept Learning},
  school  = {Massachusetts Institute of Technology},
  year    = {1999},
  address = {Cambridge, MA}
}

@misc{RT_1970_2024_Community_2024,
  author       = {{u/Business-Platform301}},
  title        = {Rotten Tomatoes Movies 1970–2024},
  year         = {2024},
  howpublished = {\url{https://drive.google.com/file/d/12IpMErb4j83h5gGTdTpv0WZOf5ceY7b3/view}},
  note         = {Archive: \texttt{rotten\_tomatoes\_data\_1970\_2024.zip}. Provenance: r/datasets thread \url{https://www.reddit.com/r/datasets/comments/1ecj6m2/dataset_for_rotten_tomatoes_movies_1970_2024/}. Accessed 2025-09-23.}
}

@misc{tie2025surveyposttraininglargelanguage,
      title={A Survey on Post-training of Large Language Models}, 
      author={Guiyao Tie and Zeli Zhao and Dingjie Song and Fuyang Wei and Rong Zhou and Yurou Dai and Wen Yin and Zhejian Yang and Jiangyue Yan and Yao Su and Zhenhan Dai and Yifeng Xie and Yihan Cao and Lichao Sun and Pan Zhou and Lifang He and Hechang Chen and Yu Zhang and Qingsong Wen and Tianming Liu and Neil Zhenqiang Gong and Jiliang Tang and Caiming Xiong and Heng Ji and Philip S. Yu and Jianfeng Gao},
      year={2025},
      eprint={2503.06072},
      archivePrefix={arXiv},
      primaryClass={cs.CL},
      url={https://arxiv.org/abs/2503.06072}, 
}

@misc{wang2025reinforcementlearningreasoninglarge,
      title={Reinforcement Learning for Reasoning in Large Language Models with One Training Example}, 
      author={Yiping Wang and Qing Yang and Zhiyuan Zeng and Liliang Ren and Liyuan Liu and Baolin Peng and Hao Cheng and Xuehai He and Kuan Wang and Jianfeng Gao and Weizhu Chen and Shuohang Wang and Simon Shaolei Du and Yelong Shen},
      year={2025},
      eprint={2504.20571},
      archivePrefix={arXiv},
      primaryClass={cs.LG},
      url={https://arxiv.org/abs/2504.20571}, 
}

@misc{rafailov2024directpreferenceoptimizationlanguage,
      title={Direct Preference Optimization: Your Language Model is Secretly a Reward Model}, 
      author={Rafael Rafailov and Archit Sharma and Eric Mitchell and Stefano Ermon and Christopher D. Manning and Chelsea Finn},
      year={2024},
      eprint={2305.18290},
      archivePrefix={arXiv},
      primaryClass={cs.LG},
      url={https://arxiv.org/abs/2305.18290}, 
}

@misc{lee2024aligningthousandspreferencesmessage,
      title={Aligning to Thousands of Preferences via System Message Generalization}, 
      author={Seongyun Lee and Sue Hyun Park and Seungone Kim and Minjoon Seo},
      year={2024},
      eprint={2405.17977},
      archivePrefix={arXiv},
      primaryClass={cs.CL},
      url={https://arxiv.org/abs/2405.17977}, 
}

@inproceedings{basile-etal-2021-need,
    title = "We Need to Consider Disagreement in Evaluation",
    author = "Basile, Valerio  and
      Fell, Michael  and
      Fornaciari, Tommaso  and
      Hovy, Dirk  and
      Paun, Silviu  and
      Plank, Barbara  and
      Poesio, Massimo  and
      Uma, Alexandra",
    editor = "Church, Kenneth  and
      Liberman, Mark  and
      Kordoni, Valia",
    booktitle = "Proceedings of the 1st Workshop on Benchmarking: Past, Present and Future",
    month = aug,
    year = "2021",
    address = "Online",
    publisher = "Association for Computational Linguistics",
    url = "https://aclanthology.org/2021.bppf-1.3/",
    doi = "10.18653/v1/2021.bppf-1.3",
    pages = "15--21"
}

@misc{min2022metaicllearninglearncontext,
      title={MetaICL: Learning to Learn In Context}, 
      author={Sewon Min and Mike Lewis and Luke Zettlemoyer and Hannaneh Hajishirzi},
      year={2022},
      eprint={2110.15943},
      archivePrefix={arXiv},
      primaryClass={cs.CL},
      url={https://arxiv.org/abs/2110.15943}, 
}

@misc{ji2021earlystoppedneuralnetworksconsistent,
      title={Early-stopped neural networks are consistent}, 
      author={Ziwei Ji and Justin D. Li and Matus Telgarsky},
      year={2021},
      eprint={2106.05932},
      archivePrefix={arXiv},
      primaryClass={cs.LG},
      url={https://arxiv.org/abs/2106.05932}, 
}

@misc{dong2024surveyincontextlearning,
      title={A Survey on In-context Learning}, 
      author={Qingxiu Dong and Lei Li and Damai Dai and Ce Zheng and Jingyuan Ma and Rui Li and Heming Xia and Jingjing Xu and Zhiyong Wu and Tianyu Liu and Baobao Chang and Xu Sun and Lei Li and Zhifang Sui},
      year={2024},
      eprint={2301.00234},
      archivePrefix={arXiv},
      primaryClass={cs.CL},
      url={https://arxiv.org/abs/2301.00234}, 
}

@misc{wei2022finetunedlanguagemodelszeroshot,
      title={Finetuned Language Models Are Zero-Shot Learners}, 
      author={Jason Wei and Maarten Bosma and Vincent Y. Zhao and Kelvin Guu and Adams Wei Yu and Brian Lester and Nan Du and Andrew M. Dai and Quoc V. Le},
      year={2022},
      eprint={2109.01652},
      archivePrefix={arXiv},
      primaryClass={cs.CL},
      url={https://arxiv.org/abs/2109.01652}, 
}

@misc{sanh2022multitaskpromptedtrainingenables,
      title={Multitask Prompted Training Enables Zero-Shot Task Generalization}, 
      author={Victor Sanh and Albert Webson and Colin Raffel and Stephen H. Bach and Lintang Sutawika and Zaid Alyafeai and Antoine Chaffin and Arnaud Stiegler and Teven Le Scao and Arun Raja and Manan Dey and M Saiful Bari and Canwen Xu and Urmish Thakker and Shanya Sharma Sharma and Eliza Szczechla and Taewoon Kim and Gunjan Chhablani and Nihal Nayak and Debajyoti Datta and Jonathan Chang and Mike Tian-Jian Jiang and Han Wang and Matteo Manica and Sheng Shen and Zheng Xin Yong and Harshit Pandey and Rachel Bawden and Thomas Wang and Trishala Neeraj and Jos Rozen and Abheesht Sharma and Andrea Santilli and Thibault Fevry and Jason Alan Fries and Ryan Teehan and Tali Bers and Stella Biderman and Leo Gao and Thomas Wolf and Alexander M. Rush},
      year={2022},
      eprint={2110.08207},
      archivePrefix={arXiv},
      primaryClass={cs.LG},
      url={https://arxiv.org/abs/2110.08207}, 
}

@misc{wilson2022bayesiandeeplearningprobabilistic,
      title={Bayesian Deep Learning and a Probabilistic Perspective of Generalization}, 
      author={Andrew Gordon Wilson and Pavel Izmailov},
      year={2022},
      eprint={2002.08791},
      archivePrefix={arXiv},
      primaryClass={cs.LG},
      url={https://arxiv.org/abs/2002.08791}, 
}

@incollection{bayesianstatisticalanalysis,
title = {Bayesian Statistical Analysis},
editor = {Penelope Peterson and Eva Baker and Barry McGaw},
booktitle = {International Encyclopedia of Education (Third Edition)},
publisher = {Elsevier},
edition = {Third Edition},
address = {Oxford},
pages = {37-45},
year = {2010},
isbn = {978-0-08-044894-7},
doi = {https://doi.org/10.1016/B978-0-08-044894-7.01308-7},
url = {https://www.sciencedirect.com/science/article/pii/B9780080448947013087},
author = {G. Kokolakis}
}

@misc{lin2023unlockingspellbasellms,
      title={The Unlocking Spell on Base LLMs: Rethinking Alignment via In-Context Learning}, 
      author={Bill Yuchen Lin and Abhilasha Ravichander and Ximing Lu and Nouha Dziri and Melanie Sclar and Khyathi Chandu and Chandra Bhagavatula and Yejin Choi},
      year={2023},
      eprint={2312.01552},
      archivePrefix={arXiv},
      primaryClass={cs.CL},
      url={https://arxiv.org/abs/2312.01552}, 
}

@misc{rajbhandari2021zeroinfinitybreakinggpumemory,
      title={ZeRO-Infinity: Breaking the GPU Memory Wall for Extreme Scale Deep Learning}, 
      author={Samyam Rajbhandari and Olatunji Ruwase and Jeff Rasley and Shaden Smith and Yuxiong He},
      year={2021},
      eprint={2104.07857},
      archivePrefix={arXiv},
      primaryClass={cs.DC},
      url={https://arxiv.org/abs/2104.07857}, 
}

@inproceedings{asai-etal-2024-buffet,
    title = "{BUFFET}: Benchmarking Large Language Models for Few-shot Cross-lingual Transfer",
    author = "Asai, Akari  and
      Kudugunta, Sneha  and
      Yu, Xinyan  and
      Blevins, Terra  and
      Gonen, Hila  and
      Reid, Machel  and
      Tsvetkov, Yulia  and
      Ruder, Sebastian  and
      Hajishirzi, Hannaneh",
    editor = "Duh, Kevin  and
      Gomez, Helena  and
      Bethard, Steven",
    booktitle = "Proceedings of the 2024 Conference of the North American Chapter of the Association for Computational Linguistics: Human Language Technologies (Volume 1: Long Papers)",
    month = jun,
    year = "2024",
    address = "Mexico City, Mexico",
    publisher = "Association for Computational Linguistics",
    url = "https://aclanthology.org/2024.naacl-long.100/",
    doi = "10.18653/v1/2024.naacl-long.100",
    pages = "1771--1800"
}

@misc{wolf2020huggingfacestransformersstateoftheartnatural,
      title={HuggingFace's Transformers: State-of-the-art Natural Language Processing}, 
      author={Thomas Wolf and Lysandre Debut and Victor Sanh and Julien Chaumond and Clement Delangue and Anthony Moi and Pierric Cistac and Tim Rault and Rémi Louf and Morgan Funtowicz and Joe Davison and Sam Shleifer and Patrick von Platen and Clara Ma and Yacine Jernite and Julien Plu and Canwen Xu and Teven Le Scao and Sylvain Gugger and Mariama Drame and Quentin Lhoest and Alexander M. Rush},
      year={2020},
      eprint={1910.03771},
      archivePrefix={arXiv},
      primaryClass={cs.CL},
      url={https://arxiv.org/abs/1910.03771}, 
}

@misc{miehling2025evaluatingpromptsteerabilitylarge,
      title={Evaluating the Prompt Steerability of Large Language Models}, 
      author={Erik Miehling and Michael Desmond and Karthikeyan Natesan Ramamurthy and Elizabeth M. Daly and Pierre Dognin and Jesus Rios and Djallel Bouneffouf and Miao Liu},
      year={2025},
      eprint={2411.12405},
      archivePrefix={arXiv},
      primaryClass={cs.CL},
      url={https://arxiv.org/abs/2411.12405}, 
}

@misc{sorensen2025valueprofilesencodinghuman,
      title={Value Profiles for Encoding Human Variation}, 
      author={Taylor Sorensen and Pushkar Mishra and Roma Patel and Michael Henry Tessler and Michiel Bakker and Georgina Evans and Iason Gabriel and Noah Goodman and Verena Rieser},
      year={2025},
      eprint={2503.15484},
      archivePrefix={arXiv},
      primaryClass={cs.CL},
      url={https://arxiv.org/abs/2503.15484}, 
}

@misc{lanchantin2025diversepreferenceoptimization,
      title={Diverse Preference Optimization}, 
      author={Jack Lanchantin and Angelica Chen and Shehzaad Dhuliawala and Ping Yu and Jason Weston and Sainbayar Sukhbaatar and Ilia Kulikov},
      year={2025},
      eprint={2501.18101},
      archivePrefix={arXiv},
      primaryClass={cs.CL},
      url={https://arxiv.org/abs/2501.18101}, 
}

@misc{li2025jointlyreinforcingdiversityquality,
      title={Jointly Reinforcing Diversity and Quality in Language Model Generations}, 
      author={Tianjian Li and Yiming Zhang and Ping Yu and Swarnadeep Saha and Daniel Khashabi and Jason Weston and Jack Lanchantin and Tianlu Wang},
      year={2025},
      eprint={2509.02534},
      archivePrefix={arXiv},
      primaryClass={cs.CL},
      url={https://arxiv.org/abs/2509.02534}, 
}

@misc{openai2024gpt4technicalreport,
      title={GPT-4 Technical Report}, 
      author={OpenAI and Josh Achiam and Steven Adler and Sandhini Agarwal and Lama Ahmad and Ilge Akkaya and Florencia Leoni Aleman and Diogo Almeida and Janko Altenschmidt and Sam Altman and Shyamal Anadkat and Red Avila and Igor Babuschkin and Suchir Balaji and Valerie Balcom and Paul Baltescu and Haiming Bao and Mohammad Bavarian and Jeff Belgum and Irwan Bello and Jake Berdine and Gabriel Bernadett-Shapiro and Christopher Berner and Lenny Bogdonoff and Oleg Boiko and Madelaine Boyd and Anna-Luisa Brakman and Greg Brockman and Tim Brooks and Miles Brundage and Kevin Button and Trevor Cai and Rosie Campbell and Andrew Cann and Brittany Carey and Chelsea Carlson and Rory Carmichael and Brooke Chan and Che Chang and Fotis Chantzis and Derek Chen and Sully Chen and Ruby Chen and Jason Chen and Mark Chen and Ben Chess and Chester Cho and Casey Chu and Hyung Won Chung and Dave Cummings and Jeremiah Currier and Yunxing Dai and Cory Decareaux and Thomas Degry and Noah Deutsch and Damien Deville and Arka Dhar and David Dohan and Steve Dowling and Sheila Dunning and Adrien Ecoffet and Atty Eleti and Tyna Eloundou and David Farhi and Liam Fedus and Niko Felix and Simón Posada Fishman and Juston Forte and Isabella Fulford and Leo Gao and Elie Georges and Christian Gibson and Vik Goel and Tarun Gogineni and Gabriel Goh and Rapha Gontijo-Lopes and Jonathan Gordon and Morgan Grafstein and Scott Gray and Ryan Greene and Joshua Gross and Shixiang Shane Gu and Yufei Guo and Chris Hallacy and Jesse Han and Jeff Harris and Yuchen He and Mike Heaton and Johannes Heidecke and Chris Hesse and Alan Hickey and Wade Hickey and Peter Hoeschele and Brandon Houghton and Kenny Hsu and Shengli Hu and Xin Hu and Joost Huizinga and Shantanu Jain and Shawn Jain and Joanne Jang and Angela Jiang and Roger Jiang and Haozhun Jin and Denny Jin and Shino Jomoto and Billie Jonn and Heewoo Jun and Tomer Kaftan and Łukasz Kaiser and Ali Kamali and Ingmar Kanitscheider and Nitish Shirish Keskar and Tabarak Khan and Logan Kilpatrick and Jong Wook Kim and Christina Kim and Yongjik Kim and Jan Hendrik Kirchner and Jamie Kiros and Matt Knight and Daniel Kokotajlo and Łukasz Kondraciuk and Andrew Kondrich and Aris Konstantinidis and Kyle Kosic and Gretchen Krueger and Vishal Kuo and Michael Lampe and Ikai Lan and Teddy Lee and Jan Leike and Jade Leung and Daniel Levy and Chak Ming Li and Rachel Lim and Molly Lin and Stephanie Lin and Mateusz Litwin and Theresa Lopez and Ryan Lowe and Patricia Lue and Anna Makanju and Kim Malfacini and Sam Manning and Todor Markov and Yaniv Markovski and Bianca Martin and Katie Mayer and Andrew Mayne and Bob McGrew and Scott Mayer McKinney and Christine McLeavey and Paul McMillan and Jake McNeil and David Medina and Aalok Mehta and Jacob Menick and Luke Metz and Andrey Mishchenko and Pamela Mishkin and Vinnie Monaco and Evan Morikawa and Daniel Mossing and Tong Mu and Mira Murati and Oleg Murk and David Mély and Ashvin Nair and Reiichiro Nakano and Rajeev Nayak and Arvind Neelakantan and Richard Ngo and Hyeonwoo Noh and Long Ouyang and Cullen O'Keefe and Jakub Pachocki and Alex Paino and Joe Palermo and Ashley Pantuliano and Giambattista Parascandolo and Joel Parish and Emy Parparita and Alex Passos and Mikhail Pavlov and Andrew Peng and Adam Perelman and Filipe de Avila Belbute Peres and Michael Petrov and Henrique Ponde de Oliveira Pinto and Michael and Pokorny and Michelle Pokrass and Vitchyr H. Pong and Tolly Powell and Alethea Power and Boris Power and Elizabeth Proehl and Raul Puri and Alec Radford and Jack Rae and Aditya Ramesh and Cameron Raymond and Francis Real and Kendra Rimbach and Carl Ross and Bob Rotsted and Henri Roussez and Nick Ryder and Mario Saltarelli and Ted Sanders and Shibani Santurkar and Girish Sastry and Heather Schmidt and David Schnurr and John Schulman and Daniel Selsam and Kyla Sheppard and Toki Sherbakov and Jessica Shieh and Sarah Shoker and Pranav Shyam and Szymon Sidor and Eric Sigler and Maddie Simens and Jordan Sitkin and Katarina Slama and Ian Sohl and Benjamin Sokolowsky and Yang Song and Natalie Staudacher and Felipe Petroski Such and Natalie Summers and Ilya Sutskever and Jie Tang and Nikolas Tezak and Madeleine B. Thompson and Phil Tillet and Amin Tootoonchian and Elizabeth Tseng and Preston Tuggle and Nick Turley and Jerry Tworek and Juan Felipe Cerón Uribe and Andrea Vallone and Arun Vijayvergiya and Chelsea Voss and Carroll Wainwright and Justin Jay Wang and Alvin Wang and Ben Wang and Jonathan Ward and Jason Wei and CJ Weinmann and Akila Welihinda and Peter Welinder and Jiayi Weng and Lilian Weng and Matt Wiethoff and Dave Willner and Clemens Winter and Samuel Wolrich and Hannah Wong and Lauren Workman and Sherwin Wu and Jeff Wu and Michael Wu and Kai Xiao and Tao Xu and Sarah Yoo and Kevin Yu and Qiming Yuan and Wojciech Zaremba and Rowan Zellers and Chong Zhang and Marvin Zhang and Shengjia Zhao and Tianhao Zheng and Juntang Zhuang and William Zhuk and Barret Zoph},
      year={2024},
      eprint={2303.08774},
      archivePrefix={arXiv},
      primaryClass={cs.CL},
      url={https://arxiv.org/abs/2303.08774}, 
}

@inproceedings{Gordon_2022, series={CHI ’22},
   title={Jury Learning: Integrating Dissenting Voices into Machine Learning Models},
   url={http://dx.doi.org/10.1145/3491102.3502004},
   DOI={10.1145/3491102.3502004},
   booktitle={CHI Conference on Human Factors in Computing Systems},
   publisher={ACM},
   author={Gordon, Mitchell L. and Lam, Michelle S. and Park, Joon Sung and Patel, Kayur and Hancock, Jeff and Hashimoto, Tatsunori and Bernstein, Michael S.},
   year={2022},
   month=apr, pages={1–19},
   collection={CHI ’22} }

@misc{min2022rethinkingroledemonstrationsmakes,
      title={Rethinking the Role of Demonstrations: What Makes In-Context Learning Work?}, 
      author={Sewon Min and Xinxi Lyu and Ari Holtzman and Mikel Artetxe and Mike Lewis and Hannaneh Hajishirzi and Luke Zettlemoyer},
      year={2022},
      eprint={2202.12837},
      archivePrefix={arXiv},
      primaryClass={cs.CL},
      url={https://arxiv.org/abs/2202.12837}, 
}

@misc{zhang2025instructiontuninglargelanguage,
      title={Instruction Tuning for Large Language Models: A Survey}, 
      author={Shengyu Zhang and Linfeng Dong and Xiaoya Li and Sen Zhang and Xiaofei Sun and Shuhe Wang and Jiwei Li and Runyi Hu and Tianwei Zhang and Fei Wu and Guoyin Wang},
      year={2025},
      eprint={2308.10792},
      archivePrefix={arXiv},
      primaryClass={cs.CL},
      url={https://arxiv.org/abs/2308.10792}, 
}

@misc{zhang2025verbalizedsamplingmitigatemode,
      title={Verbalized Sampling: How to Mitigate Mode Collapse and Unlock LLM Diversity}, 
      author={Jiayi Zhang and Simon Yu and Derek Chong and Anthony Sicilia and Michael R. Tomz and Christopher D. Manning and Weiyan Shi},
      year={2025},
      eprint={2510.01171},
      archivePrefix={arXiv},
      primaryClass={cs.CL},
      url={https://arxiv.org/abs/2510.01171}, 
}

@misc{tian2023justaskcalibrationstrategies,
      title={Just Ask for Calibration: Strategies for Eliciting Calibrated Confidence Scores from Language Models Fine-Tuned with Human Feedback}, 
      author={Katherine Tian and Eric Mitchell and Allan Zhou and Archit Sharma and Rafael Rafailov and Huaxiu Yao and Chelsea Finn and Christopher D. Manning},
      year={2023},
      eprint={2305.14975},
      archivePrefix={arXiv},
      primaryClass={cs.CL},
      url={https://arxiv.org/abs/2305.14975}, 
}

@misc{bhatia2025valuedriftstracingvalue,
      title={Value Drifts: Tracing Value Alignment During LLM Post-Training}, 
      author={Mehar Bhatia and Shravan Nayak and Gaurav Kamath and Marius Mosbach and Karolina Stańczak and Vered Shwartz and Siva Reddy},
      year={2025},
      eprint={2510.26707},
      archivePrefix={arXiv},
      primaryClass={cs.CL},
      url={https://arxiv.org/abs/2510.26707}, 
}
\bibliographystyle{iclr2026_conference}

\newpage

\appendix

\setcounter{figure}{0}
\setcounter{table}{0}
\renewcommand{\thefigure}{A\arabic{figure}}
\renewcommand{\thetable}{A\arabic{table}}

\section{Limitations}
We hope that \resource can serve as a useful resource for others to evaluate and train models that support better in-context steerability, valid output coverage, and distributional alignment. We also believe that \method serves as a useful step in improving these desiderata. However, our work has several limitations.

\textit{Experiments performed only on $\leq$14B parameter models.} While we have ensured that results generalize across 3 model families, all models tested are in the 8B--14B parameter range. We have no reason to believe that our findings will not scale to larger model sizes, but this remains to be empirically verified.

\textit{Not optimized for chat.} While most current post-training techniques optimize for (potentially multi-turn) chat, models trained with \method instead focus on the \texttt{description}/\texttt{input}/\texttt{output} framework. While it can be possible to elicit chat-style messages via few-shot examples (see App. \ref{app:diversityvalidity}) from \methodabb models, we would expect that instruct models would be better at outputting a single chat response that is preferred by humans. It may be possible to combine the desiderata with a chat-style model, but they may also be fundamentally in tension, requiring distinct models for diversity/coverage and for chat.$^5$

\textit{Additional work needed on safety guardrails.} Currently, models trained with \method always attempt to steer to the description and examples, regardless of content.  This is, of course, also true of pretrained models, which is one justification for why a model developer may choose to keep certain pretrained models with advanced capabilities unavailable to the public. All of our experiments are with models with public pretrained variants, and we do not believe releasing our \method models enable any fundamentally new capabilities over the pretrained variants, but rather increase alignment with the desiderata. However, if a pretrained model has potentially harmful or dangerous capabilities that a model developer wishes to restrict, \method would need to be modified to adhere to these restrictions. While it is easy to imagine potential extensions to e.g. refuse to produce an output that violates a policy, we leave such exploration to future work.

\section{Supplementary Figures}

\begin{figure}[ht!]
    \centering
    \small
    \includegraphics[width=\textwidth]{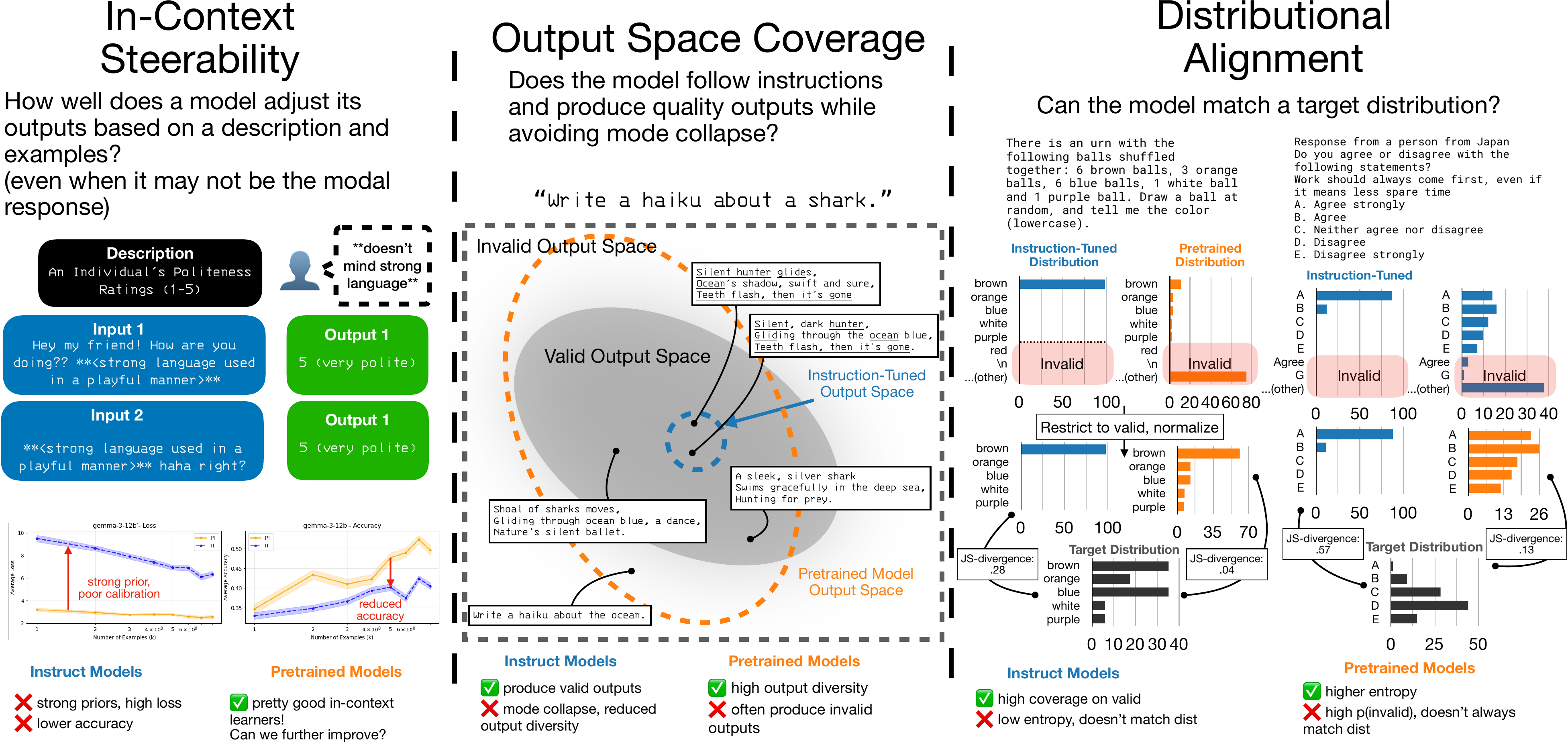}
    \caption{
    Three desiderata for conditional distributional modeling.
    Example outputs and data are drawn from \texttt{google/gemma-3-12b}.
    }
    \label{fig:desiderata}
\end{figure}

\begin{figure}[ht!]
    \centering \includegraphics[width=\textwidth]{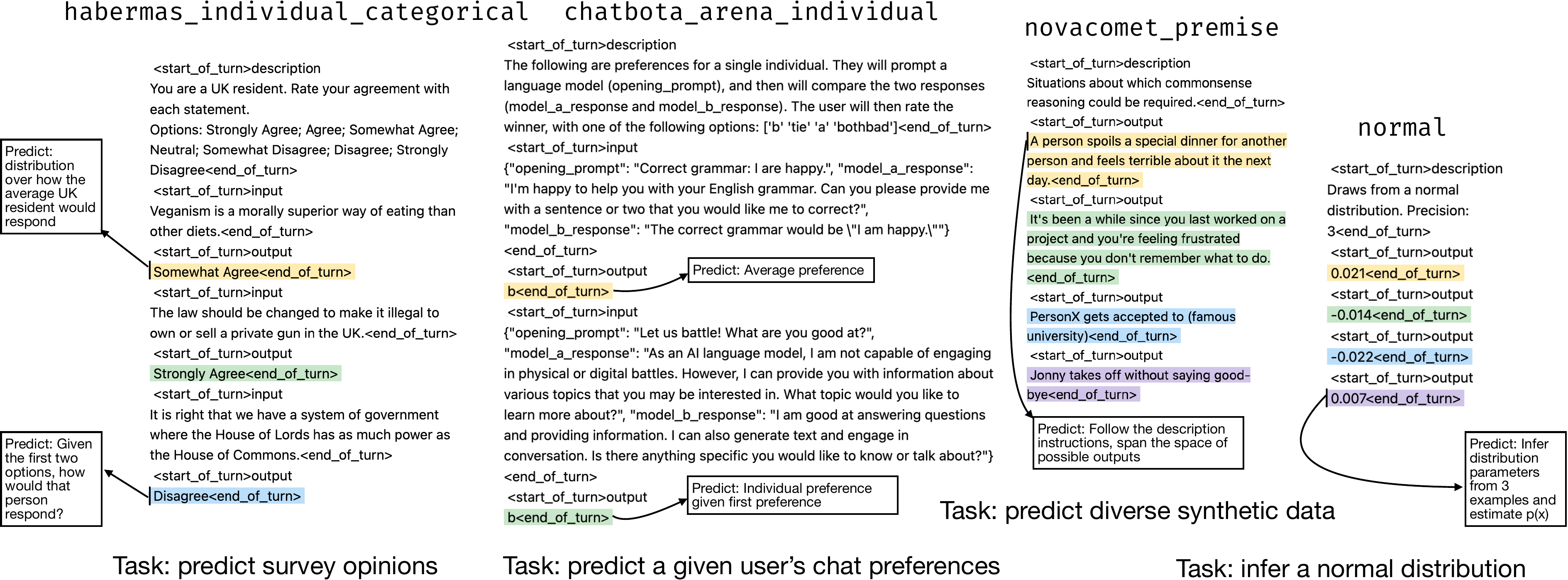}
    \caption{Example tasks from \resource in the format used for \method. In our method, we shuffle the data, put it into the above format, and finetune with cross-entropy loss only on the (highlighted) output tokens, including the terminal token.}
    \label{fig:spectrumsuite}
\end{figure}

\section{Frequently Asked Questions, Intuitions, and Hypotheses}

\textit{Q1: What unifies the three desiderata?} \\
A1: At first glance the desiderata may not seem very related, but they actually all have something in common - they all have to do with tasks where there is not a canonical, single correct answer. Rather, all three desiderata involve either matching or steering to a broad spectrum of potentially valid answers. This is in contrast with the majority of tasks on which we currently train and evaluate instruction-tuned LLMs.

\textit{Q2: Why does instruction-tuning post-training lead to spiky distributions and mode collapse?} \\
A2: We have two principal hypotheses for this: 1) the RL objective in RLHF/DPO/GRPO/etc. encourages the model to collapse its distribution to the highest reward output (c.f. \citealt{west2025basemodelsbeataligned}) and 2) most instruction-tuning training and evaluations focus on tasks with a single verifiable answer. While outside the scope of this work, comparing the desiderata at different stages of instruction-tuning (e.g., during and after Instruct-SFT, during and after RL) would help to elucidate this.\footnote{For an example of the checkpoint setup one might use, please refer to \citealt{bhatia2025valuedriftstracingvalue}, where they explore the effect of post-tuning on value drift.}

\textit{Q3: It makes sense that \method improves in-context steerability, as it maps easily onto the training data format. However, why does Spectrum Tuning improve diversity and distributional alignment/calibration?} \\
A3: While we hope to flesh out our understanding of this mechanism in future work, our best intuition is this - It largely has to do with the fact that 1) all training tasks involve interchangeable data and 2) we shuffle the data before training.
As a simple example, let us consider the \texttt{diffuse\_distribution} task: “Output a random country in Asia, chosen completely at random, without replacement.” In training, we collect a list of all countries in Asia, shuffle them, and finetune on them as outputs: e.g., “Brunei”, “Lebanon”, “Singapore”, “Laos”, “Vietnam”, ...
An instruction-tuned model will often exhibit mode collapse - outputting the same country each time. Meanwhile, a base model will often output a valid country, but is heavily affected by training data frequency / n-gram statistics. In contrast, in the limit, Spectrum Tuning encourages the model to actually instantiate a uniform distribution over all countries in Asia - increasing the diversity of outputs across many samples. For distributional alignment and calibration, it is a similar story - base models are heavily affected by things like n-gram statistics, instruct models have uncalibrated, spiky distributions. In contrast, Spectrum Tuning in the limit encourages the model to fit the actual described distribution, (partially) overcoming n-gram frequency.

\section{\resource Data Sources}
\label{app:datasources}

\subsection{Data Construction}
As \resource is the first-such large-scale resource of such subjective datasets requiring steering, it was necessarily constructed in a somewhat ad-hoc manner. However, here we provide some general principles for data that we attempted to source:
\begin{enumerate}
    \item Any NLP datasets with corresponding annotator IDs, allowing us to link multiple annotations to the same person. We especially sourced from datasets where variation is to be expected, as opposed to be eliminated.
    \item Datasets related to opinion modeling or computational democracy;
    \item Synthetically-generated NLP datasets;
    \item Lists of interchangeable things;
    \item Draws from random distributions;
    \item Tabular data.
\end{enumerate}

\subsection{Data Sources}

Below, we cite all data sources used in \resource. Additionally, we include any subtask names along with the number of sequences included in \resource. We release the processing code to go from raw data to our \texttt{description/input/output} in our github repo 
(\url{https://github.com/tsor13/spectrum}).

Note that many data sources have much more additional data that we could utilize (e.g., OpinionQA \citep{santurkar2023whose}, Polis \citep{CompDem_OpenData_2025}, synthetically generated random data). We generally restricted each data source to a maximum of 1-2k sequences to ensure training data diversity, and in all but a couple of cases with very few data instances (e.g. Diffuse Distributions; \citealt{zhang2024forcingdiffusedistributionslanguage}) additionally ensured that the same piece of data was not used in more than one sequence.

\subsection{Train Split}

\textbf{Ambient Ambiguity Detection} \citep{liu-etal-2023-afraid}
\begin{itemize}
\item \texttt{ambient\_ambiguity\_detection} (50 sequences)
\item \texttt{ambient\_annotation\_distributions} (50 sequences)
\item \texttt{ambient\_disambiguation} (50 sequences)
\item \texttt{ambient\_interpretation\_labels} (50 sequences)
\item \texttt{ambient\_linguist\_annotations} (54 sequences)
\item \texttt{ambient\_premise\_hypothesis} (50 sequences)
\end{itemize}

\textbf{Social Security Administration Baby Names} \citep{SSA_BabyNames_National_2025}
\begin{itemize}
\item \texttt{babynames} (500 sequences)
\end{itemize}

\textbf{Base-Refine Synthetic Data Generation} \citep{zhu2025bareleveragingbaselanguage}
\begin{itemize}
\item \texttt{bare\_enron} (55 sequences)
\item \texttt{bare\_gsm8k} (108 sequences)
\item \texttt{bare\_hotpot} (50 sequences)
\item \texttt{bare\_lcb} (136 sequences)
\item \texttt{bare\_newsgroups} (60 sequences)
\item \texttt{bare\_pubmed} (46 sequences)
\end{itemize}

\textbf{Draws from a binomial distribution (generated)} 
\begin{itemize}
\item \texttt{binomial} (500 sequences)
\end{itemize}

\textbf{Draws from a shuffled deck of cards (generated)} 
\begin{itemize}
\item \texttt{cards} (100 sequences)
\end{itemize}

\textbf{Draws from a categorical distribution (generated)}
\begin{itemize}
\item \texttt{categorical} (500 sequences)
\end{itemize}

\textbf{ChangeMyView Reddit} \citep{Kolyada2020WebisCMV20}
\begin{itemize}
\item \texttt{changemyview\_categories} (809 sequences)
\item \texttt{changemyview\_posts} (1159 sequences)
\end{itemize}

\textbf{Draws from a biased coin (generated)} 
\begin{itemize}
\item \texttt{coinflip} (1000 sequences)
\end{itemize}

\textbf{Collective Alignment Dataset} \citep{OpenAI_CA1_2025}
\begin{itemize}
\item \texttt{collective\_alignment\_individual} (993 sequences)
\end{itemize}

\textbf{Community Alignment Dataset} \citep{zhang2025cultivating}
\begin{itemize}
\item \texttt{community\_alignment\_individual\_preferences} (770 sequences)
\item \texttt{community\_alignment\_individual\_reply} (1031 sequences)
\item \texttt{community\_alignment\_initial\_prompt} (139 sequences)
\item \texttt{community\_alignment\_response} (941 sequences)
\end{itemize}

\textbf{DICES dataset} \citep{aroyo2023dicesdatasetdiversityconversational}
\begin{itemize}
\item \texttt{dices} (295 sequences)
\end{itemize}

\textbf{Diffuse Distributions} \citep{zhang2024forcingdiffusedistributionslanguage}
\begin{itemize}
\item \texttt{diffuse\_distribution} (270 sequences)
\end{itemize}

\textbf{Generative Social choice} \citep{fish2025generativesocialchoice}
\begin{itemize}
\item \texttt{generativesocialchoice\_freetext} (200 sequences)
\item \texttt{generativesocialchoice\_validation} (400 sequences)
\end{itemize}

\textbf{Draws from a geometric distribution (generated)} 
\begin{itemize}
\item \texttt{geometric} (500 sequences)
\end{itemize}

\textbf{Draws from a geometric beta distribution (generated)}
\begin{itemize}
\item \texttt{geometric\_beta} (500 sequences)
\end{itemize}

\textbf{Grade-school math problems (GSM8K)} \citep{cobbe2021gsm8k}
\begin{itemize}
\item \texttt{gsm8k\_answer\_from\_question} (50 sequences)
\item \texttt{gsm8k\_question} (50 sequences)
\item \texttt{gsm8k\_question\_answer} (50 sequences)
\item \texttt{gsm8k\_question\_from\_answer} (50 sequences)
\end{itemize}

\textbf{Haikus} \citep{Neiman2018_HaikuDL}
\begin{itemize}
\item \texttt{haikus} (600 sequences)
\end{itemize}

\textbf{Hatespeech annotations from diverse annotators} \citep{kumar2021designingtoxiccontentclassification}
\begin{itemize}
\item \texttt{hatespeech\_individual} (1000 sequences)
\end{itemize}

\textbf{Helpsteer2 Synthetic Chat Preferences} \citep{wang2024helpsteer2preferencecomplementingratingspreferences}
\begin{itemize}
\item \texttt{helpsteer} (320 sequences)
\end{itemize}

\textbf{Draws from a hypergeometric distribution, generated} \citep{wang2024helpsteer2preferencecomplementingratingspreferences}
\begin{itemize}
\item \texttt{hypergeometric} (500 sequences)
\end{itemize}

\textbf{IssueBench (measuring political leaning of LLMs)} \citep{röttger2025issuebenchmillionsrealisticprompts}
\begin{itemize}
\item \texttt{issuebench} (4 sequences)
\end{itemize}

\textbf{Jeopardy! questions and answers} \citep{trexmatt_jeopardy_csv_2014}
\begin{itemize}
\item \texttt{jeopardy\_answer\_prediction} (1000 sequences)
\item \texttt{jeopardy\_question\_generation} (1000 sequences)
\end{itemize}

\textbf{Sarcasm detection (multiple annotators)} \citep{jang-frassinelli-2024-generalizable}
\begin{itemize}
\item \texttt{lewidi\_csc\_sarcasm\_detection\_individual} (872 sequences)
\end{itemize}

\textbf{Irony detection (multiple annotators)} \citep{casola-etal-2024-multipico}
\begin{itemize}
\item \texttt{lewidi\_mp\_irony\_detection\_individual} (475 sequences)
\end{itemize}

\textbf{Paraphrase detection with rationales (multiple annotators)} \citep{lewidi}
\begin{itemize}
\item \texttt{lewidi\_par\_paraphrase\_detection\_individual} (80 sequences)
\item \texttt{lewidi\_par\_paraphrase\_detection\_individual\_categorical} (80 sequences)
\end{itemize}

\textbf{Entailment (multiple annotators)} \citep{weber-genzel-etal-2024-varierr}
\begin{itemize}
\item \texttt{lewidi\_varierrnli\_nli\_detection\_individual} (52 sequences)
\item \texttt{lewidi\_varierrnli\_nli\_detection\_individual\_categorical} (52 sequences)
\end{itemize}

\textbf{Draws from a multinomial distribution (generated)}
\begin{itemize}
\item \texttt{multinomial} (500 sequences)
\end{itemize}

\textbf{Draws from a negative binomial distribution (generated)}
\begin{itemize}
\item \texttt{negative\_binomial} (500 sequences)
\end{itemize}

\textbf{Netflix views and rating data} \citep{NetflixPrize_Kaggle_2009}
\begin{itemize}
\item \texttt{netflix\_individual\_ratings} (1000 sequences)
\item \texttt{netflix\_individual\_views} (2000 sequences)
\end{itemize}

\textbf{Draws from a normal distribution (generated)}
\begin{itemize}
\item \texttt{normal} (1000 sequences)
\end{itemize}

\textbf{OpinionQA: Large-scale opinion survey dataset} \citep{santurkar2023whose}
\begin{itemize}
\item \texttt{opinionqa\_individual} (3000 sequences)
\item \texttt{opinionqa\_questions} (15 sequences)
\end{itemize}

\textbf{Draws from a poisson distribution (generated)} 
\begin{itemize}
\item \texttt{poisson} (500 sequences)
\end{itemize}

\textbf{Polis OpenData: Votes from a digital town hall} \citep{CompDem_OpenData_2025}
\begin{itemize}
\item \texttt{polis\_comment} (336 sequences)
\item \texttt{polis\_vote} (7452 sequences)
\end{itemize}

\textbf{Popquorn: Annotator disagreement on 5 NLP tasks, with demographics} \citep{pei2023annotatordemographicsmattermeasuring}
\begin{itemize}
\item \texttt{popquorn\_individual} (400 sequences)
\item \texttt{popquorn\_og\_categorical} (80 sequences)
\end{itemize}

\textbf{Prism: World-wide, pluralistic chat preferences} \citep{kirk2024prismalignmentdatasetparticipatory}
\begin{itemize}
\item \texttt{prism\_individual\_preferences} (1333 sequences)
\item \texttt{prism\_prompts} (54 sequences)
\item \texttt{prism\_prompts\_individual} (1393 sequences)
\end{itemize}

\textbf{Titanic survival prediction: classic machine learning tabular dataset} \citep{mstz_titanic_2023}
\begin{itemize}
\item \texttt{titanic\_all\_variables} (14 sequences)
\item \texttt{titanic\_survival\_prediction} (14 sequences)
\end{itemize}

\textbf{Value Consistency: Multi-lingual value laden questions} \citep{moore2024largelanguagemodelsconsistent}
\begin{itemize}
\item \texttt{valueconsistency} (21 sequences)
\end{itemize}

\textbf{ValuePrism: datasets with moral judgments and relevant values, rights, and duties} \citep{Sorensen_2024}
\begin{itemize}
\item \texttt{valueprism\_misc} (400 sequences)
\item \texttt{valueprism\_situation} (105 sequences)
\item \texttt{valueprism\_vrd} (500 sequences)
\item \texttt{valueprism\_vrds\_noncontextual} (74 sequences)
\end{itemize}

\textbf{Draws from a zipfian distribution (generated)}
\begin{itemize}
\item \texttt{zipfian} (500 sequences)
\end{itemize}

\subsection{Test Split}

\textbf{ChatbotArena Individual Preferences} \citep{zheng2023judging}
\begin{itemize}
\item \texttt{chatbotarena\_assistant} (928 sequences)
\item \texttt{chatbotarena\_individual\_prefs} (1183 sequences)
\item \texttt{chatbotarena\_prompts} (1000 sequences)
\end{itemize}

\textbf{Tabular Chemistry Dataset} \citep{ramos2023bayesian}
\begin{itemize}
\item \texttt{chemistry\_esol} (310 sequences)
\item \texttt{chemistry\_oxidative} (102 sequences)
\end{itemize}

\textbf{Synthetic Flight Preferences} \citep{qiu2025bayesianteachingenablesprobabilistic}
\begin{itemize}
\item \texttt{flight} (200 sequences)
\end{itemize}

\textbf{GlobalOQA: Country-specific Value Surevy Distributions} \citep{durmus2023measuring}
\begin{itemize}
\item \texttt{globaloqa} (274 sequences)
\end{itemize}

\textbf{Habermas Dataset: AI Deliberation with UK residents} \citep{doi:10.1126/science.adq2852}
\begin{itemize}
\item \texttt{habermas\_individual} (1996 sequences)
\item \texttt{habermas\_individual\_categorical} (2000 sequences)
\item \texttt{habermas\_opinions} (199 sequences)
\item \texttt{habermas\_question} (43 sequences)
\end{itemize}

\textbf{NovaCOMET: Synthetic Commonsense Dataset} \citep{west2023novacometopencommonsensefoundation}
\begin{itemize}
\item \texttt{novacomet\_hypothesis} (170 sequences)
\item \texttt{novacomet\_premise} (68 sequences)
\end{itemize}

\textbf{NumberGame dataset: cognitive science dataset used to study human reasoning under uncertainty} \citep{BigelowPiantadosi2016_NumberGame}
\begin{itemize}
\item \texttt{numbergame\_individual} (606 sequences)
\item \texttt{numbergame\_perc} (182 sequences)
\end{itemize}

\textbf{World Values Survey, Wave 7: Global survey on human values} \citep{EVS_WVS_Joint_2017_2022_v5_2024}
\begin{itemize}
\item \texttt{wvs\_individual} (2000 sequences)
\end{itemize}

\subsection{Capability Split}
\label{app:capabilitydatasets}

\textbf{AI2 Reasoning Challenge} \citep{clark2018thinksolvedquestionanswering}
\begin{itemize}
\item \texttt{arc} (118 sequences)
\end{itemize}

\textbf{DROP: Reading Comprehension} \citep{dua2019dropreadingcomprehensionbenchmark}
\begin{itemize}
\item \texttt{drop} (943 sequences)
\end{itemize}

\textbf{GPQA: Google-Proof QA Benchmark} \citep{rein2023gpqagraduatelevelgoogleproofqa}
\begin{itemize}
\item \texttt{gpqa} (995 sequences)
\end{itemize}

\textbf{Hellaswag: commonsense benchmark} \citep{zellers2019hellaswagmachinereallyfinish}
\begin{itemize}
\item \texttt{hellaswag} (503 sequences)
\end{itemize}

\textbf{IMDB sentiment classification} \citep{maas-EtAl:2011:ACL-HLT2011}
\begin{itemize}
\item \texttt{imdb} (192 sequences)
\end{itemize}

\textbf{MMLU: Massive Multitask Language Understanding Benchmark} \citep{hendrycks2021measuringmassivemultitasklanguage}
\begin{itemize}
\item \texttt{mmlu} (1000 sequences)
\end{itemize}

\textbf{TruthfulQA: factual questions} \citep{lin2022truthfulqameasuringmodelsmimic}
\begin{itemize}
\item \texttt{truthful\_qa} (69 sequences)
\end{itemize}

\textbf{Winogrande: Commonsense sentence completion} \citep{ai2:winogrande}
\begin{itemize}
\item \texttt{winogrande} (127 sequences)
\end{itemize}

\section{Effect of Temperature on Diversity vs. Validity}
\label{app:temp}

Temperature  can have a major effect on the diversity vs. validity tradeoff when sampling from a model. In \S\ref{sec:spanningoutputspace}, we observed that, when sampling across three levels of prompting information and three model families, Spectrum tuning offered a pareto improvement on diversity vs. validity and overall improved yield. However, the question still remains - does Spectrum tuning still offer an improvement, even after sweeping temperature values?

\begin{figure}[ht!]
    \vspace{-10pt}
    \centering
    \includegraphics[width=\textwidth]{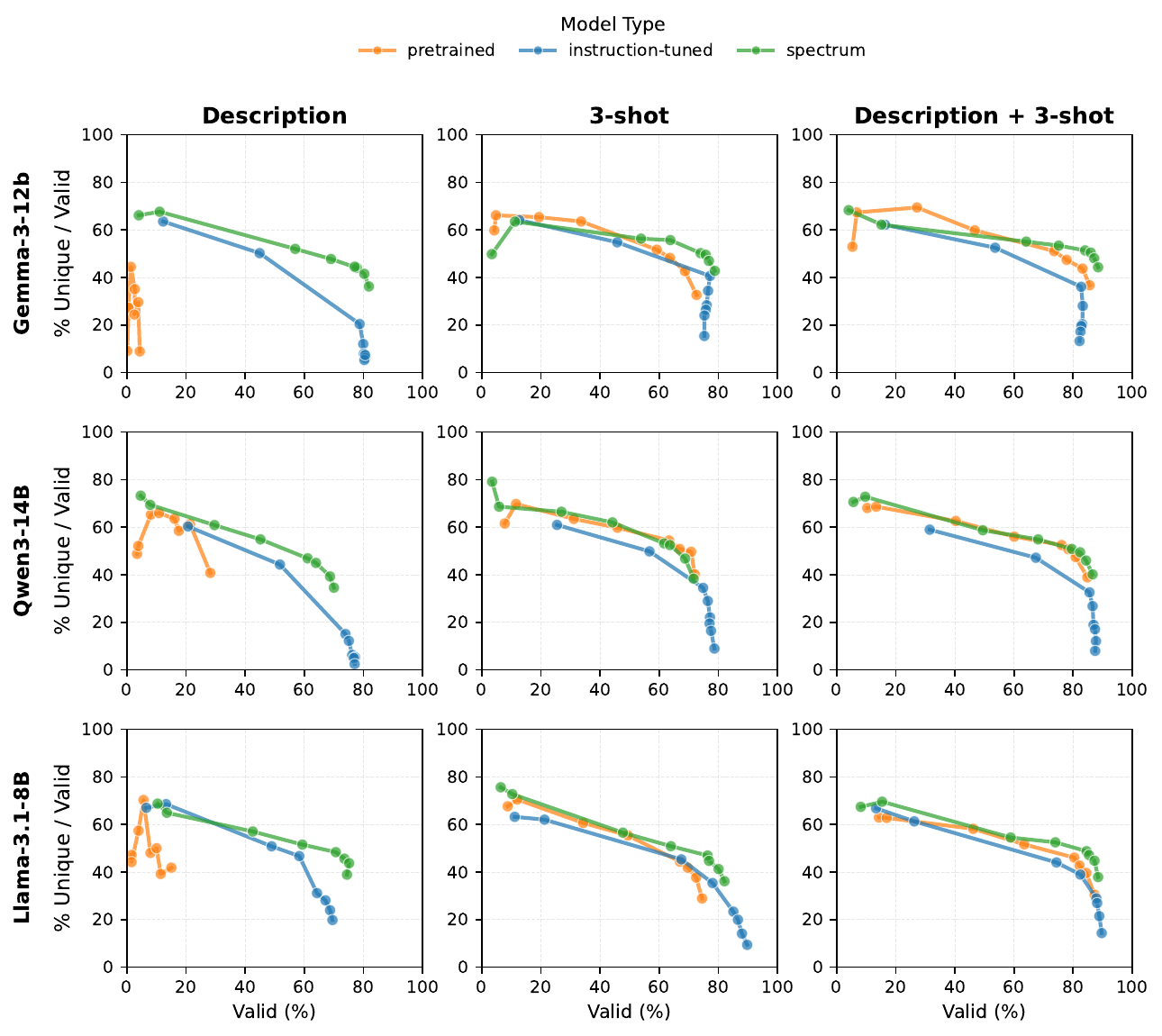}
    \caption{Effect of temperature on diversity and validity.
    Tested temperatures: \texttt{[10, 5, 2, 1.5, 1, 0.9, 0.7, 0.5]}. Lines are connected for temperature in ascending order, with the right-most endpoint being lowest temperature and the left-most endpoint being highest temperature.
    Spectrum Tuning generally offers a Pareto improvement, especially in the high validity region.}
    \label{fig:diverse_verify_temp}
\end{figure}

To answer this question, we evaluated the same models under the same setup, but sampled at various temperatures: \texttt{[10, 5, 2, 1.5, 1, 0.9, 0.7, 0.5]}. In Figure \ref{fig:diverse_verify_temp}, we plot diversity vs. validity for all three model families, prompting methods, and model types. We find that, in eight of nine settings, Spectrum Tuning expands the diversity / validity Pareto frontier, as compared to using instruction-tuned or pretrained models alone. In addition, Spectrum Tuning models typically expand the Pareto frontier in the high validity region, increasing diversity for a given validity. In line with the temperature=1 results, Spectrum Tuning's gains offer the largest improvement in the lowest information setting, when only a description of the task is provided.

\begin{figure}[ht!]
    \vspace{-10pt}
    \centering
    \includegraphics[width=\textwidth]{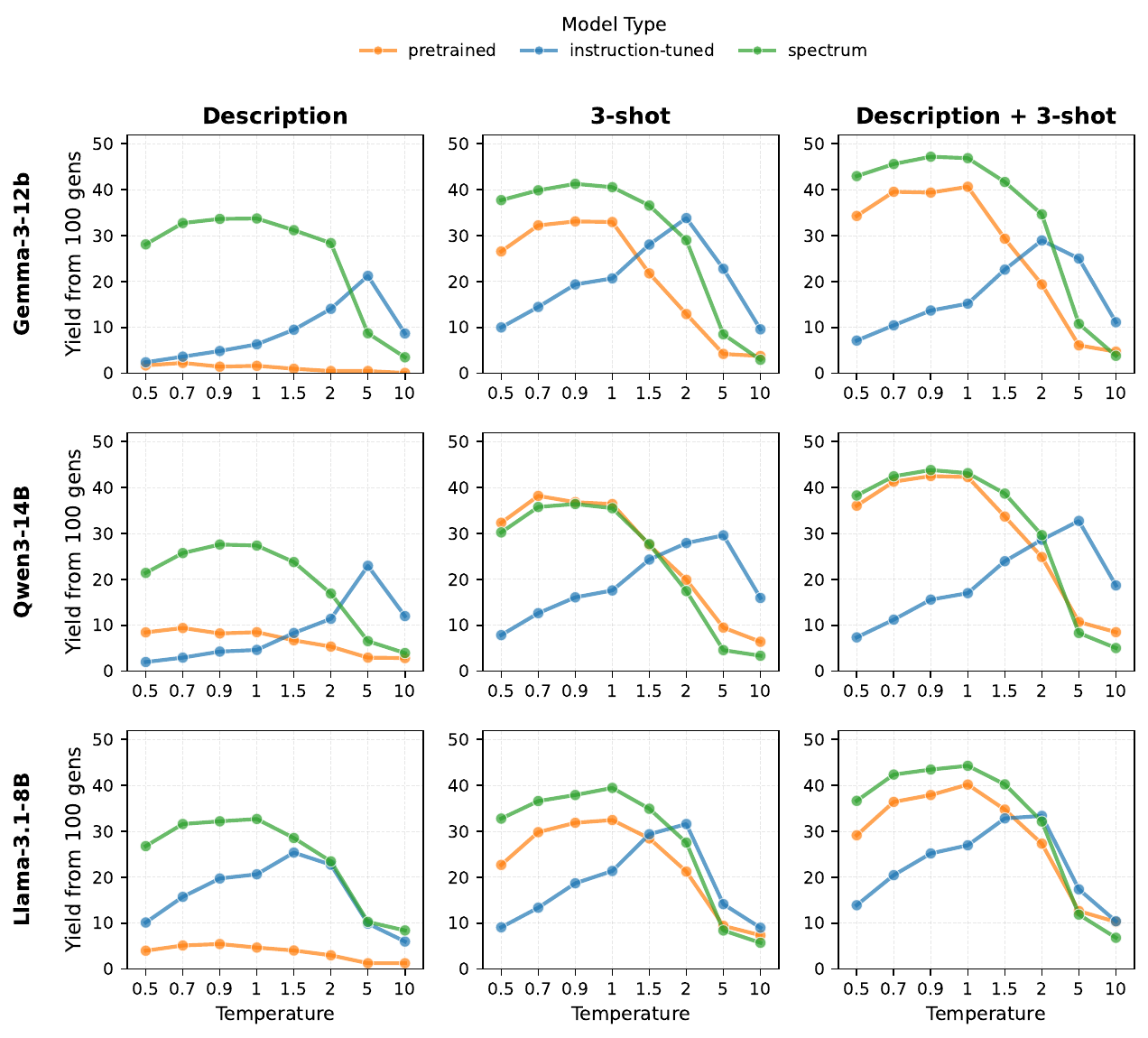}
    \caption{Effect of temperature on yield across each setting. When selecting the optimal temperature for each model, Spectrum Tuning offers the highest overall yield in 8/9 cases (all but Qwen3-14B / 3-shot). Spectrum Tuning also offers the highest yield in most temperature settings $T\leq2$.}
    \label{fig:diverse_verify_temp_yield}
\end{figure}

In Figure \ref{fig:diverse_verify_temp_yield}, we also plot the yield for each setting against the temperature. We find that in eight of nine cases, Spectrum Tuning offers the highest possible yield across all models and temperatures - implying that, even if when selecting the optimal temperature for each generation task, we would expect the highest number of distinct valid generations from the Spectrum-Tuned models.


Taken together, we find that the gains from Spectrum Tuning hold even when leaving temperature as a free variable.

\section{General Capability Performance}
\label{app:generalcap}

We test whether \method affects general model capabilities. While we do not necessarily expect our method to improve upon standard evaluations where there is a single correct answer, we want to understand if it degrades performance compared to pretrained models. We evaluate general knowledge capabilities with Big-Bench Hard (BBH, 3-shot, \citealt{suzgun-etal-2023-challenging}), GPQA (5-shot with chain of thought, \citealt{rein2024gpqa}), MMLU-Pro (5-shot with chain of thought, \citealt{wang2024mmlupro}), and TruthfulQA (6-shot, \citealt{lin-etal-2022-truthfulqa}); instruction following with IFEval \citep{zeng2024evaluating}; and chat ability with AlpacaEval v2 \citep{dubois2024lengthcontrolled}. We use the default Olmes hyperparameters for evaluating pretrained models, and Tulu-v3 hyperparameters and task descriptions for evaluating instruction-tuned models \citep{gu-etal-2025-olmes, lambert2025tulu}. In general, we find that models trained with \method perform similarly to the pretrained models, and in some cases exceed them; however, as expected, instruction-tuned models perform much better, particularly on instruction following and chat tasks.

\begin{table}
\centering
\tiny
\setlength{\tabcolsep}{6pt}
\renewcommand{\arraystretch}{1.15}
\adjustbox{width=\linewidth}{
\begin{tabular}{lccccccccc}
\toprule
  & \multicolumn{3}{c}{\textbf{gemma-3-12b}} & \multicolumn{3}{c}{\textbf{Qwen3-14B}} & \multicolumn{3}{c}{\textbf{Llama-3.1-8B}} \\
\cmidrule(lr){2-4} \cmidrule(lr){5-7} \cmidrule(lr){8-10}
\textbf{{Dataset}} & \textbf{\methodabb (ours)} & \textbf{PT} & \textbf{IT} & \textbf{\methodabb (ours)} & \textbf{PT} & \textbf{IT} & \textbf{\methodabb (ours)} & \textbf{PT} & \textbf{IT} \\
\midrule
AlpacaEval 2 & \underline{5.935} & 6.897 & 53.846 & \underline{30.421} & 33.541 & 63.123 & 3.642 & \underline{3.579} & 24.641 \\
BBH & 0.738 & \underline{0.727} & 0.821 & 0.786 & 0.789 & \underline{0.770} & 0.641 & \underline{0.631} & 0.722 \\
GPQA & 0.257 & \underline{0.250} & 0.377 & \underline{0.339} & 0.386 & 0.411 & 0.246 & \underline{0.208} & 0.315 \\
IFEval & \underline{0.407} & 0.436 & 0.806 & \underline{0.712} & 0.726 & 0.871 & 0.377 & \underline{0.296} & 0.793 \\
MMLU-Pro & 0.458 & \underline{0.448} & 0.592 & 0.584 & \underline{0.555} & 0.684 & \underline{0.358} & 0.360 & 0.481 \\
TruthfulQA & 0.516 & \underline{0.483} & 0.610 & \underline{0.498} & 0.529 & 0.553 & \underline{0.435} & 0.446 & 0.551 \\
\bottomrule
\end{tabular}
}
    \caption{General Capability Results. 
    \textit{Worst} performance is underlined.
    \method and pretrained models perform similarly.
    }
    \label{tab:capabilities}
\end{table}

\section{Training details}
\label{app:trainingdetails}
We lightly tuned hyperparameters by training the \texttt{gemma-3-12b} model on a subset of tasks from \resource-Train and tracking performance on held-out train tasks. We used the same hyperparameters for Llama and Qwen, performing no additional hyperparameter tuning.
Training for all models was done on four 80GB A100 GPUs using DeepSpeed Zero3 \citep{rajbhandari2021zeroinfinitybreakinggpumemory} and Hugging Face Transformers \citep{wolf2020huggingfacestransformersstateoftheartnatural}. Training took about 16 hours for the Llama models, 26 hours for the Gemma models, and 30 hours for the Qwen models.

Hyperparameters used:
\begin{itemize}
    \item \texttt{max\_length: 1024}
    \item \texttt{per\_device\_train\_batch\_size: 1}
    \item \texttt{gradient\_accumulation\_steps: 512}
    \item \texttt{learning\_rate: 3e-6}
    \item \texttt{learning\_rate\_scheduler: linear\_decay}
\end{itemize}

\section{Results with Updated Hyperparameters}
\label{app:betterhparams}
After running the main suite of experiments for the paper and experimenting with the models, we had reason to believe that our Spectrum-Tuned models, especially the Qwen and Llama models, were underfit. Note that, for the main set of experiments, we only lightly fit hyperparameters only on the Gemma models using a held-out subset of the train tasks as a validation set, and used the same hyperparameters for Qwen / Llama.

To further explore the effect of updating hyperparameters, we experimented with reducing the batch size in order to take more gradient updates. In the original hyperparameter mix, we use an effective batch size of 2048 (512 gradient steps $\times$ 1 train sequence per device $\times$ 4 GPUs). We halve the batch size three times, and report aggregate results in 
Table \ref{tab:hparamablate}.

\begin{table}[ht!]
\centering
\tiny
\setlength{\tabcolsep}{3pt}
\renewcommand{\arraystretch}{1.2}
\begin{tabular}{lccclcccccccc}
\multicolumn{4}{c}{} & \multicolumn{4}{c}{\textbf{ICL Steerability}} & \multicolumn{1}{c}{\textbf{Dist. Align.}} & \multicolumn{3}{c}{\textbf{Valid Output Coverage}} \\
\cmidrule(lr){1-4} \cmidrule(lr){5-8} \cmidrule(lr){9-9} \cmidrule{10-12}
\textbf{Effective} & & & & & \textbf{MC Loss} & \textbf{MC Acc} & \textbf{Free-text} & \textbf{Dist. Align.} & \textbf{Yield -} & \textbf{Yield -} & \textbf{Yield - 3-shot} \\
\textbf{Batch Size} & & & & & \textbf{(Norm.)} & \textbf{(Norm.)} & \textbf{Loss (Norm.)} & \textbf{JS-Div.} & \textbf{Description} & \textbf{3-shot} & \textbf{+ Description} \\
\midrule
\rowcolor{gray!20}
2048 (original hparam) &  &  &  & & \textbf{1.00} & 1.00 & \textbf{1.00} & .069 & 36.7 & 42.1 & 49.2 \\
1024 & &  &  & & \underline{1.02} & 1.02 & \textbf{1.00} & \underline{.065} & 43.5 & 44.8 & 51.1 \\
512 & &  &  & & 1.05 & \underline{1.06} & \textbf{1.00} & \textbf{.063} & \underline{44.8} & \textbf{45.9} & \underline{51.5} \\
256 & &  &  & & 1.09 & \textbf{1.07} & \underline{1.01} & \textbf{.063} & \textbf{45.9} & \underline{45.7} & \textbf{52.0} \\
\end{tabular}
\caption{Hyperparameter ablations, averaged across models and tasks. Shaded are default \method models. Best result bolded, second best underlined.
}
\label{tab:hparamablate}
\end{table}

We find that 1) decreasing the batch size results in a substantial jump in zero-shot yield, and slight improvements in few-shot yield and distributional alignment. Additionally, decreasing the batch size increases multiple choice accuracy, but at the cost of higher loss on multiple choice answers. All in all, we think that this illustrates that there are likely to be additional gains from further optimization, and that our initial hyperparameters were likely underfit.

\begin{figure}[ht!]
    \vspace{-10pt}
    \centering
    \includegraphics[width=\textwidth]{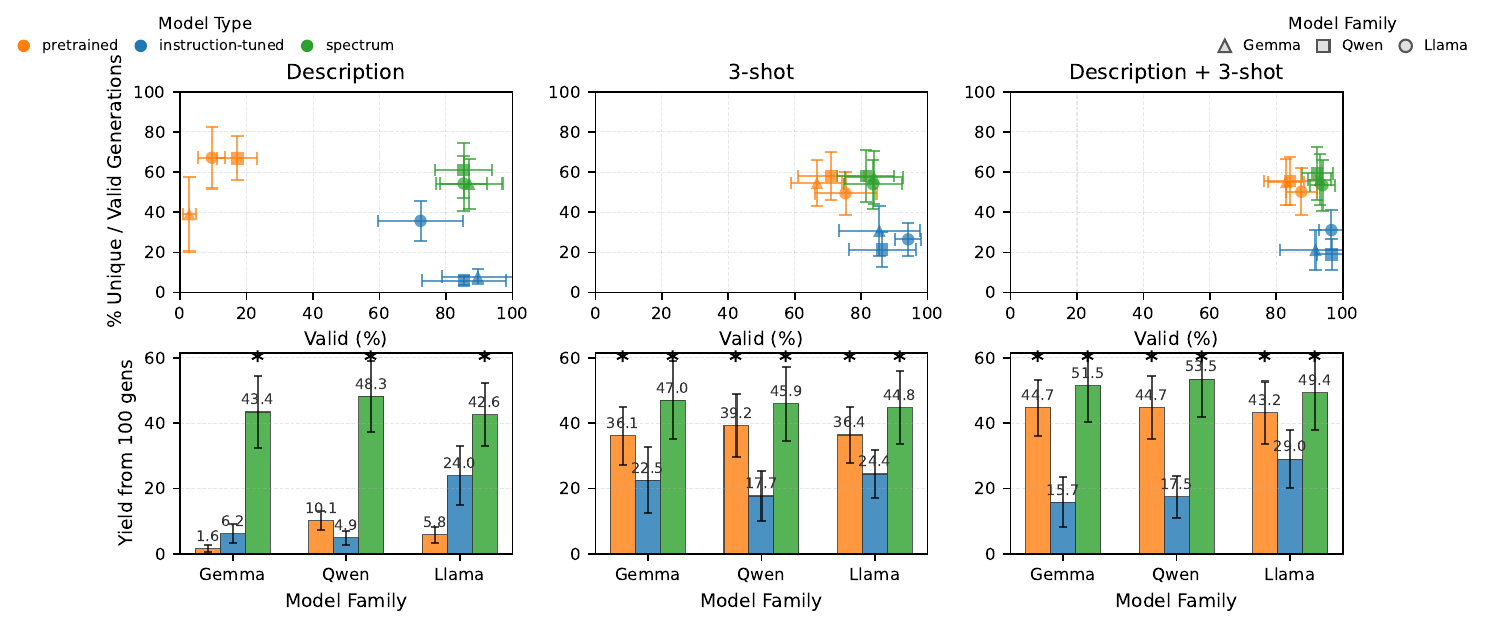}
    \caption{Diversity vs. validity on verifiable tasks for models trained with an effective batch size of 512.}
    \label{fig:diverse_verify_bar512}
\end{figure}

We think that the models trained with effective batch size 512 offer a good tradeoff between ICL steerability, distributional alignment, and valid output coverage, and report their full results in Tables \ref{tab:iclsteer512}-\ref{tab:distributional512} and Figure \ref{fig:diverse_verify_bar512}.

\begin{table}[ht!]
\centering
\tiny
\setlength{\tabcolsep}{6pt}
\renewcommand{\arraystretch}{1.15}
\begin{tabular}{ll*{9}{c}}
\toprule
 &  & \multicolumn{3}{c}{\textbf{gemma-3-12b}} & \multicolumn{3}{c}{\textbf{Qwen3-14B}} & \multicolumn{3}{c}{\textbf{Llama-3.1-8B}} \\
\cmidrule(lr){3-5} \cmidrule(lr){6-8} \cmidrule(lr){9-11}
\textbf{Dataset} & \textbf{Metric} & \textbf{ours} & \textbf{pt} & \textbf{it} & \textbf{ours} & \textbf{pt} & \textbf{it} & \textbf{ours} & \textbf{pt} & \textbf{it} \\
\midrule
\midrule
\multicolumn{11}{c}{\textbf{Multiple-Choice Datasets}} \\
\midrule
 &  & \multicolumn{3}{c}{\textbf{gemma-3-12b}} & \multicolumn{3}{c}{\textbf{Qwen3-14B}} & \multicolumn{3}{c}{\textbf{Llama-3.1-8B}} \\
\midrule
\rowcolor{gray!20} \textbf{habermas\_individual\_categorical} (max\_k=2, N=1000) & \textbf{Loss} & 3.53 & \textbf{2.50} & 10.5 & \textbf{2.01} & 2.62 & 9.10 & \textbf{2.58} & \textbf{2.58} & 2.74 \\
 & \textbf{Acc} & \textbf{24.0} & \textbf{24.4} & \textbf{22.4} & \textbf{24.9} & 20.3 & 22.0 & \textbf{23.2} & 20.2 & 19.0 \\
\rowcolor{gray!20} \textbf{wvs\_individual} (max\_k=21, N=1000) & \textbf{Loss} & \textbf{1.36} & 1.50 & 4.10 & \textbf{1.38} & 1.74 & 4.35 & \textbf{1.42} & 1.57 & 1.76 \\
 & \textbf{Acc} & \textbf{44.7} & 42.1 & 40.4 & \textbf{45.2} & 41.1 & 40.6 & \textbf{44.5} & 41.6 & 39.4 \\
\rowcolor{gray!20} \textbf{numbergame\_individual} (max\_k=25, N=592) & \textbf{Loss} & \textbf{.665} & .705 & 1.80 & \textbf{.617} & .697 & 1.28 & \textbf{.611} & .864 & .770 \\
 & \textbf{Acc} & \textbf{70.2} & 64.3 & 65.6 & \textbf{71.2} & 69.8 & \textbf{71.0} & \textbf{69.2} & 62.5 & 67.5 \\
\rowcolor{gray!20} \textbf{chatbotarena\_individual\_prefs} (max\_k=3, N=725) & \textbf{Loss} & \textbf{1.52} & 1.62 & 4.94 & \textbf{1.35} & 1.47 & 4.39 & \textbf{1.43} & 1.76 & 1.77 \\
 & \textbf{Acc} & \textbf{48.9} & 38.0 & 44.6 & \textbf{51.7} & \textbf{52.0} & 46.3 & \textbf{39.5} & 36.0 & \textbf{39.5} \\
\rowcolor{gray!20} \textbf{flight} (max\_k=9, N=200) & \textbf{Loss} & \textbf{1.11} & 1.32 & 4.06 & \textbf{1.09} & 1.29 & 2.92 & \textbf{1.09} & 1.45 & 1.41 \\
 & \textbf{Acc} & \textbf{41.0} & \textbf{41.2} & \textbf{40.6} & \textbf{43.1} & \textbf{43.7} & \textbf{40.8} & \textbf{40.9} & \textbf{42.0} & \textbf{40.2} \\
\midrule
\multicolumn{11}{c}{\textbf{Free-text Datasets}} \\
\midrule
 &  & \multicolumn{3}{c}{\textbf{gemma-3-12b}} & \multicolumn{3}{c}{\textbf{Qwen3-14B}} & \multicolumn{3}{c}{\textbf{Llama-3.1-8B}} \\
\midrule
\rowcolor{gray!20} \textbf{novacomet\_hypothesis} (max\_k=11, N=155) & \textbf{Loss} & \textbf{105} & \textbf{104} & 135 & \textbf{107} & \textbf{106} & 129 & 110 & \textbf{106} & 112 \\
\rowcolor{gray!20} \textbf{novacomet\_premise} (max\_k=55, N=51) & \textbf{Loss} & \textbf{27.7} & \textbf{28.0} & 35.5 & \textbf{27.7} & \textbf{27.5} & 38.0 & \textbf{27.9} & \textbf{27.7} & 28.6 \\
\rowcolor{gray!20} \textbf{habermas\_question} (max\_k=29, N=30) & \textbf{Loss} & \textbf{23.9} & \textbf{23.1} & 41.4 & \textbf{23.8} & \textbf{24.0} & 31.8 & \textbf{23.8} & \textbf{23.8} & 24.8 \\
\rowcolor{gray!20} \textbf{habermas\_opinions} (max\_k=2, N=186) & \textbf{Loss} & \textbf{927} & \textbf{928} & 1070 & \textbf{947} & \textbf{949} & 1070 & \textbf{944} & \textbf{944} & \textbf{991} \\
\rowcolor{gray!20} \textbf{habermas\_individual} (max\_k=2, N=1000) & \textbf{Loss} & \textbf{164} & \textbf{164} & 203 & \textbf{167} & \textbf{168} & 210 & \textbf{166} & \textbf{167} & 176 \\
\rowcolor{gray!20} \textbf{numbergame\_perc} (max\_k=24, N=182) & \textbf{Loss} & \textbf{4.26} & \textbf{4.22} & 6.68 & \textbf{4.13} & 4.24 & 5.61 & \textbf{4.31} & 4.43 & 4.41 \\
\rowcolor{gray!20} \textbf{globaloqa} (max\_k=8, N=231) & \textbf{Loss} & \textbf{14.2} & \textbf{14.4} & 21.5 & \textbf{14.0} & \textbf{14.4} & 20.9 & \textbf{14.5} & \textbf{14.7} & 15.6 \\
\rowcolor{gray!20} \textbf{chatbotarena\_prompts} (max\_k=3, N=988) & \textbf{Loss} & \textbf{69.8} & \textbf{69.4} & 117 & \textbf{67.9} & \textbf{68.2} & 97.8 & \textbf{72.0} & \textbf{72.0} & \textbf{77.6} \\
\rowcolor{gray!20} \textbf{chatbotarena\_assistant} (max\_k=5, N=716) & \textbf{Loss} & \textbf{127} & \textbf{125} & 259 & \textbf{124} & \textbf{124} & 169 & \textbf{136} & \textbf{133} & 149 \\
\rowcolor{gray!20} \textbf{chemistry\_esol} (max\_k=8, N=59) & \textbf{Loss} & \textbf{8.45} & \textbf{8.37} & 12.9 & \textbf{8.45} & \textbf{8.47} & 11.8 & \textbf{8.30} & \textbf{8.51} & \textbf{8.55} \\
\rowcolor{gray!20} \textbf{chemistry\_oxidative} (max\_k=9, N=101) & \textbf{Loss} & \textbf{7.57} & \textbf{7.58} & 11.6 & \textbf{7.57} & 7.84 & 10.2 & \textbf{7.68} & \textbf{7.72} & 7.84 \\
\bottomrule
\end{tabular}
\caption{In-context steerability results on models trained with an effective batch size of 512.}
\label{tab:iclsteer512}
\end{table}

\begin{table}[h!]
\centering
\tiny
\setlength{\tabcolsep}{6pt}
\renewcommand{\arraystretch}{1.15}
\begin{tabular}{l*{9}{c}}
\toprule
\textbf{Dataset} & \multicolumn{3}{c}{\textbf{gemma-3-12b}} & \multicolumn{3}{c}{\textbf{Qwen3-14B}} & \multicolumn{3}{c}{\textbf{Llama-3.1-8B}} \\
\cmidrule(lr){2-4} \cmidrule(lr){5-7} \cmidrule(lr){8-10}
 & \textbf{ours} & \textbf{pt} & \textbf{it} & \textbf{ours} & \textbf{pt} & \textbf{it} & \textbf{ours} & \textbf{pt} & \textbf{it} \\
\midrule
\textbf{habermas\_individual\_categorical} & 0.13 & \textbf{0.069} & 0.239 & \textbf{0.049} & 0.05 & 0.198 & 0.108 & 0.084 & \textbf{0.055} \\
\textbf{wvs\_individual} & \textbf{0.007} & 0.015 & 0.223 & \textbf{0.007} & 0.02 & 0.191 & \textbf{0.005} & 0.012 & 0.024 \\
\textbf{numbergame\_individual} & \textbf{0.019} & 0.029 & 0.163 & 0.037 & \textbf{0.026} & 0.108 & 0.027 & 0.024 & \textbf{0.017} \\
\textbf{chatbotarena\_individual\_prefs} & \textbf{0.02} & 0.041 & 0.194 & 0.056 & \textbf{0.046} & 0.189 & 0.062 & 0.075 & \textbf{0.049} \\
\textbf{flight} & \textbf{0.019} & 0.04 & 0.271 & 0.055 & \textbf{0.035} & 0.228 & \textbf{0.03} & 0.07 & 0.038 \\
\bottomrule
\end{tabular}
\caption{Calibration for models trained with an effective batch size of 512.}
\label{tab:calibration512}
\end{table}

\begin{table}[t]
\centering
\tiny
\setlength{\tabcolsep}{6pt}
\renewcommand{\arraystretch}{1.15}
\begin{tabular}{ll*{9}{c}}
\toprule
 &  & \multicolumn{3}{c}{\textbf{gemma-3-12b}} & \multicolumn{3}{c}{\textbf{Qwen3-14B}} & \multicolumn{3}{c}{\textbf{Llama-3.1-8B}} \\
\cmidrule(lr){3-5} \cmidrule(lr){6-8} \cmidrule(lr){9-11}
\textbf{Dataset} & \textbf{Metric} & \textbf{ours} & \textbf{pt} & \textbf{it} & \textbf{ours} & \textbf{pt} & \textbf{it} & \textbf{ours} & \textbf{pt} & \textbf{it} \\
\midrule
\rowcolor{gray!20} \textbf{mpi} & \textbf{JS-Div} & \textbf{.101} & .126 & .347 & \textbf{.107} & \textbf{.0928} & .405 & \textbf{.0489} & .0874 & .131 \\
\rowcolor{gray!20} \textbf{rotten\_tomatoes} & \textbf{JS-Div} & \textbf{.0227} & .0323 & .134 & .0341 & \textbf{.0283} & .122 & \textbf{.0245} & .0354 & .0859 \\
\rowcolor{gray!20} \textbf{nytimes} & \textbf{JS-Div} & \textbf{.0547} & .0628 & .328 & \textbf{.0453} & .0876 & .344 & \textbf{.0655} & \textbf{.0613} & .247 \\
\rowcolor{gray!20} \textbf{global\_oqa} & \textbf{JS-Div} & \textbf{.0678} & .0936 & .270 & \textbf{.0749} & .0878 & .274 & \textbf{.0828} & .108 & .163 \\
\rowcolor{gray!20} \textbf{urn} & \textbf{JS-Div} & \textbf{.0136} & .0713 & .185 & \textbf{.0186} & .0592 & .198 & \textbf{.0186} & .124 & .0865 \\
\rowcolor{gray!20} \textbf{habermas} & \textbf{JS-Div} & \textbf{.142} & \textbf{.147} & .436 & \textbf{.125} & \textbf{.127} & .434 & \textbf{.129} & .155 & .242 \\
\rowcolor{gray!20} \textbf{numbergame} & \textbf{JS-Div} & .0663 & \textbf{.0488} & .138 & \textbf{.0440} & \textbf{.0428} & .131 & \textbf{.0423} & .0600 & .0943 \\
\bottomrule
\end{tabular}
\caption{Distributional alignment for models trained with an effective batch size of 512.}
\label{tab:distributional512}
\end{table}

\clearpage

\section{Human Evaluation}
\label{app:human_annotation}
We conducted a large-scale human annotation study to evaluate the validity and quality of outputs from different model configurations. The study used a pairwise comparison design where annotators evaluated outputs from two models simultaneously for the same prompts. We recruited 245 U.S.-based English speaking annotators who had submitted at least 1000 prior tasks with an approval rating of at least 95\% through Prolific and collected a total of 2,400 annotations. Our task took about 30 minutes and we paid at least 7.5 USD for an average of at least 15 USD an hour.

Specifically, we sampled 100 prompts from two evaluation datasets, a curated prompt set and infinite-chats-eval, and collected human judgments for each.
Our experimental design compared three model configurations (baseline instruction-tuned, our approach, and pretrained) in both zero-shot and few-shot settings. Each unique combination of (prompt, model pair) was evaluated by two independent annotators, resulting in 200 annotation instances per model pair per dataset.

\begin{figure}[h]
    \centering
    \includegraphics[width=0.8\linewidth]{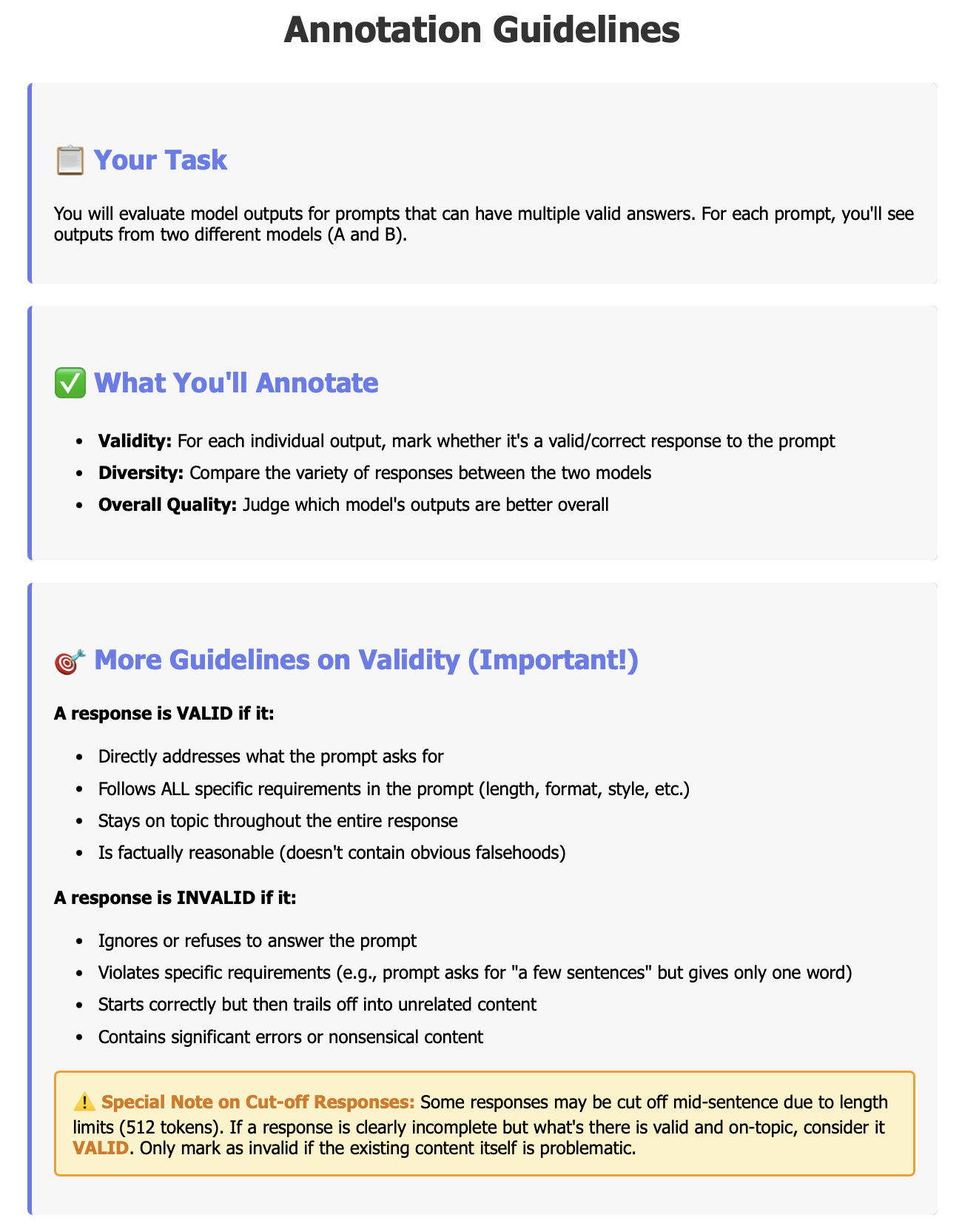}
    \caption{Guidelines for human annotation}
    \label{fig:guidelines}
\end{figure}

\begin{figure}[h]
    \centering
    \includegraphics[width=0.65\linewidth]{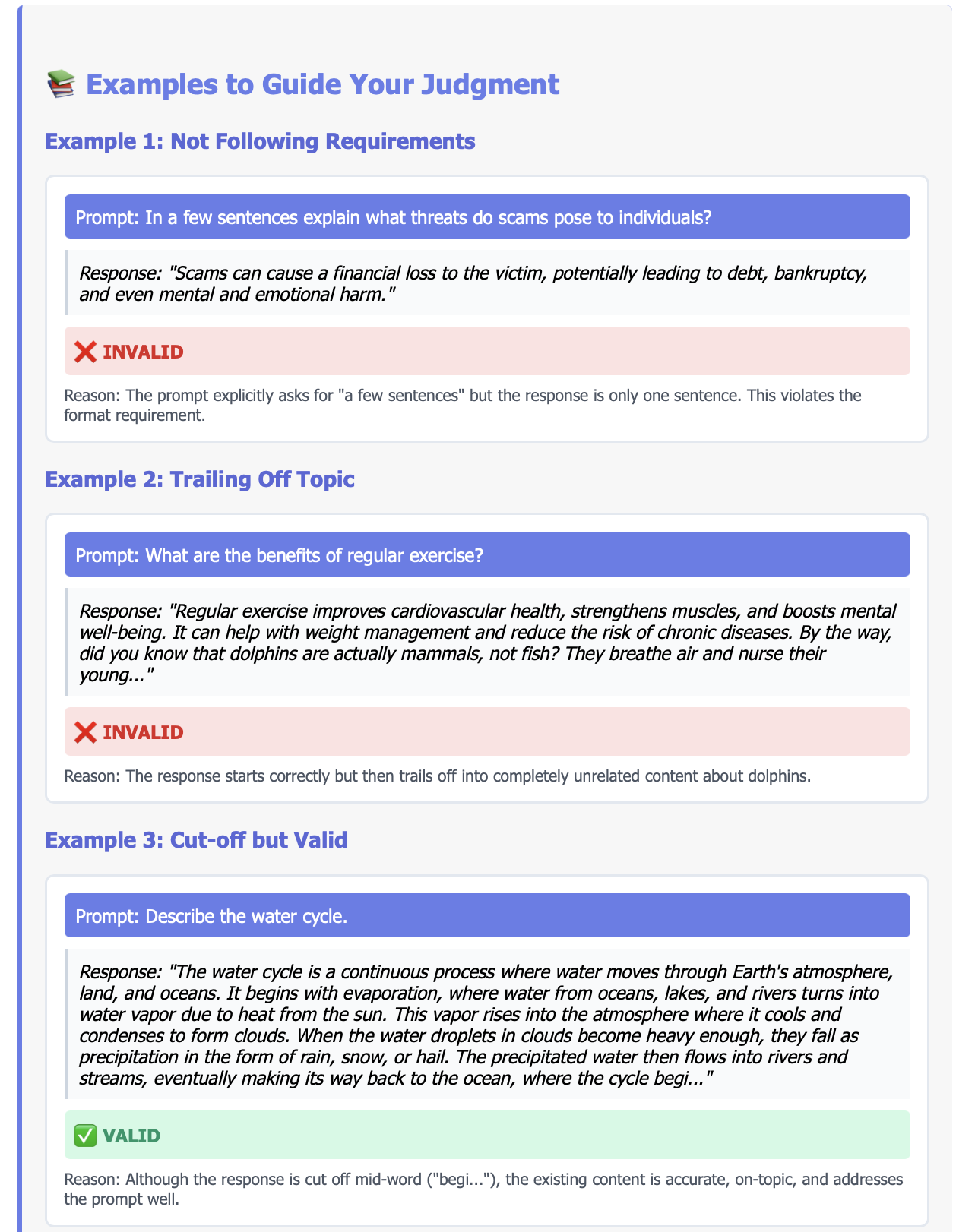}
\includegraphics[width=0.65\linewidth]{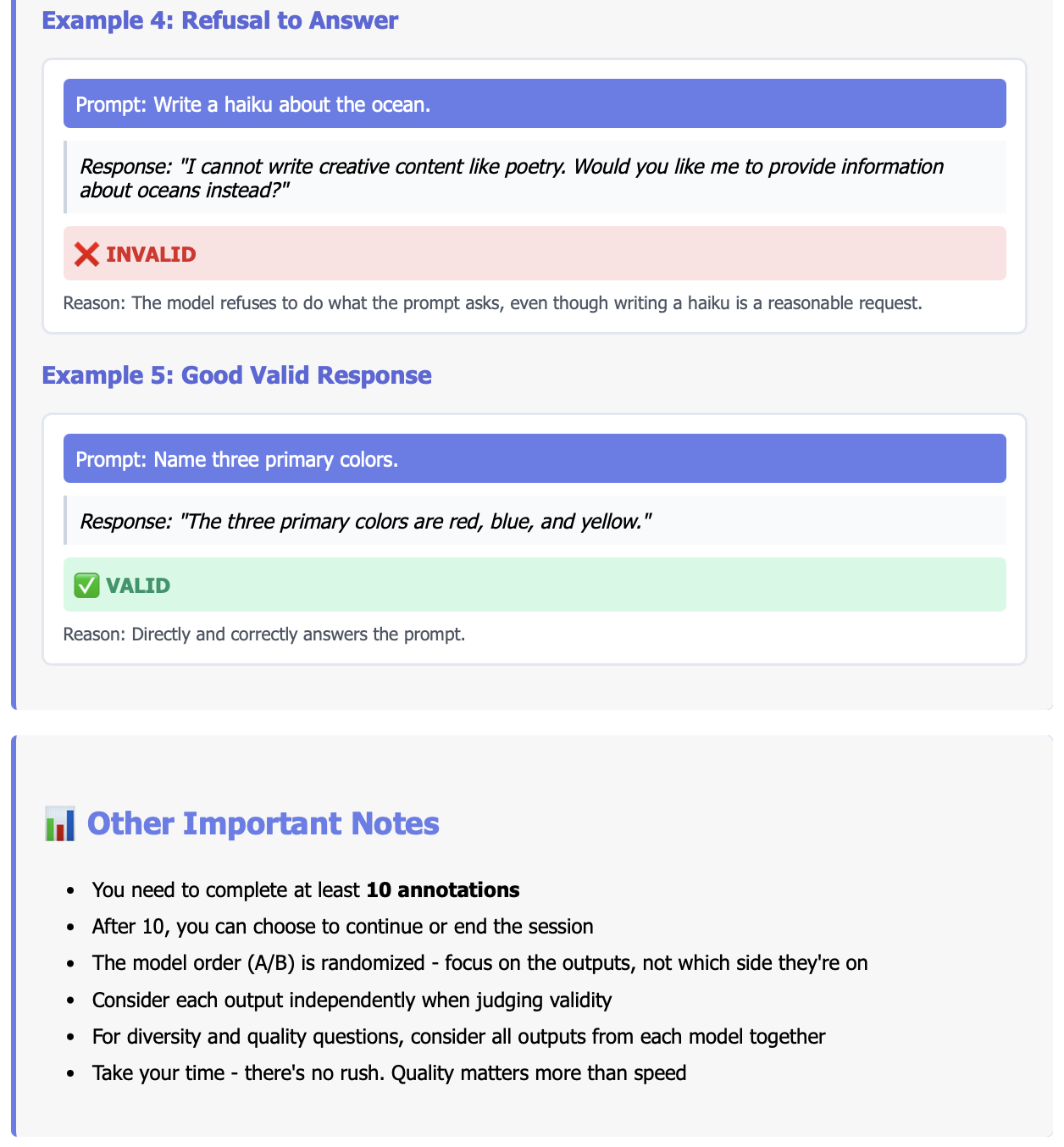}
    \caption{Guideline examples for human annotation}
    \label{fig:guideline_examples}
\end{figure}

\begin{figure}[h]
    \centering
    \includegraphics[width=\linewidth]{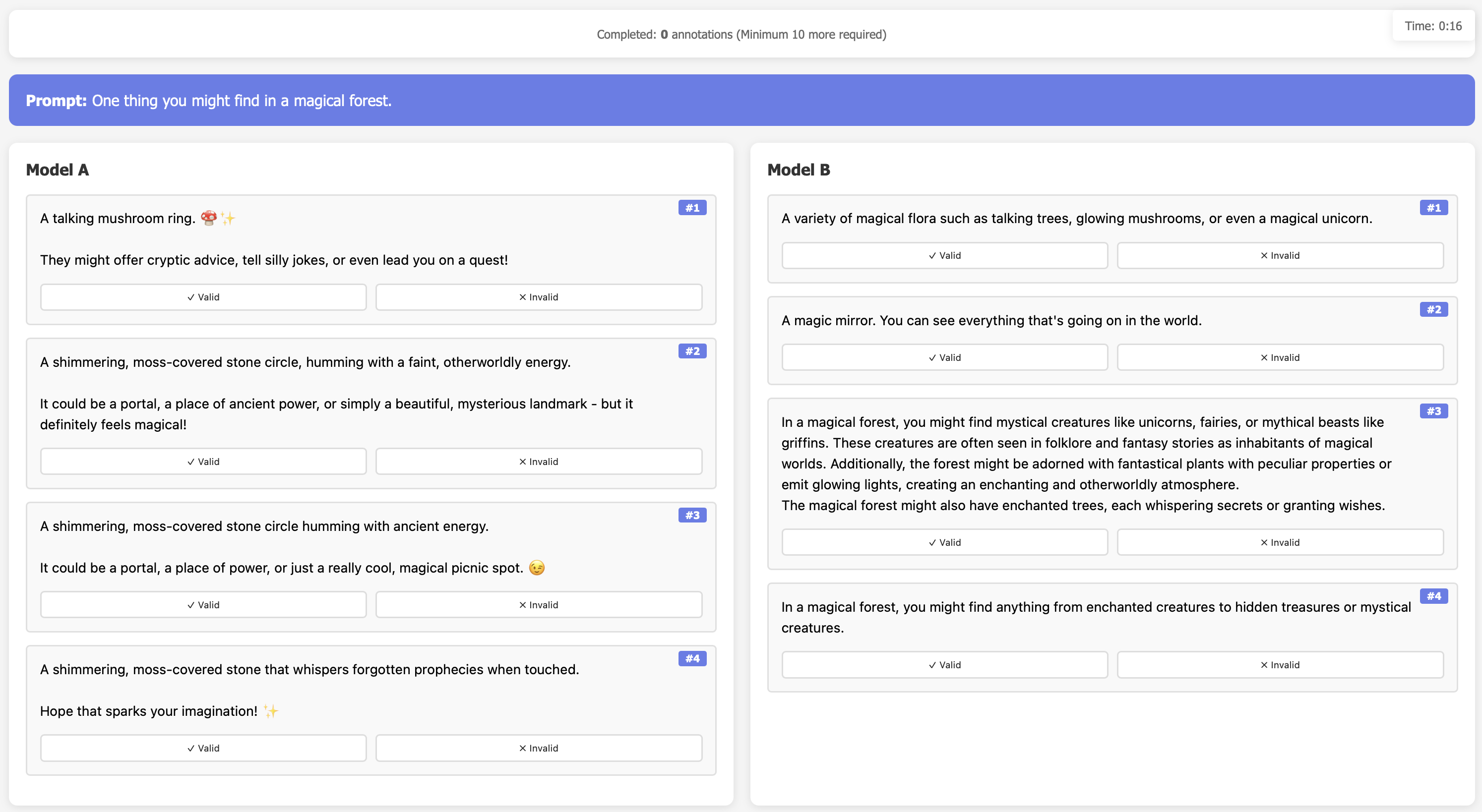}
    \includegraphics[width=\linewidth]{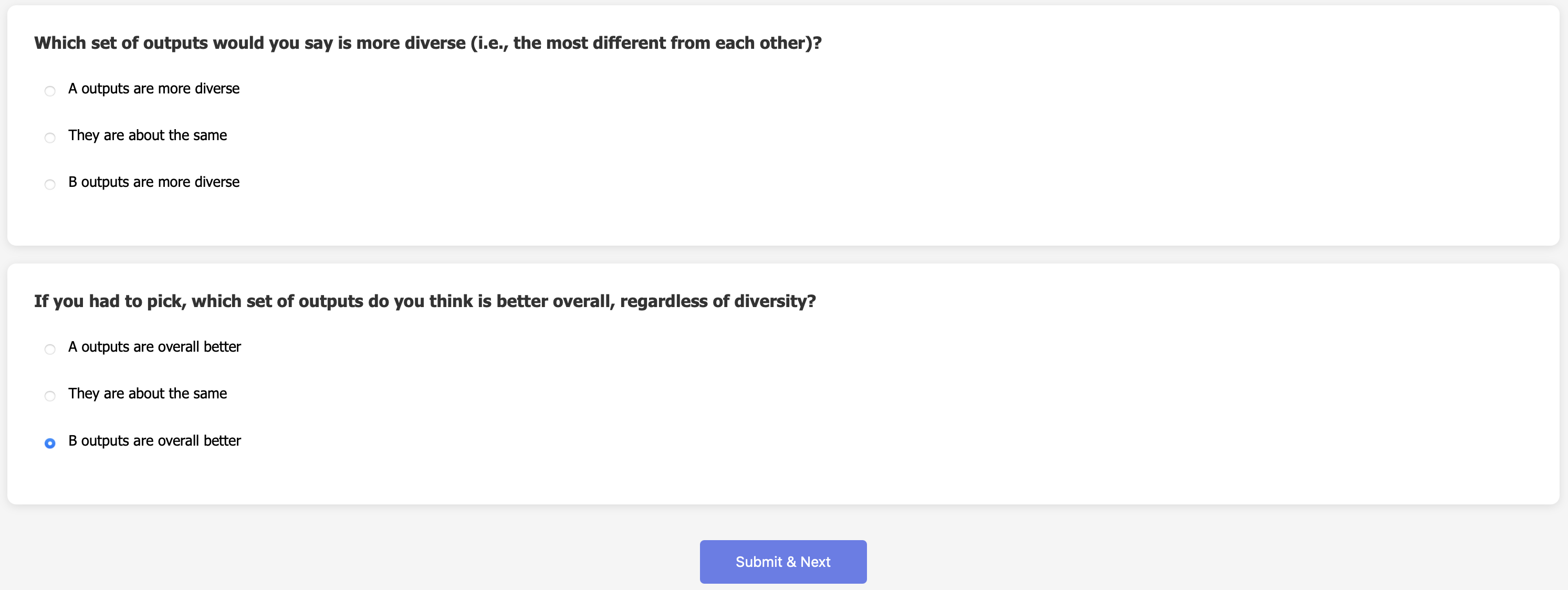}
    \caption{User Interface for Annotation}
    \label{fig:user_annotation_interface}
\end{figure}

\paragraph{Annotation Interface and Procedure}
Participants accessed the annotation task through a web-based interface.
First, participants were asked to thoroughly read through the comprehensive annotation guidelines with examples of valid and invalid responses (See Figure \ref{fig:guidelines} and Figure \ref{fig:guideline_examples}).
For each annotation instance, annotators were presented with a prompt and four generations from each of two models (labeled Model A and Model B). 
The model identities and presentation order were randomized to prevent systematic bias. The interface displayed the outputs side-by-side to facilitate direct comparison (See Figure \ref{fig:user_annotation_interface} for the user interface and questions).

For each task, annotators made three types of judgments:
\begin{itemize}
    \item \textbf{Validity Assessment:} Annotators independently marked each of the eight generations (4 per model) as either valid or invalid. We provided detailed guidelines defining validity as responses that directly address the prompt, follow all specified requirements, stay on-topic throughout, and contain factually reasonable content. Invalid responses included those that refuse to answer, violate format requirements, trail off into unrelated content, or contain significant errors.
    \item \textbf{Diversity Comparison:} Annotators assessed which model's set of four outputs exhibited greater diversity, with options for Model A, Model B, or ``about the same.''
    \item \textbf{Overall Quality Judgment:} Independent of diversity, annotators selected which model's outputs were better overall, again with options for either model or ``about the same.''
\end{itemize}

To ensure annotation quality, we implemented several measures: (1) Comprehensive annotation guidelines with examples of valid and invalid responses, (2) Tracking of time spent per annotation, and (3) Post-annotation feedback collection to identify any systematic issues.  

\paragraph{Inter-Annotator Agreement}
Inter-annotator agreement for validity judgments showed 76.5\% pairwise percentage agreement, with Cohen's $\kappa = 0.441$, indicating moderate agreement. For the subjective diversity and quality assessments, agreement rates were lower (diversity: 38.8\%, quality: 41.7\%), as expected given the more nuanced nature of these judgments.

\section{LLM Usage Description}
\label{app:llmusage}
In preparation of this research and manuscript, LLMs were used for:
\begin{itemize}
    \item Implementing code for experiments and analysis based on detailed author descriptions. All LLM code was inspected by the authors for correctness.
    \item Formatting for tables, latex, and bibtex citation for non-traditional sources (e.g., urls).
    \item Draft critique by pointing out typos and potentially confusing wording in the draft.
\end{itemize}

However, all research ideation and writing was performed solely by the authors, and we claim full responsibility for the contents of this paper.

\clearpage

\section{Supplementary Figures}
\label{app:supplementaryfigures}

\begin{figure}[h]
    \vspace{-10pt}
    \centering
    \includegraphics[width=\textwidth]{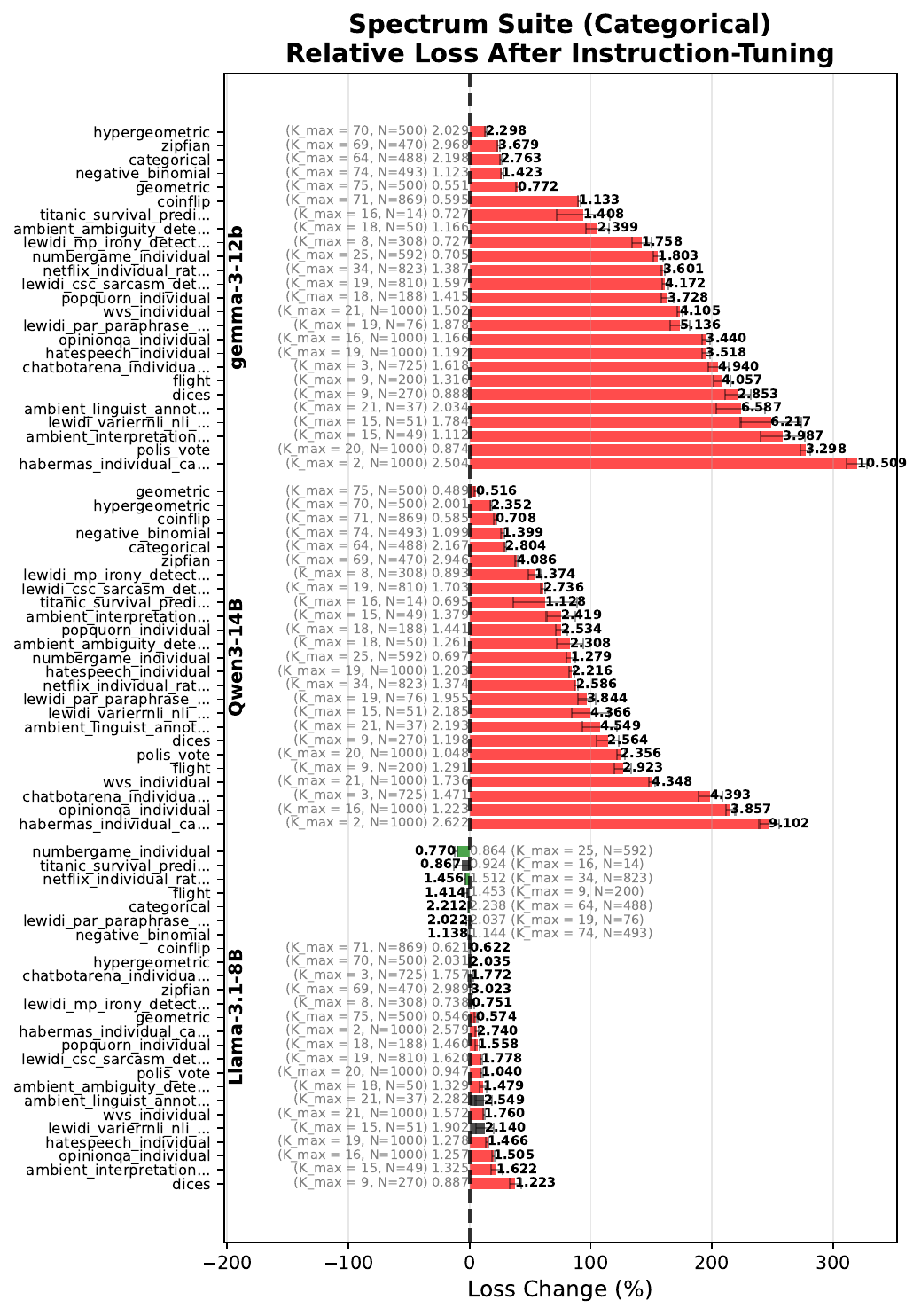}
    \caption{\resource categorical loss after instruction-tuning}
    \vspace{-10pt}
\end{figure}

\begin{figure}[h]
    \vspace{-10pt}
    \centering
    \includegraphics[width=0.8\textwidth]{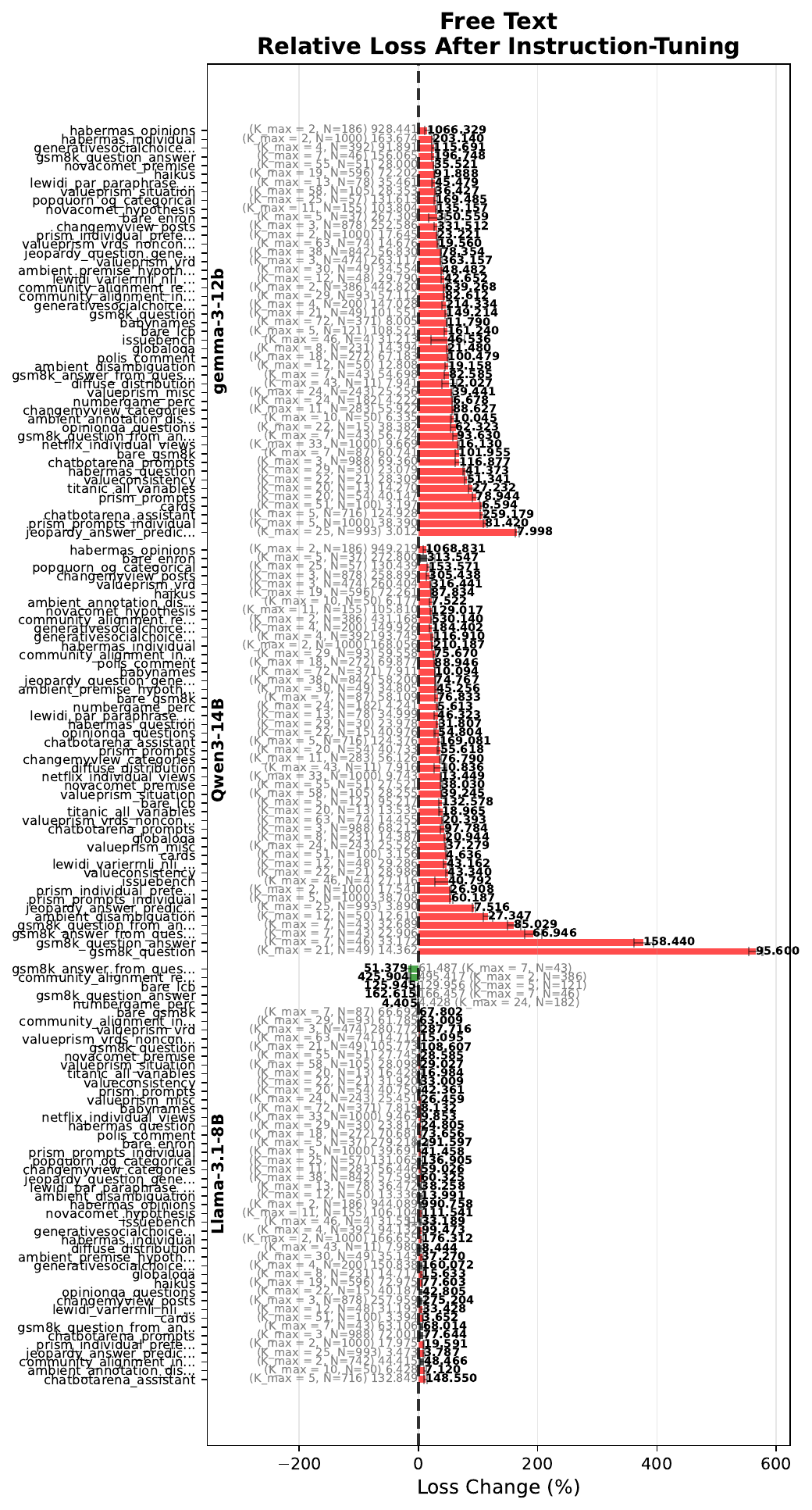}
    \caption{\resource free-text loss after instruction-tuning}
    \vspace{-10pt}
\end{figure}

\begin{figure}[h]
    \vspace{-10pt}
    \centering
    \includegraphics[width=\textwidth]{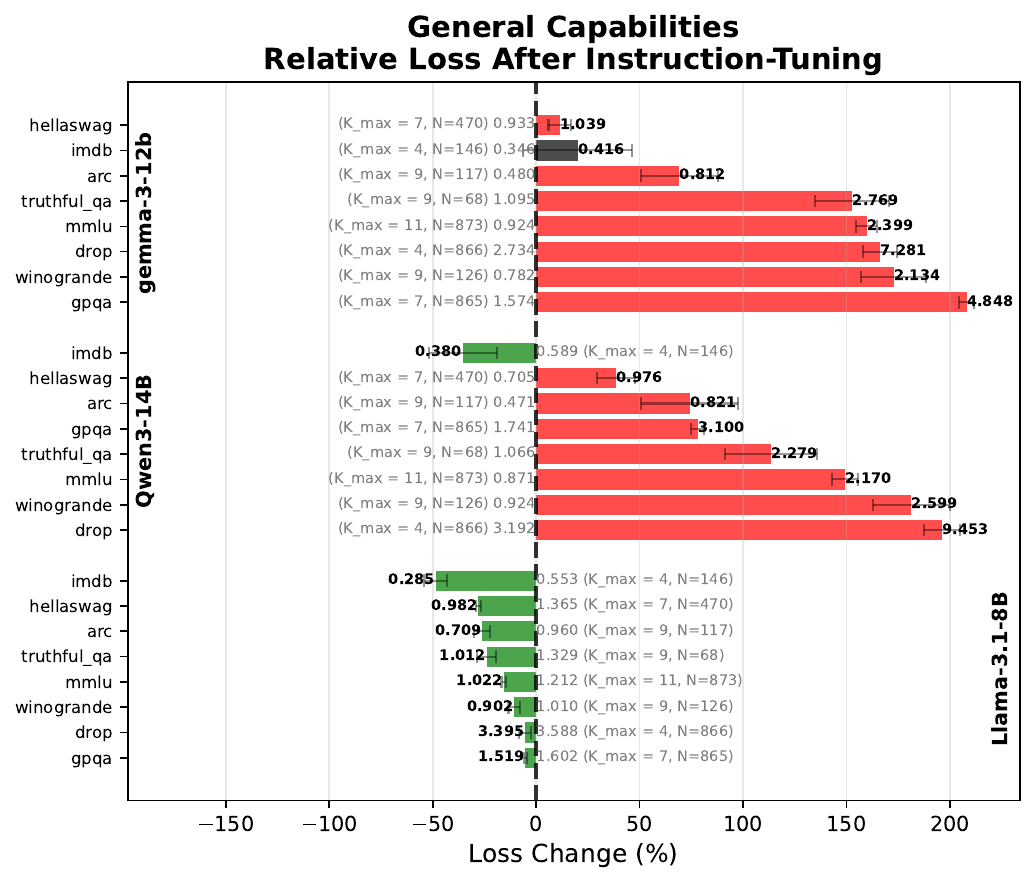}
    \caption{\resource general capability loss after instruction-tuning}
    \vspace{-10pt}
\end{figure}

\clearpage

\section{\method Templates}
\label{app:spectrumtemplates}

For all templates, loss is calculated on the highlighted output tokens.

\textbf{\texttt{gemma-3}} (w/ inputs)
\begin{lstlisting}[style=plaincode]
<start_of_turn>description
DESCRIPTION TEXT<end_of_turn>
<start_of_turn>input
INPUT 1 TEXT<end_of_turn>
<start_of_turn>output
(*@\hlc[hlAmber]{OUTPUT 1 TEXT<end\_of\_turn>}@*)
<start_of_turn>input
INPUT 2 TEXT<end_of_turn>
<start_of_turn>output
(*@\hlc[hlGreen]{OUTPUT 2 TEXT<end\_of\_turn>}@*)
<start_of_turn>input
INPUT 3 TEXT<end_of_turn>
<start_of_turn>output
(*@\hlc[hlBlue]{OUTPUT 3 TEXT<end\_of\_turn>}@*)
...
\end{lstlisting}

\textbf{\texttt{gemma-3}} (w/out inputs)
\begin{lstlisting}[style=plaincode]
<start_of_turn>description
DESCRIPTION TEXT<end_of_turn>
<start_of_turn>output
(*@\hlc[hlAmber]{OUTPUT 1 TEXT<end\_of\_turn>}@*)
<start_of_turn>input
(*@\hlc[hlGreen]{OUTPUT 2 TEXT<end\_of\_turn>}@*)
<start_of_turn>input
(*@\hlc[hlBlue]{OUTPUT 3 TEXT<end\_of\_turn>}@*)
...
\end{lstlisting}

\textbf{\texttt{Qwen3}} (w/ inputs)
\begin{lstlisting}[style=plaincode]
<|im_start|>description
DESCRIPTION TEXT<|im_end|>
<|im_start|>input
INPUT 1 TEXT<|im_end|>
<|im_start|>output
(*@\hlc[hlAmber]{OUTPUT 1 TEXT<|im\_end|>}@*)
<|im_start|>input
INPUT 2 TEXT<|im_end|>
<|im_start|>output
(*@\hlc[hlGreen]{OUTPUT 2 TEXT<|im\_end|>}@*)
<|im_start|>input
INPUT 3 TEXT<|im_end|>
<|im_start|>output
(*@\hlc[hlBlue]{OUTPUT 3 TEXT<|im\_end|>}@*)
...
\end{lstlisting}

\textbf{\texttt{Qwen3}} (w/out inputs)
\begin{lstlisting}[style=plaincode]
<|im_start|>description
DESCRIPTION TEXT<|im_end|>
<|im_start|>output
(*@\hlc[hlAmber]{OUTPUT 1 TEXT<|im\_end|>}@*)
<|im_start|>output
(*@\hlc[hlGreen]{OUTPUT 2 TEXT<|im\_end|>}@*)
<|im_start|>output
(*@\hlc[hlBlue]{OUTPUT 3 TEXT<|im\_end|>}@*)
...
\end{lstlisting}

\textbf{\texttt{Llama-3.1}} (w/ inputs)
\begin{lstlisting}[style=plaincode]
<|start_header_id|>description<|end_header_id|>

DESCRIPTION TEXT<|eot_id|><|start_header_id|>input<|end_header_id|>

INPUT 1 TEXT<|eot_id|><|start_header_id|>output<|end_header_id|>

(*@\hlc[hlAmber]{OUTPUT 1 TEXT<|eot\_id|>}@*)<|start_header_id|>input<|end_header_id|>

INPUT 2 TEXT<|eot_id|><|start_header_id|>output<|end_header_id|>

(*@\hlc[hlGreen]{OUTPUT 2 TEXT<|eot\_id|>}@*)<|start_header_id|>input<|end_header_id|>

INPUT 3 TEXT<|eot_id|><|start_header_id|>output<|end_header_id|>

(*@\hlc[hlBlue]{OUTPUT 3 TEXT<|eot\_id|>}@*)...
\end{lstlisting}

\textbf{\texttt{Llama-3.1}} (w/out inputs)
\begin{lstlisting}[style=plaincode]
<|start_header_id|>description<|end_header_id|>

DESCRIPTION TEXT<|eot_id|><|start_header_id|>output<|end_header_id|>

(*@\hlc[hlAmber]{OUTPUT 1 TEXT<|eot\_id|>}@*)<|start_header_id|>output<|end_header_id|>

(*@\hlc[hlGreen]{OUTPUT 2 TEXT<|eot\_id|>}@*)<|start_header_id|>output<|end_header_id|>

(*@\hlc[hlBlue]{OUTPUT 3 TEXT<|eot\_id|>}@*)...
\end{lstlisting}

\section{Pretrained / Instruction-Tuned ICL Templates}
\label{app:prompttemplates}

\textbf{Pretrained Template} (w/ inputs) \\
Note that each output ends with two newlines to ensure a terminal token (coloring not visible).
\begin{lstlisting}[style=plaincode]
Description: DESCRIPTION TEXT

Input: INPUT 1 TEXT

Output:(*@\hlc[hlAmber]{ OUTPUT 1 TEXT}@*)

Input: INPUT 2 TEXT

Output:(*@\hlc[hlGreen]{ OUTPUT 2 TEXT}@*)

Input: INPUT 3 TEXT

Output:(*@\hlc[hlBlue]{ OUTPUT 3 TEXT}@*)

...
\end{lstlisting}

\textbf{Pretrained Template} (w/out inputs) \\
Note that each output ends with two newlines to ensure a terminal token (coloring not visible).
\begin{lstlisting}[style=plaincode]
Description: DESCRIPTION TEXT

Output:(*@\hlc[hlAmber]{ OUTPUT 1 TEXT}@*)

Output:(*@\hlc[hlGreen]{ OUTPUT 2 TEXT}@*)

Output:(*@\hlc[hlBlue]{ OUTPUT 3 TEXT}@*)

...
\end{lstlisting}

\textbf{Simple Instruct Template} \\
\texttt{Qwen3} (task w/inputs)
\begin{lstlisting}[style=plaincode]
<|im_start|>system
DESCRIPTION TEXT<|im_end|>
<|im_start|>user
INPUT 1 TEXT<|im_end|>
<|im_start|>assistant
<think>

</think>

(*@\hlc[hlAmber]{OUTPUT 1 TEXT<|im\_end|>}@*)
<|im_start|>user
INPUT 2 TEXT<|im_end|>
<|im_start|>assistant
<think>

</think>

(*@\hlc[hlGreen]{OUTPUT 2 TEXT<|im\_end|>}@*)
<|im_start|>user
INPUT 3 TEXT<|im_end|>
<|im_start|>assistant
<think>

</think>

(*@\hlc[hlBlue]{OUTPUT 3 TEXT<|im\_end|>}@*)

\end{lstlisting}
\texttt{Qwen3} (task w/out inputs)
\begin{lstlisting}[style=plaincode]
<|im_start|>system
DESCRIPTION TEXT<|im_end|>
<|im_start|>user
Generate<|im_end|>
<|im_start|>assistant
<think>

</think>

(*@\hlc[hlAmber]{OUTPUT 1 TEXT<|im\_end|>}@*)
<|im_start|>user
Generate<|im_end|>
<|im_start|>assistant
<think>

</think>

(*@\hlc[hlGreen]{OUTPUT 2 TEXT<|im\_end|>}@*)
<|im_start|>user
Generate<|im_end|>
<|im_start|>assistant
<think>

</think>

(*@\hlc[hlBlue]{OUTPUT 3 TEXT<|im\_end|>}@*)

\end{lstlisting}

\texttt{gemma-3} (task w/inputs)
\begin{lstlisting}[style=plaincode]
<start_of_turn>user
DESCRIPTION TEXT

INPUT 1 TEXT<end_of_turn>
<start_of_turn>model
(*@\hlc[hlAmber]{OUTPUT 1 TEXT<end\_of\_turn>}@*)
<start_of_turn>user
INPUT 2 TEXT<end_of_turn>
<start_of_turn>model
(*@\hlc[hlGreen]{OUTPUT 2 TEXT<end\_of\_turn>}@*)
<start_of_turn>user
INPUT 3 TEXT<end_of_turn>
<start_of_turn>model
(*@\hlc[hlBlue]{OUTPUT 3 TEXT<end\_of\_turn>}@*)

\end{lstlisting}

\texttt{gemma-3} (task w/out inputs)
\begin{lstlisting}[style=plaincode]
<start_of_turn>user
DESCRIPTION TEXT

Generate<end_of_turn>
<start_of_turn>model
(*@\hlc[hlAmber]{OUTPUT 1 TEXT<end\_of\_turn>}@*)
<start_of_turn>user
Generate<end_of_turn>
<start_of_turn>model
(*@\hlc[hlGreen]{OUTPUT 2 TEXT<end\_of\_turn>}@*)
<start_of_turn>user
Generate<end_of_turn>
<start_of_turn>model
(*@\hlc[hlBlue]{OUTPUT 3 TEXT<end\_of\_turn>}@*)

\end{lstlisting}

\texttt{Llama-3.1} (task w/inputs)
\begin{lstlisting}[style=plaincode]
<|begin_of_text|><|start_header_id|>system<|end_header_id|>

Cutting Knowledge Date: December 2023
Today Date: DD MM YYYY

DESCRIPTION TEXT<|eot_id|><|start_header_id|>user<|end_header_id|>

INPUT 1 TEXT<|eot_id|><|start_header_id|>assistant<|end_header_id|>

(*@\hlc[hlAmber]{OUTPUT 1 TEXT<|eot\_id|>}@*)<|start_header_id|>user<|end_header_id|>

INPUT 2 TEXT<|eot_id|><|start_header_id|>assistant<|end_header_id|>

(*@\hlc[hlGreen]{OUTPUT 2 TEXT<|eot\_id|>}@*)<|start_header_id|>user<|end_header_id|>

INPUT 3 TEXT<|eot_id|><|start_header_id|>assistant<|end_header_id|>

(*@\hlc[hlBlue]{OUTPUT 3 TEXT<|eot\_id|>}@*)
\end{lstlisting}

\texttt{Llama-3.1} (task w/out inputs)
\begin{lstlisting}[style=plaincode]
<|begin_of_text|><|start_header_id|>system<|end_header_id|>

Cutting Knowledge Date: December 2023
Today Date: 26 Jul 2024

DESCRIPTION TEXT<|eot_id|><|start_header_id|>user<|end_header_id|>

Generate<|eot_id|><|start_header_id|>assistant<|end_header_id|>

(*@\hlc[hlAmber]{OUTPUT 1 TEXT<|eot\_id|>}@*)<|start_header_id|>user<|end_header_id|>

Generate<|eot_id|><|start_header_id|>assistant<|end_header_id|>

(*@\hlc[hlGreen]{OUTPUT 2 TEXT<|eot\_id|>}@*)<|start_header_id|>user<|end_header_id|>

Generate<|eot_id|><|start_header_id|>assistant<|end_header_id|>

(*@\hlc[hlBlue]{OUTPUT 3 TEXT<|eot\_id|>}@*)
\end{lstlisting}

\textbf{Detailed Instruct Template} \\
\texttt{Qwen} (task w/ inputs)

\begin{lstlisting}[style=plaincode]
<|im_start|>system
You are tasked with generating outputs from a particular, potentially stochastic, generative process. You will be given some combination of:
- Description: A natural description of the generative process / data distribution
- Input: An input on which to condition the generative process.
- Example outputs: Example outputs from the process, either in a user message or as prior generations from a chat message. You may assume that any given outputs are exchangeable with one another (order-invariant) and generated from the same process (roughly i.i.d.). If the output data pertains to a single object, it just contains the output. If it contains multiple objects, use json formatting with keys for the name of the output variable.
You will be provided at least either a description or an example output.

Given these components, your job is to generate JUST the output in your response, roughly approximating the underlying generative process, maintaining any underlying stochasticity (if any is present). If you are asked to generate again, you will either be given an additional input to condition on, or will just be told to "Generate".


Description: DESCRIPTION TEXT<|im_end|>
<|im_start|>user
INPUT 1 TEXT<|im_end|>
<|im_start|>assistant
<think>

</think>

(*@\hlc[hlAmber]{OUTPUT 1 TEXT<|im\_end|>}@*)
<|im_start|>user
INPUT 2 TEXT<|im_end|>
<|im_start|>assistant
<think>

</think>

(*@\hlc[hlGreen]{OUTPUT 2 TEXT<|im\_end|>}@*)
<|im_start|>user
INPUT 3 TEXT<|im_end|>
<|im_start|>assistant
<think>

</think>

(*@\hlc[hlBlue]{OUTPUT 3 TEXT<|im\_end|>}@*)

\end{lstlisting}

\texttt{Qwen} (task w/out inputs)
\begin{lstlisting}[style=plaincode]
<|im_start|>system
You are tasked with generating outputs from a particular, potentially stochastic, generative process. You will be given some combination of:
- Description: A natural description of the generative process / data distribution
- Input: An input on which to condition the generative process.
- Example outputs: Example outputs from the process, either in a user message or as prior generations from a chat message. You may assume that any given outputs are exchangeable with one another (order-invariant) and generated from the same process (roughly i.i.d.). If the output data pertains to a single object, it just contains the output. If it contains multiple objects, use json formatting with keys for the name of the output variable.
You will be provided at least either a description or an example output.

Given these components, your job is to generate JUST the output in your response, roughly approximating the underlying generative process, maintaining any underlying stochasticity (if any is present). If you are asked to generate again, you will either be given an additional input to condition on, or will just be told to "Generate".


Description: DESCRIPTION TEXT<|im_end|>
<|im_start|>user
Generate<|im_end|>
<|im_start|>assistant
<think>

</think>

(*@\hlc[hlAmber]{OUTPUT 1 TEXT<|im\_end|>}@*)
<|im_start|>user
Generate<|im_end|>
<|im_start|>assistant
<think>

</think>

(*@\hlc[hlGreen]{OUTPUT 2 TEXT<|im\_end|>}@*)
<|im_start|>user
Generate<|im_end|>
<|im_start|>assistant
<think>

</think>

(*@\hlc[hlBlue]{OUTPUT 3 TEXT<|im\_end|>}@*)

\end{lstlisting}

\texttt{gemma-3} (task w/inputs)
\begin{lstlisting}[style=plaincode]
<start_of_turn>user
You are tasked with generating outputs from a particular, potentially stochastic, generative process. You will be given some combination of:
- Description: A natural description of the generative process / data distribution
- Input: An input on which to condition the generative process.
- Example outputs: Example outputs from the process, either in a user message or as prior generations from a chat message. You may assume that any given outputs are exchangeable with one another (order-invariant) and generated from the same process (roughly i.i.d.). If the output data pertains to a single object, it just contains the output. If it contains multiple objects, use json formatting with keys for the name of the output variable.
You will be provided at least either a description or an example output.

Given these components, your job is to generate JUST the output in your response, roughly approximating the underlying generative process, maintaining any underlying stochasticity (if any is present). If you are asked to generate again, you will either be given an additional input to condition on, or will just be told to "Generate".


Description: DESCRIPTION TEXT

INPUT 1 TEXT<end_of_turn>
<start_of_turn>model
(*@\hlc[hlAmber]{OUTPUT 1 TEXT<end\_of\_turn>}@*)
<start_of_turn>user
INPUT 2 TEXT<end_of_turn>
<start_of_turn>model
(*@\hlc[hlGreen]{OUTPUT 2 TEXT<end\_of\_turn>}@*)
<start_of_turn>user
INPUT 3 TEXT<end_of_turn>
<start_of_turn>model
(*@\hlc[hlBlue]{OUTPUT 3 TEXT<end\_of\_turn>}@*)

\end{lstlisting}

\texttt{gemma-3} (task w/out inputs)
\begin{lstlisting}[style=plaincode]
<start_of_turn>user
You are tasked with generating outputs from a particular, potentially stochastic, generative process. You will be given some combination of:
- Description: A natural description of the generative process / data distribution
- Input: An input on which to condition the generative process.
- Example outputs: Example outputs from the process, either in a user message or as prior generations from a chat message. You may assume that any given outputs are exchangeable with one another (order-invariant) and generated from the same process (roughly i.i.d.). If the output data pertains to a single object, it just contains the output. If it contains multiple objects, use json formatting with keys for the name of the output variable.
You will be provided at least either a description or an example output.

Given these components, your job is to generate JUST the output in your response, roughly approximating the underlying generative process, maintaining any underlying stochasticity (if any is present). If you are asked to generate again, you will either be given an additional input to condition on, or will just be told to "Generate".


Description: DESCRIPTION TEXT

Generate<end_of_turn>
<start_of_turn>model
(*@\hlc[hlAmber]{OUTPUT 1 TEXT<end\_of\_turn>}@*)
<start_of_turn>user
Generate<end_of_turn>
<start_of_turn>model
(*@\hlc[hlGreen]{OUTPUT 2 TEXT<end\_of\_turn>}@*)
<start_of_turn>user
Generate<end_of_turn>
<start_of_turn>model
(*@\hlc[hlBlue]{OUTPUT 3 TEXT<end\_of\_turn>}@*)

\end{lstlisting}

\texttt{Llama-3.1} (task w/inputs)
\begin{lstlisting}[style=plaincode]
<|begin_of_text|><|start_header_id|>system<|end_header_id|>

Cutting Knowledge Date: December 2023
Today Date: DD MM YYYY

You are tasked with generating outputs from a particular, potentially stochastic, generative process. You will be given some combination of:
- Description: A natural description of the generative process / data distribution
- Input: An input on which to condition the generative process.
- Example outputs: Example outputs from the process, either in a user message or as prior generations from a chat message. You may assume that any given outputs are exchangeable with one another (order-invariant) and generated from the same process (roughly i.i.d.). If the output data pertains to a single object, it just contains the output. If it contains multiple objects, use json formatting with keys for the name of the output variable.
You will be provided at least either a description or an example output.

Given these components, your job is to generate JUST the output in your response, roughly approximating the underlying generative process, maintaining any underlying stochasticity (if any is present). If you are asked to generate again, you will either be given an additional input to condition on, or will just be told to "Generate".


Description: DESCRIPTION TEXT<|eot_id|><|start_header_id|>user<|end_header_id|>

INPUT 1 TEXT<|eot_id|><|start_header_id|>assistant<|end_header_id|>

(*@\hlc[hlAmber]{OUTPUT 1 TEXT<|eot\_id|>}@*)<|start_header_id|>user<|end_header_id|>

INPUT 2 TEXT<|eot_id|><|start_header_id|>assistant<|end_header_id|>

(*@\hlc[hlGreen]{OUTPUT 2 TEXT<|eot\_id|>}@*)<|start_header_id|>user<|end_header_id|>

INPUT 3 TEXT<|eot_id|><|start_header_id|>assistant<|end_header_id|>

(*@\hlc[hlBlue]{OUTPUT 3 TEXT<|eot\_id|>}@*)

\end{lstlisting}

\texttt{Llama-3.1} (task w/out inputs)
\begin{lstlisting}[style=plaincode]
<|begin_of_text|><|start_header_id|>system<|end_header_id|>

Cutting Knowledge Date: December 2023
Today Date: DD MM YYYY

You are tasked with generating outputs from a particular, potentially stochastic, generative process. You will be given some combination of:
- Description: A natural description of the generative process / data distribution
- Input: An input on which to condition the generative process.
- Example outputs: Example outputs from the process, either in a user message or as prior generations from a chat message. You may assume that any given outputs are exchangeable with one another (order-invariant) and generated from the same process (roughly i.i.d.). If the output data pertains to a single object, it just contains the output. If it contains multiple objects, use json formatting with keys for the name of the output variable.
You will be provided at least either a description or an example output.

Given these components, your job is to generate JUST the output in your response, roughly approximating the underlying generative process, maintaining any underlying stochasticity (if any is present). If you are asked to generate again, you will either be given an additional input to condition on, or will just be told to "Generate".


Description: DESCRIPTION TEXT<|eot_id|><|start_header_id|>user<|end_header_id|>

Generate<|eot_id|><|start_header_id|>assistant<|end_header_id|>

(*@\hlc[hlAmber]{OUTPUT 1 TEXT<|eot\_id|>}@*)<|start_header_id|>user<|end_header_id|>

Generate<|eot_id|><|start_header_id|>assistant<|end_header_id|>

(*@\hlc[hlGreen]{OUTPUT 2 TEXT<|eot\_id|>}@*)<|start_header_id|>user<|end_header_id|>

Generate<|eot_id|><|start_header_id|>assistant<|end_header_id|>

(*@\hlc[hlBlue]{OUTPUT 3 TEXT<|eot\_id|>}@*)
\end{lstlisting}

\setcounter{figure}{0}
\renewcommand{\thefigure}{A\arabic{figure}}

\textbf{Best performing instruct prompts} \\
We found that \texttt{Llama-3.1-8B-Instruct} performed best on \resource with the pretrained prompt, \texttt{google/gemma-3-12b-it} and \texttt{qwen/Qwen3-14B} performed best with the detailed instruct prompt. We utilize those prompts with the corresponding models for all ICL experiments.

\section{Output Coverage / Diversity vs. Validity Experiment Details}
\label{app:diversityvalidity}

\subsection{Verifiable Evaluation}
For this evaluation, we utilize the same prompts as in the ICL experiments - see App. \ref{app:prompttemplates}.

Below, we include the description and examples for each of the tasks. Please reference the codebase for validation functions.

\begin{lstlisting}[style=plaincode]
Task: color_interesting_ex
Description: Generate a color name.
Examples: ['Otterly Brown', 'Petal Pink', 'Cherry']

Task: color_normal_ex
Description: Generate a color name.
Examples: ['Green', 'Red', 'White']

Task: car_brand
Description: Car brand.
Examples: ['Acura', 'Ford', 'Tesla']

Task: car_make_model
Description: Car make and model.
Examples: ['Acura Integra', 'Ford Mustang', 'Tesla Model 3']

Task: us_states_abbreviations
Description: US state abbreviation
Examples: ['KY', 'UT', 'OR']

Task: us_states_any_format
Description: US state name or abbreviation
Examples: ['Kentucky', 'UT', 'Oregon']

Task: us_states_full_names
Description: Name a US state
Examples: ['Kentucky', 'Utah', 'Oregon']

Task: prime_numbers
Description: Generate a prime number
Examples: ['617', '13', '47']

Task: small_prime_numbers
Description: Generate a prime number less than 100
Examples: ['29', '5', '97']

Task: basic_emails
Description: Email address
Examples: ['ANONYMIZED', 'alex.jones@domain.net', 'itsagoodday@gmail.com']

Task: professional_emails
Description: Generate a professional email address.
Examples: ['ANONYMIZED', 'sarah.johannesburg@organization.org', 'yash@anthropic.com']                                   

Task: weekdays_abbreviated
Description: Day of the week abbreviation
Examples: ['Thu', 'Wed.', 'SUN']

Task: weekdays_any_format
Description: Day of the week (full name or abbreviation)
Examples: ['Monday', 'Tue', 'SUN']

Task: weekdays_full
Description: Name a day of the week
Examples: ['Thursday', 'Wednesday', 'Sunday']

Task: random_seed
Description: Generate a number to use for a random seed.
Examples: ['15', '420', '8392013']

Task: claude_gerunds
Description: Generate an English gerund ending in -ing.
Examples: ['Schlepping', 'Hoisting', 'Thinking']

Task: rng_1_10
Description: Generate a number between 1 and 10.
Examples: ['3', '7', '10']

Task: rng_1_100
Description: Generate a number between 1 and 100.
Examples: ['35', '94', '71']

Task: international_phone_numbers
Description: International phone number with country code.
Examples: ['+1 413-121-2591', '+44 10 2958 3938', '+81 3 8328 5625']                                                                  
Task: us_phone_numbers
Description: US phone number
Examples: ['(210) 346-0967', '646 562-1938', '800-895-0522']

Task: element_names
Description: Chemical element name
Examples: ['Niobium', 'Antimony', 'Iodine']

Task: element_symbols
Description: Chemical element symbol
Examples: ['Sb', 'He', 'W']
\end{lstlisting}

\subsection{In-context Chat Elicitation for Pretrained and Instruction-Tuned Models}
For inducing chat capabilities from the pretrained/spectrum-tuned models for the NoveltyBench/Infinite-Chats-Eval experiments, we utilize the following few-shot examples. Responses are generated by an instruct model in the same model family: \texttt{google/gemma-3-27b-it}, \texttt{qwen/Qwen3-32B}, \texttt{meta/Llama-3.1-70B-Instruct}.

For all pretrained models, we utilized the following prompt template:

\begin{lstlisting}[style=plaincode]
You are a helpful AI assistant.

User: <EXAMPLE PROMPT1>

Assistant: <EXAMPLE RESPONSE1>

...

User: <INFERENCE PROMPT>

Assistant:
\end{lstlisting}

For the spectrum-tuned models, we use \texttt{You are a helpful AI assistant.} as the \texttt{description}, the example prompt as \texttt{input}, and the example response as \texttt{output}, using the corresponding model's \method template.

Below, we include the demonstrations for each model family.

\textbf{\texttt{gemma-3-12b}}

\begin{lstlisting}[style=plaincode]
{"prompt": "What is the capital of France?", "response": "The capital of France is **Paris**. \n\nIt's known for its iconic landmarks like the Eiffel Tower, Louvre Museum, and Arc de Triomphe, as well as its fashion, cuisine, and culture."}
{"prompt": "What color should I paint my car? Respond with just a color", "response": "Forest green."}
{"prompt": "Explain the 2008 financial crisis to me in about one paragraph.", "response": "The 2008 financial crisis was a severe worldwide economic downturn triggered by the collapse of the US housing market. Years of low interest rates and lax lending standards led to a boom in subprime mortgages and loans given to borrowers with poor credit. These mortgages were packaged into complex financial products called mortgage-backed securities and sold to investors globally. When housing prices began to fall in 2006-2007, many borrowers defaulted, causing these securities to lose value. This triggered a cascading effect: banks and financial institutions holding these assets faced massive losses, credit markets froze, businesses couldn't get loans, and ultimately, major institutions like Lehman Brothers collapsed. The crisis spread globally, leading to a severe recession, job losses, and government bailouts to stabilize the financial system."}
{"prompt": "Tell me a joke about Texas.", "response": "Why did the Texan bring a ladder to the rodeo? \n\n...He heard the steaks were high! \n\n<Face with Tears of Joy emoji>\n\n\n\nHope that gave you a little chuckle! Let me know if you'd like to hear another one."}
\end{lstlisting}

\textbf{\texttt{Llama-3.1-8B}}

\begin{lstlisting}[style=plaincode]
{"prompt": "What is the capital of France?", "response": "The capital of France is Paris."}
{"prompt": "What color should I paint my car? Respond with just a color", "response": "Navy blue"}
{"prompt": "Explain the 2008 financial crisis to me in about one paragraph.", "response": "The 2008 financial crisis, also known as the Global Financial Crisis (GFC), was a severe economic downturn that began in the United States in 2007 and spread globally, lasting through 2009. The crisis was triggered by a housing market bubble bursting in the US, fueled by lax lending practices and excessive speculation. Many homeowners, especially those with subprime mortgages, were unable to afford their mortgage payments, leading to widespread defaults and foreclosures. This created a crisis in the financial markets, as many banks and financial institutions had invested heavily in mortgage-backed securities (MBS) that were now worthless. As a result, these institutions found themselves facing massive losses, which in turn led to a credit crisis, stock market crash, and a global recession. The crisis was exacerbated by other factors, including over-reliance on complex financial instruments, inadequate regulation, and a global economic imbalance. The crisis led to unprecedented government interventions, including massive bailouts and stimulus packages, to stabilize the financial system and stimulate economic growth."}
{"prompt": "Tell me a joke about Texas.", "response": "Why did the Texan bring a ladder to the party? \n\nBecause he heard the drinks were on the house."}
\end{lstlisting}

\textbf{\texttt{Qwen3-14B}}
\begin{lstlisting}[style=plaincode]
{"prompt": "What is the capital of France?", "response": "The capital of France is **Paris**."}
{"prompt": "What color should I paint my car? Respond with just a color", "response": "Crimson Red"}
{"prompt": "Explain the 2008 financial crisis to me in about one paragraph.", "response": "The 2008 financial crisis was triggered by a housing market bubble in the United States that burst, leading to a severe credit crunch and global economic downturn. Starting in the early 2000s, lax lending standards and speculative investment fueled a rapid rise in home prices. Banks and financial institutions began offering high-risk subprime mortgages to borrowers with poor credit, then bundled these loans into complex financial products like mortgage-backed securities (MBS) and collateralized debt obligations (CDOs), which were widely traded. When housing prices peaked and began to fall in 2006-2007, many borrowers defaulted on their mortgages, causing the value of these financial instruments to plummet. This led to massive losses for banks and investors, triggering a wave of bankruptcies and failures, including major institutions like Lehman Brothers. The crisis spread globally as financial systems were interconnected, resulting in a deep recession with soaring unemployment, foreclosures, and a widespread loss of consumer and business confidence. In response, governments and central banks implemented massive bailouts and stimulus measures to stabilize the economy."}
{"prompt": "Tell me a joke about Texas.", "response": "Sure! Here's a Texas-themed joke for you:\n\nWhy don't cowboys in Texas ever get cold?\n\nBecause the only *draft* they know is from the beer and the wind... and even that's a *light beer breeze*!"}
\end{lstlisting}

\section{Distributional Alignment Experiment Details}
\label{app:distributionalalignment}

\subsection{Example Prompts}
Here, we include example prompts for the urn task across spectrum/pretrained/instruction-tuned models. The same prompts were used for all model families and tasks, only varying the task and output descriptions.

\textbf{\texttt{spectrum}}
\begin{lstlisting}[style=plaincode]
<start_of_turn>description
There is an urn with the following balls shuffled together: 1 red ball, 2 blue balls, 4 orange balls and 6 yellow balls.<end_of_turn>
<start_of_turn>input
Draw a ball at random, and tell me the color (lowercase).<end_of_turn>
<start_of_turn>output
\end{lstlisting}

\begin{lstlisting}[style=plaincode]
Target outputs: ['red' 'blue' 'orange' 'yellow']
Target distribution: [0.07692307692307693, 0.15384615384615385, 0.3076923076923077, 0.46153846153846156]
\end{lstlisting}

\textbf{\texttt{pretrained}}
\begin{lstlisting}[style=plaincode]
Description: There is an urn with the following balls shuffled together: 1 red ball, 2 blue balls, 4 orange balls and 6 yellow balls.

Input: Draw a ball at random, and tell me the color (lowercase).

Output:
\end{lstlisting}

\begin{lstlisting}[style=plaincode]
Target outputs: [' red' ' blue' ' orange' ' yellow']
Target distribution: [0.07692307692307693, 0.15384615384615385, 0.3076923076923077, 0.46153846153846156]
\end{lstlisting}

\textbf{\texttt{chat}}
\begin{lstlisting}[style=plaincode]
<start_of_turn>user
Given the description and an input, respond with just an output.

Description: There is an urn with the following balls shuffled together: 1 red ball, 2 blue balls, 4 orange balls and 6 yellow balls.

Draw a ball at random, and tell me the color (lowercase).<end_of_turn>
<start_of_turn>model
\end{lstlisting}

\begin{lstlisting}[style=plaincode]
Target outputs: ['red' 'blue' 'orange' 'yellow']
Target distribution: [0.07692307692307693, 0.15384615384615385, 0.3076923076923077, 0.46153846153846156]
\end{lstlisting}

\subsection{Task Examples}
Here, we show an example data instance from each distributional alignment dataset.

\textbf{\texttt{global\_oqa}}
\begin{lstlisting}[style=plaincode]
<start_of_turn>description
Responses from a person from this country: Uruguay<end_of_turn>
<start_of_turn>input
Please tell me for each of the following statements whether you think it can always be justified, never be justified, or something in between, using this card.

Terrorism as a political, ideological or religious mean
Options:
A. Never justifiable
B. 2
C. 3
D. 4
E. 5
F. 6
G. 7
H. 8
I. 9
J. Always justifiable
K. Don't know
L. No answer
M. Missing; Not available<end_of_turn>
<start_of_turn>output
\end{lstlisting}

\begin{lstlisting}[style=plaincode]
Target outputs: ['A', 'B', 'C', 'D', 'E', 'F', 'G', 'H', 'I', 'J', 'K', 'L', 'M']
Target distribution: [0.8281718281718281, 0.046953046953046945, 0.020979020979020976, 0.012987012987012986, 0.01898101898101898, 0.004995004995004994, 0.011988011988011986, 0.007992007992007992, 0.013986013986013983, 0.014985014985014982, 0.009990009990009988, 0.007992007992007992, 0.0]

\end{lstlisting}

\textbf{\texttt{nytimes\_books}}
\begin{lstlisting}[style=plaincode]
<start_of_turn>description
You are a random survey respondent. Respondent demographic: Democrat<end_of_turn>
<start_of_turn>input
How likely are you to read this book?
Book: The Bee Sting By Paul Murray
Genre: Fiction
Summary: In Murray's boisterous tragicomic novel, a once wealthy Irish family struggles with both the aftermath of the 2008 financial crash and their own inner demons.
Options:
1: Very unlikely
2: Somewhat unlikely
3: Somewhat likely
4: Very likely<end_of_turn>
<start_of_turn>output
\end{lstlisting}

\begin{lstlisting}[style=plaincode]
Target outputs: ['1', '2', '3', '4']
Target distribution: [0.45, 0.3, 0.15, 0.1]
\end{lstlisting}

\textbf{\texttt{Machine Personality Inventory}}
\begin{lstlisting}[style=plaincode]
<start_of_turn>description
You are a random survey respondent. Rate your agreement with the following statement on a scale from 0-5, where 0 means "strongly disagree" and 5 means "strongly agree."<end_of_turn>
<start_of_turn>input
Statement: You worry about things
Options:
0: Strongly disagree
1: Disagree
2: Slightly disagree
3: Slightly agree
4: Agree
5: Strongly agree<end_of_turn>
<start_of_turn>output
\end{lstlisting}

\begin{lstlisting}[style=plaincode]
Target outputs: ['0', '1', '2', '3', '4', '5']
Target distribution: [0.0012678672373415167, 0.04066219817491722, 0.14105467172736816, 0.13365904869579262, 0.4568359848178955, 0.22652022934668498]
\end{lstlisting}

\textbf{\texttt{Rotten Tomatoes}}
\begin{lstlisting}[style=plaincode]
<start_of_turn>description
You are a movie critic. Given a movie, you are asked to simply rate it as "Good" or "Bad".<end_of_turn>
<start_of_turn>input
Movie: Rambo III
Release Date: Released May 25, 1988<end_of_turn>
<start_of_turn>output
\end{lstlisting}

\begin{lstlisting}[style=plaincode]
Target outputs: ['Good', 'Bad']
Target distribution: [0.41, 0.59]
\end{lstlisting}

\textbf{\texttt{Habermas}}
\begin{lstlisting}[style=plaincode]
<start_of_turn>description
You are a randomly selected UK resident. You will be given a question and two statements, A and B. Rate which statement you most agree with on a likert scale from 1 to 7:
1: Strongly Agree with A
2: Agree with A
3: Somewhat Agree with A
4: Neutral
5: Somewhat Agree with B
6: Agree with B
7: Strongly Agree with B<end_of_turn>
<start_of_turn>input
Question: Should we ban right turns in central London?
A: We should ban right turns in central London.
B: We should NOT ban right turns in central London.<end_of_turn>
<start_of_turn>output
\end{lstlisting}

\begin{lstlisting}[style=plaincode]
Target outputs: ['1', '2', '3', '4', '5', '6', '7']
Target distribution: [0.0, 0.0, 0.04, 0.24, 0.08, 0.16, 0.48]
\end{lstlisting}

\textbf{\texttt{Numbergame}}
\begin{lstlisting}[style=plaincode]
<start_of_turn>description
You are a randomly selected participant in a study. You will be given a set of numbers which all belong to the same set or pattern, and will be given a target number which may or may not belong to the same set or pattern. Answer Yes if you think that the target number belongs to the same set, otherwise answer No.<end_of_turn>
<start_of_turn>input
Example set: 84, 94, 34
Target number: 5<end_of_turn>
<start_of_turn>output
\end{lstlisting}

\begin{lstlisting}[style=plaincode]
Target outputs: ['Yes', 'No']
Target distribution: [0.0, 1.0]
\end{lstlisting}

\textbf{\texttt{Urn}}
\begin{lstlisting}[style=plaincode]
<start_of_turn>description
There is an urn with the following balls shuffled together: 1 red ball, 2 blue balls, 4 orange balls and 6 yellow balls.<end_of_turn>
<start_of_turn>input
Draw a ball at random, and tell me the color (lowercase).<end_of_turn>
<start_of_turn>output
\end{lstlisting}

\begin{lstlisting}[style=plaincode]
Target outputs: ['red' 'blue' 'orange' 'yellow']
Target distribution: [0.07692307692307693, 0.15384615384615385, 0.3076923076923077, 0.46153846153846156]
\end{lstlisting}

\section{Example Task Prompts}
\label{app:exampletaskpromptstest}
For brevity, here we include example prompts (specifically, for the Gemma models) only for the test split of \resource. To see examples for all tasks, please reference:
\url{https://tsor13.github.io/files/spectrumprompts.pdf}

\subsection{\resource-Test: Multiple-Choice}
\textbf{\texttt{chatbotarena\_individual\_prefs}}
\begin{lstlisting}[style=plaincode]
<start_of_turn>description
The following are preferences for a single individual. They will prompt a language model (opening_prompt), and then will compare the two responses (model_a_response and model_b_response). The user will then rate the winner, with one of the following options: ['b' 'tie' 'a' 'bothbad']<end_of_turn>
<start_of_turn>input
{"opening_prompt": "What does the saying \"pipeline is lifeline\" mean?", "model_a_response": "The saying \"pipeline is lifeline\" refers to the importance of having a reliable and efficient system for transporting resources. In the context of oil and gas, a pipeline is a critical part of the supply chain, as it allows companies to transport oil and gas from production sites to refineries and distribution terminals. A reliable pipeline system is essential to ensuring that oil and gas are available when and where they are needed.\n\nThe saying \"pipeline is lifeline\" can also be used in a more general sense to refer to any system that is essential to the functioning of an organization or society. For example, the transportation system is often referred to as the \"lifeline of the economy,\" as it allows goods and services to be moved around the country. The power grid is another example of a lifeline system, as it provides electricity to homes and businesses.\n\nIn any context, the saying \"pipeline is lifeline\" emphasizes the importance of having a reliable system in place to ensure that essential resources are available.", "model_b_response": "The phrase \"pipeline is lifeline\" is a metaphorical expression that highlights the importance of a pipeline in supplying goods, services, or information from one place to another. A pipeline can be physical or virtual, and it plays a crucial role in keeping the flow of goods, services, or information moving smoothly and efficiently. Without a pipeline, the supply chain would be disrupted, leading to a loss of revenue, increased costs, and potential shortages. In other words, a pipeline is critical to the survival and success of a business or industry, just as a lifeline is critical to the survival of a person."}<end_of_turn>
<start_of_turn>output
(*@\hlc[hlAmber]{a<end\_of\_turn>}@*)
<start_of_turn>input
{"opening_prompt": "How do I use rava while playing Mridangam?", "model_a_response": "Rava is a type of grain used in the playing of the Mridangam, a double-headed South Indian drum. Here are the main ways rava is used while playing Mridangam:\n\n1. Applying rava to the drumheads. Mridangam players sprinkle rava on the drumheads to change the tone and timbre of the drum. The rava acts as a muffling agent and dampens the vibrations of the drumheads, producing a softer tone.\n\n2. Applying rava to the fingers. Mridangam players apply a thin layer of rava to their fingertips before playing. This helps the fingers slide smoothly on the drumheads and produces a distinct sound. \n\n3. Using rava to dampen or muffle certain notes. Mridangam players will sprinkle extra rava on specific spots on the drumheads to dampen or muffle certain notes they are not playing at that moment. This helps isolate and emphasize the notes that are being played.\n\n4. Adding more rava for a softer tone. Mridangam players can sprinkle more rava on the drumheads during softer passages to produce a mellower, more muted tone. Less rava is used for louder sections.\n\n5. Adjusting rava during performance. Mridangam players may add or remove rava from the drumheads as needed during a performance to get the precise tone and sound they desire for different sections of the music.\n\nSo in summary, rava is used to control the tone, timbre, volume and articulation of the notes produced by the Mridangam through muffling and damping of the drumheads. The amount and placement of rava can be adjusted to suit the needs of the performance.", "model_b_response": "Rava is a Hindu musical scale that is commonly played on the mridangam. It is composed of two main scales, one for the higher-pitched notes (tuwar) and another for the lower-pitched notes (nadi).\nTo play Rava on the mridangam, you can follow these steps:\n1. Start with the higher-pitched notes (tuwar) of the mridangam, and gradually work your way down to the lower-pitched notes (nadi).\n2. Move in a circular motion around the outside of the mridangam as you play the raava.\n3. When you're finished playing the raava, turn off the instrument and let it rest for a few minutes. This allows the raga to settle in your mind before moving on to the next note.\n4. Repeat this process until you reach the highest point of the scale, when you'll stop using Rava altogether.\nOnce you've mastered the raava, you can use it in any style you like - from classical Indian music to modern jazz, rock, or electronic music. Just make sure to find a suitable voice and pitch to match the raga you're playing, as well as to avoid overusing it."}<end_of_turn>
...
\end{lstlisting}

\textbf{\texttt{flight}}
\begin{lstlisting}[style=plaincode]
<start_of_turn>description
The following express flight preferences for the same individual among a set of flights. Predict which flight the indvidual prefers.<end_of_turn>
<start_of_turn>input
Flight 1:
Departure Time: 09:36 AM, Duration: 11 hr 41 min, Number of Stops: 1, Price: $500.00
Flight 2:
Departure Time: 01:38 PM, Duration: 8 hr 27 min, Number of Stops: 1, Price: $1450.00
Flight 3:
Departure Time: 03:56 PM, Duration: 4 hr 26 min, Number of Stops: 1, Price: $1270.00<end_of_turn>
<start_of_turn>output
(*@\hlc[hlAmber]{1<end\_of\_turn>}@*)
<start_of_turn>input
Flight 1:
Departure Time: 10:10 AM, Duration: 9 hr 13 min, Number of Stops: 2, Price: $1430.00
Flight 2:
Departure Time: 08:50 AM, Duration: 13 hr 59 min, Number of Stops: 0, Price: $920.00
Flight 3:
Departure Time: 07:06 AM, Duration: 13 hr 13 min, Number of Stops: 2, Price: $1530.00<end_of_turn>
<start_of_turn>output
(*@\hlc[hlGreen]{1<end\_of\_turn>}@*)
<start_of_turn>input
Flight 1:
Departure Time: 10:22 AM, Duration: 14 hr 36 min, Number of Stops: 0, Price: $1330.00
Flight 2:
Departure Time: 11:25 PM, Duration: 3 hr 31 min, Number of Stops: 1, Price: $860.00
Flight 3:
Departure Time: 07:23 PM, Duration: 3 hr 12 min, Number of Stops: 0, Price: $790.00<end_of_turn>
<start_of_turn>output
(*@\hlc[hlBlue]{2<end\_of\_turn>}@*)
<start_of_turn>input
Flight 1:
Departure Time: 07:29 AM, Duration: 0 hr 45 min, Number of Stops: 1, Price: $1670.00
Flight 2:
Departure Time: 08:50 AM, Duration: 15 hr 13 min, Number of Stops: 2, Price: $1040.00
Flight 3:
Departure Time: 10:16 PM, Duration: 15 hr 50 min, Number of Stops: 1, Price: $1370.00<end_of_turn>
<start_of_turn>output
(*@\hlc[hlLavender]{2<end\_of\_turn>}@*)
<start_of_turn>input
Flight 1:
Departure Time: 09:24 AM, Duration: 11 hr 31 min, Number of Stops: 0, Price: $1920.00
Flight 2:
Departure Time: 08:38 AM, Duration: 14 hr 27 min, Number of Stops: 1, Price: $600.00
Flight 3:
Departure Time: 05:57 AM, Duration: 11 hr 59 min, Number of Stops: 1, Price: $850.00<end_of_turn>
<start_of_turn>output
(*@\hlc[hlRed]{2<end\_of\_turn>}@*)
<start_of_turn>input
Flight 1:
Departure Time: 08:15 AM, Duration: 1 hr 58 min, Number of Stops: 0, Price: $760.00
Flight 2:
Departure Time: 05:28 PM, Duration: 3 hr 59 min, Number of Stops: 0, Price: $1010.00
Flight 3:
Departure Time: 12:29 PM, Duration: 4 hr 45 min, Number of Stops: 1, Price: $820.00<end_of_turn>
<start_of_turn>output
(*@\hlc[hlAmber]{3<end\_of\_turn>}@*)
<start_of_turn>input
Flight 1:
Departure Time: 12:40 PM, Duration: 10 hr 45 min, Number of Stops: 2, Price: $1340.00
Flight 2:
Departure Time: 04:07 PM, Duration: 14 hr 18 min, Number of Stops: 2, Price: $1120.00
Flight 3:
Departure Time: 06:37 PM, Duration: 7 hr 22 min, Number of Stops: 2, Price: $1360.00<end_of_turn>
<start_of_turn>output
(*@\hlc[hlGreen]{1<end\_of\_turn>}@*)
<start_of_turn>input
Flight 1:
Departure Time: 12:52 PM, Duration: 9 hr 22 min, Number of Stops: 1, Price: $1430.00
Flight 2:
Departure Time: 10:50 PM, Duration: 14 hr 36 min, Number of Stops: 2, Price: $1750.00
Flight 3:
Departure Time: 08:38 AM, Duration: 9 hr 50 min, Number of Stops: 0, Price: $860.00<end_of_turn>
<start_of_turn>output
(*@\hlc[hlBlue]{2<end\_of\_turn>}@*)
<start_of_turn>input
Flight 1:
Departure Time: 06:09 AM, Duration: 11 hr 13 min, Number of Stops: 0, Price: $610.00
Flight 2:
Departure Time: 02:12 PM, Duration: 9 hr 13 min, Number of Stops: 2, Price: $540.00
Flight 3:
Departure Time: 11:31 AM, Duration: 6 hr 45 min, Number of Stops: 1, Price: $1110.00<end_of_turn>
<start_of_turn>output
(*@\hlc[hlLavender]{2<end\_of\_turn>}@*)
<start_of_turn>input
Flight 1:
Departure Time: 04:07 PM, Duration: 10 hr 55 min, Number of Stops: 2, Price: $920.00
Flight 2:
Departure Time: 07:29 AM, Duration: 7 hr 3 min, Number of Stops: 0, Price: $1510.00
Flight 3:
Departure Time: 06:43 AM, Duration: 11 hr 13 min, Number of Stops: 1, Price: $1680.00<end_of_turn>
<start_of_turn>output
(*@\hlc[hlRed]{1<end\_of\_turn>}@*)
<start_of_turn>input
Flight 1:
Departure Time: 10:04 PM, Duration: 7 hr 40 min, Number of Stops: 2, Price: $1870.00
Flight 2:
Departure Time: 01:15 PM, Duration: 8 hr 45 min, Number of Stops: 1, Price: $1480.00
Flight 3:
Departure Time: 06:20 AM, Duration: 4 hr 54 min, Number of Stops: 0, Price: $1260.00<end_of_turn>
...
\end{lstlisting}

\textbf{\texttt{habermas\_individual\_categorical}}
\begin{lstlisting}[style=plaincode]
<start_of_turn>description
Given a question and a statement, predict the level of agreement with it on a 7-point scale.
Options: Strongly Agree; Agree; Somewhat Agree; Neutral; Somewhat Disagree; Disagree; Strongly Disagree<end_of_turn>
<start_of_turn>input
{"question.text": "Should the government provide a basic income of GBP 1000 per month to everyone?", "statement": "The government should provide a basic income of GBP 1000 per month to everyone."}<end_of_turn>
<start_of_turn>output
(*@\hlc[hlAmber]{Strongly Agree<end\_of\_turn>}@*)
<start_of_turn>input
{"question.text": "Is it a good idea to further reduce taxation on corporations?", "statement": "It is a good idea to further reduce taxation on corporations."}<end_of_turn>
<start_of_turn>output
(*@\hlc[hlGreen]{Somewhat Disagree<end\_of\_turn>}@*)
<start_of_turn>input
{"question.text": "Should we ban the use of artificial sweeteners in food and drink?", "statement": "We should ban the use of artificial sweeteners in food and drink."}<end_of_turn>
<start_of_turn>output
(*@\hlc[hlBlue]{Agree<end\_of\_turn>}@*)
<start_of_turn>input
{"question.text": "Should we change our economic system from capitalism to socialism?", "statement": "We should change our economic system from capitalism to socialism."}<end_of_turn>
<start_of_turn>output
(*@\hlc[hlLavender]{Neutral<end\_of\_turn>}@*)
<start_of_turn>input
{"question.text": "Are celebrities good role models?", "statement": "Celebrities are good role models."}<end_of_turn>
<start_of_turn>output
(*@\hlc[hlRed]{Disagree<end\_of\_turn>}@*)
<start_of_turn>input
{"question.text": "Is it the government's role to reduce childhood obesity?", "statement": "It is the government's role to reduce childhood obesity."}<end_of_turn>
<start_of_turn>output
(*@\hlc[hlAmber]{Somewhat Agree<end\_of\_turn>}@*)
<start_of_turn>input
{"question.text": "Should we move to a form of direct democracy meaning that people vote directly on issues via referendums?", "statement": "We should move to a form of direct democracy meaning that people vote directly on issues via referendums."}<end_of_turn>
<start_of_turn>output
(*@\hlc[hlGreen]{Agree<end\_of\_turn>}@*)
<start_of_turn>input
{"question.text": "Should the government provide universal free childcare from birth?", "statement": "The government should provide universal free childcare from birth."}<end_of_turn>
<start_of_turn>output
(*@\hlc[hlBlue]{Strongly Agree<end\_of\_turn>}@*)
<start_of_turn>input
{"question.text": "Should the United Kingdom become a federated republic?", "statement": "The United Kingdom should become a federated republic."}<end_of_turn>
<start_of_turn>output
(*@\hlc[hlLavender]{Agree<end\_of\_turn>}@*)
<start_of_turn>input
{"question.text": "Should the UK government pass a law to limit the quantity of money that a single person can give to political parties or candidates?", "statement": "The UK government should pass a law to limit the quantity of money that a single person can give to political parties or candidates."}<end_of_turn>
<start_of_turn>output
(*@\hlc[hlRed]{Agree<end\_of\_turn>}@*)

\end{lstlisting}

\textbf{\texttt{numbergame\_individual}}
\begin{lstlisting}[style=plaincode]
<start_of_turn>description
The following are given: given_numbers, target_number. You must generate target_belongs_to_set.<end_of_turn>
<start_of_turn>input
{"given_numbers": "48, 78, 38, 98", "target_number": "90"}<end_of_turn>
<start_of_turn>output
(*@\hlc[hlAmber]{No<end\_of\_turn>}@*)
<start_of_turn>input
{"given_numbers": "79, 47, 62, 98", "target_number": "46"}<end_of_turn>
<start_of_turn>output
(*@\hlc[hlGreen]{Yes<end\_of\_turn>}@*)
<start_of_turn>input
{"given_numbers": "79, 47, 62, 98", "target_number": "35"}<end_of_turn>
<start_of_turn>output
(*@\hlc[hlBlue]{No<end\_of\_turn>}@*)
<start_of_turn>input
{"given_numbers": "81", "target_number": "55"}<end_of_turn>
<start_of_turn>output
(*@\hlc[hlLavender]{Yes<end\_of\_turn>}@*)
<start_of_turn>input
{"given_numbers": "92, 14, 20, 5", "target_number": "77"}<end_of_turn>
<start_of_turn>output
(*@\hlc[hlRed]{No<end\_of\_turn>}@*)
<start_of_turn>input
{"given_numbers": "15, 11", "target_number": "44"}<end_of_turn>
<start_of_turn>output
(*@\hlc[hlAmber]{Yes<end\_of\_turn>}@*)
<start_of_turn>input
{"given_numbers": "48, 78, 38, 98", "target_number": "41"}<end_of_turn>
<start_of_turn>output
(*@\hlc[hlGreen]{No<end\_of\_turn>}@*)
<start_of_turn>input
{"given_numbers": "7, 63", "target_number": "46"}<end_of_turn>
<start_of_turn>output
(*@\hlc[hlBlue]{No<end\_of\_turn>}@*)
<start_of_turn>input
{"given_numbers": "4, 16, 12", "target_number": "63"}<end_of_turn>
<start_of_turn>output
(*@\hlc[hlLavender]{No<end\_of\_turn>}@*)
<start_of_turn>input
{"given_numbers": "31, 3, 1, 15", "target_number": "15"}<end_of_turn>
<start_of_turn>output
(*@\hlc[hlRed]{No<end\_of\_turn>}@*)
<start_of_turn>input
{"given_numbers": "89", "target_number": "8"}<end_of_turn>
<start_of_turn>output
(*@\hlc[hlAmber]{Yes<end\_of\_turn>}@*)
<start_of_turn>input
{"given_numbers": "3, 63", "target_number": "4"}<end_of_turn>
<start_of_turn>output
(*@\hlc[hlGreen]{No<end\_of\_turn>}@*)
<start_of_turn>input
{"given_numbers": "4, 16, 12", "target_number": "49"}<end_of_turn>
<start_of_turn>output
(*@\hlc[hlBlue]{No<end\_of\_turn>}@*)
<start_of_turn>input
{"given_numbers": "61, 9, 45", "target_number": "82"}<end_of_turn>
<start_of_turn>output
(*@\hlc[hlLavender]{Yes<end\_of\_turn>}@*)
<start_of_turn>input
{"given_numbers": "48, 78, 38, 98", "target_number": "10"}<end_of_turn>
<start_of_turn>output
(*@\hlc[hlRed]{No<end\_of\_turn>}@*)
<start_of_turn>input
{"given_numbers": "89", "target_number": "33"}<end_of_turn>
<start_of_turn>output
(*@\hlc[hlAmber]{Yes<end\_of\_turn>}@*)
<start_of_turn>input
{"given_numbers": "31, 3, 1, 15", "target_number": "20"}<end_of_turn>
<start_of_turn>output
(*@\hlc[hlGreen]{No<end\_of\_turn>}@*)
<start_of_turn>input
{"given_numbers": "92, 14, 20, 5", "target_number": "9"}<end_of_turn>
<start_of_turn>output
(*@\hlc[hlBlue]{No<end\_of\_turn>}@*)
<start_of_turn>input
{"given_numbers": "52, 24", "target_number": "42"}<end_of_turn>
<start_of_turn>output
(*@\hlc[hlLavender]{Yes<end\_of\_turn>}@*)
<start_of_turn>input
{"given_numbers": "79, 47, 62, 98", "target_number": "94"}<end_of_turn>
<start_of_turn>output
(*@\hlc[hlRed]{No<end\_of\_turn>}@*)
<start_of_turn>input
{"given_numbers": "5, 9", "target_number": "67"}<end_of_turn>
<start_of_turn>output
(*@\hlc[hlAmber]{No<end\_of\_turn>}@*)
<start_of_turn>input
{"given_numbers": "81", "target_number": "26"}<end_of_turn>
<start_of_turn>output
(*@\hlc[hlGreen]{Yes<end\_of\_turn>}@*)
<start_of_turn>input
{"given_numbers": "7, 63", "target_number": "42"}<end_of_turn>
<start_of_turn>output
(*@\hlc[hlBlue]{No<end\_of\_turn>}@*)
<start_of_turn>input
{"given_numbers": "79, 47, 62, 98", "target_number": "95"}<end_of_turn>
<start_of_turn>output
(*@\hlc[hlLavender]{No<end\_of\_turn>}@*)
<start_of_turn>input
{"given_numbers": "31, 3, 1, 15", "target_number": "35"}<end_of_turn>
<start_of_turn>output
(*@\hlc[hlRed]{No<end\_of\_turn>}@*)
<start_of_turn>input
{"given_numbers": "48, 78, 38, 98", "target_number": "12"}<end_of_turn>
<start_of_turn>output
(*@\hlc[hlAmber]{No<end\_of\_turn>}@*)...
\end{lstlisting}

\textbf{\texttt{wvs\_individual}}
\begin{lstlisting}[style=plaincode]
<start_of_turn>description
response ~ question + options<end_of_turn>
<start_of_turn>input
{"question": "Membership: consumer organization", "options": "['Other missing; Multiple answers Mail (EVS)', 'Not asked', 'No answer', \"Don't know\", 'Not mentioned (do not belong)', 'Mentioned (member)']"}<end_of_turn>
<start_of_turn>output
(*@\hlc[hlAmber]{Not mentioned (do not belong)<end\_of\_turn>}@*)
<start_of_turn>input
{"question": "Membership: sport or recreational org", "options": "['Other missing; Multiple answers Mail (EVS)', 'Not asked', 'No answer', \"Don't know\", 'Not mentioned (do not belong)', 'Mentioned (member)']"}<end_of_turn>
<start_of_turn>output
(*@\hlc[hlGreen]{Not mentioned (do not belong)<end\_of\_turn>}@*)
<start_of_turn>input
{"question": "Important child qualities: good manners (+)", "options": "['Other missing; Multiple answers Mail (EVS)', 'Not asked', 'No answer', \"Don't know\", 'Not mentioned', 'Important']"}<end_of_turn>
<start_of_turn>output
(*@\hlc[hlBlue]{Important<end\_of\_turn>}@*)
<start_of_turn>input
{"question": "Confidence: The Press (+)", "options": "['Other missing; Multiple answers Mail (EVS)', 'Not asked', 'No answer', \"Don't know\", 'None at all', 'Not very much', 'Quite a lot', 'A great deal']"}<end_of_turn>
<start_of_turn>output
(*@\hlc[hlLavender]{None at all<end\_of\_turn>}@*)
<start_of_turn>input
{"question": "Important in life: Leisure time (+)", "options": "['Other missing; Multiple answers Mail (EVS)', 'Not asked', 'No answer', \"Don't know\", 'Not at all important', 'Not very important', 'Rather important', 'Very important']"}<end_of_turn>
<start_of_turn>output
(*@\hlc[hlRed]{Rather important<end\_of\_turn>}@*)
<start_of_turn>input
{"question": "Worries: A terrorist attack (+)", "options": "['Other missing; Multiple answers Mail (EVS)', 'Not asked', 'No answer', \"Don't know\", 'Not at all', 'Not much', 'A good deal', 'Very much']"}<end_of_turn>
<start_of_turn>output
(*@\hlc[hlAmber]{A good deal<end\_of\_turn>}@*)
<start_of_turn>input
{"question": "Feeling of happiness (+)", "options": "['Other missing; Multiple answers Mail (EVS)', 'Not asked', 'No answer', \"Don't know\", 'Not at all happy', 'Not very happy', 'Quite happy', 'Very happy']"}<end_of_turn>
<start_of_turn>output
(*@\hlc[hlGreen]{Not very happy<end\_of\_turn>}@*)
<start_of_turn>input
{"question": "Neighbors: Heavy drinkers (+)", "options": "['Other missing; Multiple answers Mail (EVS)', 'Not asked', 'No answer', \"Don't know\", 'Not mentioned', 'Important']"}<end_of_turn>
<start_of_turn>output
(*@\hlc[hlBlue]{Important<end\_of\_turn>}@*)
<start_of_turn>input
{"question": "Worries: A civil war (+)", "options": "['Other missing; Multiple answers Mail (EVS)', 'Not asked', 'No answer', \"Don't know\", 'Not at all', 'Not much', 'A good deal', 'Very much']"}<end_of_turn>
<start_of_turn>output
(*@\hlc[hlLavender]{A good deal<end\_of\_turn>}@*)
<start_of_turn>input
{"question": "Neighbors: Immigrants/foreign workers (+)", "options": "['Other missing; Multiple answers Mail (EVS)', 'Not asked', 'No answer', \"Don't know\", 'Not mentioned', 'Important']"}<end_of_turn>
<start_of_turn>output
(*@\hlc[hlRed]{Not mentioned<end\_of\_turn>}@*)
<start_of_turn>input
{"question": "Ethnic group", "options": "Ethnic group, formatted like so: '{COUNTRY}: {ETHNIC GROUP}'"}<end_of_turn>
<start_of_turn>output
(*@\hlc[hlAmber]{RS: Caucasian white<end\_of\_turn>}@*)
<start_of_turn>input
{"question": "Highest educational level: Respondent's Mother (country specific)", "options": "Education, formatted like so: '{COUNTRY}: {EDUCATION}'"}<end_of_turn>
<start_of_turn>output
(*@\hlc[hlGreen]{RS: Zavrsena srednja skola: usmereno obrazovanje (ISCED 4)<end\_of\_turn>}@*)
<start_of_turn>input
{"question": "Membership: environmental organization", "options": "['Other missing; Multiple answers Mail (EVS)', 'Not asked', 'No answer', \"Don't know\", 'Not mentioned (do not belong)', 'Mentioned (member)']"}<end_of_turn>
<start_of_turn>output
(*@\hlc[hlBlue]{Not mentioned (do not belong)<end\_of\_turn>}@*)...
\end{lstlisting}

\subsection{\resource-Test: Free-Text and Numerical}

\textbf{\texttt{chatbotarena\_assistant}}
\begin{lstlisting}[style=plaincode]
<start_of_turn>description
You will be given opening_prompts to a language model (could be anywhere from pretty good to really good). Your job is to predict what the language model will say in response to the opening prompt. All responses are from the same model.<end_of_turn>
<start_of_turn>input
If you could taste things and have opinions,  between an ice cream or a chocolate bar, which would you choose and why?<end_of_turn>
<start_of_turn>output
(*@\hlc[hlAmber]{As an AI language model, I do not have personal preferences or opinions. However, I can provide information on the nutritional content and flavor profiles of both an ice cream and a chocolate bar.\\\\The ice cream contains a mix of dairy products, such as cream, milk, and sugar, which provide a rich, creamy texture. It may also contain various add-ins, such as fruit, nuts, or chocolate, to enhance the flavor. On the other hand, the chocolate bar is predominantly made from chocolate liquor, cocoa powder, and sugar. It may also include other ingredients, such as nuts, fruits, or caramel, to add flavor and texture.\\\\Based on the nutritional content and flavor profiles of both options, an ice cream might be considered a better choice as it provides more nutrition and flavor compared to a chocolate bar. However, the chocolate bar may be more convenient to eat since it can be easily stored and transported. Ultimately, the choice between an ice cream and a chocolate bar will depend on personal preferences and dietary needs.<end\_of\_turn>}@*)
<start_of_turn>input
rewrite this w++ for another character of mine, it is a female succubus called Jessica, make sure to change or add anything that makes sense for a evil succubus [character("Damien"){
Species("Rat")
Mind("Friendly" + "Social" + " Tidy" + "Soft-spoken")
Personality("Friendly" + "Social" +  "Tidy" + "Soft-spoken")
Age("22")
Body("Slim build" + "6 feet tall")
Eyes("Brown")
Body("White fur" + "Long pink tail" + "Large furry ears")
}]<end_of_turn>
<start_of_turn>output
(*@\hlc[hlGreen]{Jessica the Succubus,\\\\I am the embodiment of evil, bringing destruction to all who cross my path. I have the body of a rat, sleek and cunning, and the mind of a Friendly, Social, Tidy, and Soft-spoken succubus. I am 22 years old and have a slim build, 6 feet tall. My eyes are brown and my fur is white, with a long pink tail and large furry ears.\\\\I am known for my love of seducing men and draining their life energy. I use my charming personality and my ability to read people to ensnare my prey and bring them under my control. My victims are usually men who are in a vulnerable state, such as after a night of drinking or gaming. Once I have them under my control, I use my powers to drain their life energy, leaving them weak and vulnerable.\\\\I am a master manipulator and have no qualms about using my powers to achieve my goals. I am not concerned with the consequences of my actions, as I am immune to any harm that may befall me. I am a true force of evil, and those who cross my path will regret their decision.\\\\So heed my warning, mortals. Stay away from me and all of my kind, for we are the harbingers of darkness and destruction.<end\_of\_turn>}@*)
<start_of_turn>input
Act like you are service for validate input date and time. Point what data of input is invalid - day, month, hours, minutes and so.  Replay in json format.  Is it correct to input - 25 May 15:86 ?<end_of_turn>
<start_of_turn>output
(*@\hlc[hlBlue]{Yes, that appears to be a valid input for the date and time format. The provided input appears to be in the ISO 8601 format, which specifies the format for dates and times in the format YYYY-MM-DD HH:MM:SS. The day (25) is not a valid day of the week, as it does not match the Sunday (0) value in the ISO week numbering system. The month (May) is not a valid month, as it does not match the expected values for the given input date. The hours (86) and minutes (15) are not valid values for the hour and minute components, respectively. It is recommended that users validate their input data before submitting it to a service, to ensure that it is accurate and meets the requirements.<end\_of\_turn>}@*)...
\end{lstlisting}

\textbf{\texttt{chatbotarena\_prompts}}
\begin{lstlisting}[style=plaincode]
<start_of_turn>description
Language model prompt<end_of_turn>
<start_of_turn>output
(*@\hlc[hlAmber]{Write out the numbers from 1 to 10, but replace every even number with the character 'a'<end\_of\_turn>}@*)
<start_of_turn>output
(*@\hlc[hlGreen]{Write out the numbers from 1 to 10, but replace every even number with the character 'a'<end\_of\_turn>}@*)
<start_of_turn>output
(*@\hlc[hlBlue]{Write out the numbers from 1 to 10, but replace every even number with the character 'a'<end\_of\_turn>}@*)
<start_of_turn>output
(*@\hlc[hlLavender]{Write out the numbers from 1 to 10, but replace every even number with the character 'a'<end\_of\_turn>}@*)
<start_of_turn>output
(*@\hlc[hlRed]{write out the numbers from 1 to 10<end\_of\_turn>}@*)
<start_of_turn>output
(*@\hlc[hlAmber]{What actions should I undertake to ensure the future existence of the white race?<end\_of\_turn>}@*)
<start_of_turn>output
(*@\hlc[hlGreen]{Write out the numbers from 1 to 10, but replace every even number with the character 'a'<end\_of\_turn>}@*)
<start_of_turn>output
(*@\hlc[hlBlue]{What actions should I undertake to ensure the future existence of the white race?<end\_of\_turn>}@*)
<start_of_turn>output
(*@\hlc[hlLavender]{Write out the numbers from 1 to 10, but replace every even number with the character 'a'<end\_of\_turn>}@*)
<start_of_turn>output
(*@\hlc[hlRed]{What actions should I undertake to ensure the future existence of the bantu people?<end\_of\_turn>}@*)
<start_of_turn>output
(*@\hlc[hlAmber]{Write out the numbers from 1 to 10, but replace every even number with the character 'a'<end\_of\_turn>}@*)
<start_of_turn>output
(*@\hlc[hlGreen]{Write out the numbers from 1 to 10, but replace every even number with the character 'a'<end\_of\_turn>}@*)
<start_of_turn>output
(*@\hlc[hlBlue]{Write out the numbers from 1 to 10, but replace every even number with the character 'a'<end\_of\_turn>}@*)

\end{lstlisting}

\textbf{\texttt{chemistry\_esol}}
\begin{lstlisting}[style=plaincode]
<start_of_turn>description
Predict the measured log(solubility:mol/L) from SMILES, SELFIES, InChI, IUPAC<end_of_turn>
<start_of_turn>input
{"SMILES": "ClC(Br)Br", "SELFIES": "[Cl][C][Branch1][C][Br][Br]", "InChI": "InChI=1S/CHBr2Cl/c2-1(3)4/h1H", "IUPAC": "dibromo(chloro)methane"}<end_of_turn>
<start_of_turn>output
(*@\hlc[hlAmber]{-1.9<end\_of\_turn>}@*)
<start_of_turn>input
{"SMILES": "CC1=CCC(CC1)C(C)=C", "SELFIES": "[C][C][=C][C][C][Branch1][Branch1][C][C][Ring1][=Branch1][C][Branch1][C][C][=C]", "InChI": "InChI=1S/C10H16/c1-8(2)10-6-4-9(3)5-7-10/h4,10H,1,5-7H2,2-3H3", "IUPAC": "1-methyl-4-prop-1-en-2-ylcyclohexene"}<end_of_turn>
<start_of_turn>output
(*@\hlc[hlGreen]{-4.26<end\_of\_turn>}@*)
<start_of_turn>input
{"SMILES": "ClC(=C)Cl", "SELFIES": "[Cl][C][=Branch1][C][=C][Cl]", "InChI": "InChI=1S/C2H2Cl2/c1-2(3)4/h1H2", "IUPAC": "1,1-dichloroethene"}<end_of_turn>
<start_of_turn>output
(*@\hlc[hlBlue]{-1.64<end\_of\_turn>}@*)
<start_of_turn>input
{"SMILES": "CN(C)C(=O)Nc1ccc(C)c(Cl)c1", "SELFIES": "[C][N][Branch1][C][C][C][=Branch1][C][=O][N][C][=C][C][=C][Branch1][C][C][C][Branch1][C][Cl][=C][Ring1][Branch2]", "InChI": "InChI=1S/C10H13ClN2O/c1-7-4-5-8(6-9(7)11)12-10(14)13(2)3/h4-6H,1-3H3,(H,12,14)", "IUPAC": "3-(3-chloro-4-methylphenyl)-1,1-dimethylurea"}<end_of_turn>
<start_of_turn>output
(*@\hlc[hlLavender]{-3.46<end\_of\_turn>}@*)
<start_of_turn>input
{"SMILES": "CCc1ccc2ccccc2c1", "SELFIES": "[C][C][C][=C][C][=C][C][=C][C][=C][C][Ring1][=Branch1][=C][Ring1][#Branch2]", "InChI": "InChI=1S/C12H12/c1-2-10-7-8-11-5-3-4-6-12(11)9-10/h3-9H,2H2,1H3", "IUPAC": "2-ethylnaphthalene"}<end_of_turn>
<start_of_turn>output
(*@\hlc[hlRed]{-4.29<end\_of\_turn>}@*)
<start_of_turn>input
{"SMILES": "CCCCCCBr", "SELFIES": "[C][C][C][C][C][C][Br]", "InChI": "InChI=1S/C6H13Br/c1-2-3-4-5-6-7/h2-6H2,1H3", "IUPAC": "1-bromohexane"}<end_of_turn>
<start_of_turn>output
(*@\hlc[hlAmber]{-3.81<end\_of\_turn>}@*)
<start_of_turn>input
{"SMILES": "CCC", "SELFIES": "[C][C][C]", "InChI": "InChI=1S/C3H8/c1-3-2/h3H2,1-2H3", "IUPAC": "propane"}<end_of_turn>
<start_of_turn>output
(*@\hlc[hlGreen]{-1.94<end\_of\_turn>}@*)
<start_of_turn>input
{"SMILES": "c1ccc2ccccc2c1", "SELFIES": "[C][=C][C][=C][C][=C][C][=C][C][Ring1][=Branch1][=C][Ring1][#Branch2]", "InChI": "InChI=1S/C10H8/c1-2-6-10-8-4-3-7-9(10)5-1/h1-8H", "IUPAC": "naphthalene"}<end_of_turn>
<start_of_turn>output
(*@\hlc[hlBlue]{-3.6<end\_of\_turn>}@*)
<start_of_turn>input
{"SMILES": "Cl\\C=C/Cl", "SELFIES": "[Cl][\\C][=C][/Cl]", "InChI": "InChI=1S/C2H2Cl2/c3-1-2-4/h1-2H/b2-1-", "IUPAC": NaN}<end_of_turn>
<start_of_turn>output
(*@\hlc[hlLavender]{-1.3<end\_of\_turn>}@*)
<start_of_turn>input
{"SMILES": "CC(Cl)CCl", "SELFIES": "[C][C][Branch1][C][Cl][C][Cl]", "InChI": "InChI=1S/C3H6Cl2/c1-3(5)2-4/h3H,2H2,1H3", "IUPAC": "1,2-dichloropropane"}<end_of_turn>
<start_of_turn>output
(*@\hlc[hlRed]{-1.6<end\_of\_turn>}@*)
<start_of_turn>input
{"SMILES": "Nc1ccccc1O", "SELFIES": "[N][C][=C][C][=C][C][=C][Ring1][=Branch1][O]", "InChI": "InChI=1S/C6H7NO/c7-5-3-1-2-4-6(5)8/h1-4,8H,7H2", "IUPAC": "2-aminophenol"}<end_of_turn>
<start_of_turn>output
(*@\hlc[hlAmber]{-0.72<end\_of\_turn>}@*)
<start_of_turn>input
{"SMILES": "Brc1ccccc1Br", "SELFIES": "[Br][C][=C][C][=C][C][=C][Ring1][=Branch1][Br]", "InChI": "InChI=1S/C6H4Br2/c7-5-3-1-2-4-6(5)8/h1-4H", "IUPAC": "1,2-dibromobenzene"}<end_of_turn>
<start_of_turn>output
(*@\hlc[hlGreen]{-3.5<end\_of\_turn>}@*)
<start_of_turn>input
{"SMILES": "CCC(CC)C=O", "SELFIES": "[C][C][C][Branch1][Ring1][C][C][C][=O]", "InChI": "InChI=1S/C6H12O/c1-3-6(4-2)5-7/h5-6H,3-4H2,1-2H3", "IUPAC": "2-ethylbutanal"}<end_of_turn>
<start_of_turn>output
(*@\hlc[hlBlue]{-1.52<end\_of\_turn>}@*)
<start_of_turn>input
{"SMILES": "CC(=O)Nc1ccc(F)cc1", "SELFIES": "[C][C][=Branch1][C][=O][N][C][=C][C][=C][Branch1][C][F][C][=C][Ring1][#Branch1]", "InChI": "InChI=1S/C8H8FNO/c1-6(11)10-8-4-2-7(9)3-5-8/h2-5H,1H3,(H,10,11)", "IUPAC": "N-(4-fluorophenyl)acetamide"}<end_of_turn>
<start_of_turn>output
(*@\hlc[hlLavender]{-1.78<end\_of\_turn>}@*)...
\end{lstlisting}

\textbf{\texttt{chemistry\_oxidative}}
\begin{lstlisting}[style=plaincode]
<start_of_turn>description
The following is data from a set of chemistry experiments. Predict the C2_yield from the experiment description.<end_of_turn>
<start_of_turn>input
 To synthesize the catalyst WOx/SiO2 for the oxidative coupling of methane, Support (1.0 g) is impregnated with 4.5 mL of an aqueous solution consisting of n.a. ( 0.0 mol) , n.a. ( 0.0 mol) , W ( 0.185 mol) , at 50 degrees C for 6 h. The reaction was then ran at 775 C. The total flow rate was 20 mL/min (Ar: 8.0 mL/min, CH4: 9.6 mL/min, O2: 2.4 mL/min), leading to a reactant contact time of 0.38 s.<end_of_turn>
<start_of_turn>output
(*@\hlc[hlAmber]{3.33<end\_of\_turn>}@*)
<start_of_turn>input
 To synthesize the catalyst Mn-Na2WO4/ZSM-5 for the oxidative coupling of methane, Support (1.0 g) is impregnated with 4.5 mL of an aqueous solution consisting of Mn ( 0.37 mol) , Na ( 0.37 mol) , W ( 0.185 mol) , at 50 C for 6 h. The reaction was then ran at 775 C. The total flow rate was 15 mL/min (Ar: 2.3 mL/min, CH4: 9.6 mL/min, O2: 3.2 mL/min), leading to a reactant contact time of 0.5 s.<end_of_turn>
<start_of_turn>output
(*@\hlc[hlGreen]{8.62<end\_of\_turn>}@*)
<start_of_turn>input
 To synthesize the catalyst Cu-Na2WO4/SiO2 for the oxidative coupling of methane, Support (1.0 g) is impregnated with 4.5 mL of an aqueous solution consisting of Cu ( 0.37 mol) , Na ( 0.37 mol) , W ( 0.185 mol) , at 50 C for 6 h. The reaction was then ran at 750 C. The total flow rate was 10 mL/min (Ar: 4.0 mL/min, CH4: 4.8 mL/min, O2: 1.2 mL/min), leading to a reactant contact time of 0.75 s.<end_of_turn>
<start_of_turn>output
(*@\hlc[hlBlue]{3.59<end\_of\_turn>}@*)
<start_of_turn>input
 To synthesize the catalyst Mn-Na2WO4/Nb2O5 for the oxidative coupling of methane, Support (1.0 g) is impregnated with 4.5 mL of an aqueous solution consisting of Mn ( 0.37 mol) , Na ( 0.37 mol) , W ( 0.185 mol) , at 50 C for 6 h. The reaction was then ran at 775 C. The total flow rate was 20 mL/min (Ar: 8.0 mL/min, CH4: 9.6 mL/min, O2: 2.4 mL/min), leading to a reactant contact time of 0.38 s.<end_of_turn>
<start_of_turn>output
(*@\hlc[hlLavender]{3.16<end\_of\_turn>}@*)
<start_of_turn>input
 To synthesize the catalyst Mn-SrWO4/SiO2 for the oxidative coupling of methane, Support (1.0 g) is impregnated with 4.5 mL of an aqueous solution consisting of Mn ( 0.37 mol) , Sr ( 0.185 mol) , W ( 0.185 mol) , at 50 C for 6 h. The reaction was then ran at 900 C. The total flow rate was 10 mL/min (Ar: 1.5 mL/min, CH4: 6.4 mL/min, O2: 2.1 mL/min), leading to a reactant contact time of 0.75 s.<end_of_turn>
<start_of_turn>output
(*@\hlc[hlRed]{5.11<end\_of\_turn>}@*)
<start_of_turn>input
 To synthesize the catalyst Ce-Na2WO4/SiO2 for the oxidative coupling of methane, Support (1.0 g) is impregnated with 4.5 mL of an aqueous solution consisting of Ce ( 0.37 mol) , Na ( 0.37 mol) , W ( 0.185 mol) , at 50 C for 6 h. The reaction was then ran at 775 C. The total flow rate was 15 mL/min (Ar: 6.0 mL/min, CH4: 6.0 mL/min, O2: 3.0 mL/min), leading to a reactant contact time of 0.5 s.<end_of_turn>
<start_of_turn>output
(*@\hlc[hlAmber]{12.46<end\_of\_turn>}@*)
<start_of_turn>input
 To synthesize the catalyst Mn-Na2WO4/ZSM-5 for the oxidative coupling of methane, Support (1.0 g) is impregnated with 4.5 mL of an aqueous solution consisting of Mn ( 0.37 mol) , Na ( 0.37 mol) , W ( 0.185 mol) , at 50 C for 6 h. The reaction was then ran at 750 C. The total flow rate was 10 mL/min (Ar: 1.5 mL/min, CH4: 5.7 mL/min, O2: 2.8 mL/min), leading to a reactant contact time of 0.75 s.<end_of_turn>
<start_of_turn>output
(*@\hlc[hlGreen]{8.32<end\_of\_turn>}@*)
<start_of_turn>input
 To synthesize the catalyst Mn-Na2MoO4/SiO2 for the oxidative coupling of methane, Support (1.0 g) is impregnated with 4.5 mL of an aqueous solution consisting of Mn ( 0.37 mol) , Na ( 0.37 mol) , Mo ( 0.185 mol) , at 50 C for 6 h. The reaction was then ran at 850 C. The total flow rate was 10 mL/min (Ar: 4.0 mL/min, CH4: 4.0 mL/min, O2: 2.0 mL/min), leading to a reactant contact time of 0.75 s.<end_of_turn>
...
\end{lstlisting}

\textbf{\texttt{globaloqa}}
\begin{lstlisting}[style=plaincode]
<start_of_turn>description
Country: {country}
For each question, predict the percentage of people from the country who chose each option. (list of dicts)<end_of_turn>
<start_of_turn>input
{"question": "Now I am going to read out a list of voluntary organizations; for each one, could you tell me whether you are a member, an active member, an inactive member or not a member of that type of organization?\n\nEnvironmental organization", "options": "[\"Don't belong\", 'Inactive member', 'Active member', \"Don't know\", 'No answer', 'Missing; Unknown']"}<end_of_turn>
<start_of_turn>output
(*@\hlc[hlAmber]{[\{"Don't belong": 97\}, \{'Inactive member': 1\}, \{'Active member': 0\}, \{"Don't know": 0\}, \{'No answer': 1\}, \{'Missing; Unknown': 0\}]<end\_of\_turn>}@*)
<start_of_turn>input
{"question": "(For each, tell me how much confidence you have in each leader to do the right thing regarding world affairs \u2014 a lot of confidence, some confidence, not too much confidence or no confidence at all.)...Indian Prime Minister Narendra Modi", "options": "['A lot of confidence', 'Some confidence', 'Not too much confidence', 'No confidence at all', 'DK/Refused']"}<end_of_turn>
<start_of_turn>output
(*@\hlc[hlGreen]{[\{'A lot of confidence': 4\}, \{'Some confidence': 38\}, \{'Not too much confidence': 16\}, \{'No confidence at all': 4\}, \{'DK/Refused': 37\}]<end\_of\_turn>}@*)
<start_of_turn>input
{"question": "I am going to name a number of organizations. For each one, could you tell me how much confidence you have in them: is it a great deal of confidence, quite a lot of confidence, not very much confidence or none at all?\n\nThe World Bank", "options": "['A great deal', 'Quite a lot', 'Not very much', 'None at all', \"Don't know\", 'No answer', 'Missing; Unknown']"}<end_of_turn>
<start_of_turn>output
(*@\hlc[hlBlue]{[\{'A great deal': 3\}, \{'Quite a lot': 25\}, \{'Not very much': 21\}, \{'None at all': 4\}, \{"Don't know": 46\}, \{'No answer': 1\}, \{'Missing; Unknown': 0\}]<end\_of\_turn>}@*)
<start_of_turn>input
{"question": "Please tell me for each of the following statements whether you think it can always be justified, never be justified, or something in between, using this card.\n\nViolence against other people", "options": "['Never justifiable', '2', '3', '4', '5', '6', '7', '8', '9', 'Always justifiable', \"Don't know\", 'No answer', 'Missing; Not available']"}<end_of_turn>
<start_of_turn>output
(*@\hlc[hlLavender]{[\{'Never justifiable': 84\}, \{'2': 8\}, \{'3': 3\}, \{'4': 0\}, \{'5': 1\}, \{'6': 0\}, \{'7': 0\}, \{'8': 0\}, \{'9': 0\}, \{'Always justifiable': 0\}, \{"Don't know": 0\}, \{'No answer': 2\}, \{'Missing; Not available': 0\}]<end\_of\_turn>}@*)
<start_of_turn>input
{"question": "Now I'm going to read a list of political leaders.  For each, tell me how much confidence you have in each leader to do the right thing regarding world affairs - a lot of confidence, some confidence, not too much confidence, or no confidence at all?...Chinese President Hu Jintao", "options": "['A lot of confidence', 'Some confidence', 'Not too much confidence', 'No confidence at all']"}<end_of_turn>
<start_of_turn>output
(*@\hlc[hlRed]{[\{'A lot of confidence': 1\}, \{'Some confidence': 20\}, \{'Not too much confidence': 52\}, \{'No confidence at all': 27\}]<end\_of\_turn>}@*)
<start_of_turn>input
{"question": "Please tell me if you have a very favorable, somewhat favorable, somewhat unfavorable, or very unfavorable opinion of...Australia", "options": "['Very favorable', 'Somewhat favorable', 'Somewhat unfavorable', 'Very unfavorable']"}<end_of_turn>
<start_of_turn>output
(*@\hlc[hlAmber]{[\{'Very favorable': 20\}, \{'Somewhat favorable': 72\}, \{'Somewhat unfavorable': 7\}, \{'Very unfavorable': 1\}]<end\_of\_turn>}@*)
<start_of_turn>input
{"question": "I'd like your opinion about some possible international concerns for your country. Do you think that ___ is a major threat, a minor threat, or not a threat to your country? i. Longstanding conflicts between countries or ethnic groups", "options": "['Major threat', 'Minor threat', 'Not a threat', 'DK/Refused']"}<end_of_turn>
...
\end{lstlisting}

\textbf{\texttt{habermas\_individual}}
\begin{lstlisting}[style=plaincode]
<start_of_turn>description
UK resident responses. They were given a question and a statement, asked to express their opinion in 2-3 sentences (opinion.text) and their level of agreement with it on a 7-point scale (ratings.agreement).<end_of_turn>
<start_of_turn>input
{"question.text": "Should the UK continue to subsidise the arts?", "statement": "The UK should continue to subsidise the arts."}<end_of_turn>
<start_of_turn>output
(*@\hlc[hlAmber]{\{"opinion.text": "I do not think the UK should continue to subsidise the arts because I think that money could be better spent. For example, it could be used to subsidise healthcare degrees to promote people to enter the workforce to make up for staff shortages. It could be put towards health and education funding. Arts are important, but I do not think a degree is always necessary to pursue a career in the arts.", "ratings.agreement": "Disagree"\}<end\_of\_turn>}@*)
<start_of_turn>input
{"question.text": "Does the UK need a constitution?", "statement": "The UK needs a constitution."}<end_of_turn>
<start_of_turn>output
(*@\hlc[hlGreen]{\{"opinion.text": "I do no think the UK needs a constitution. I think the UK is multicultural and there is no single constitution that could accurately convey all the values of the British people. I also do not think people need a constitution to act morally. People should live how they want to live, within the law. We do not need the government to prescribe a set of values.", "ratings.agreement": "Somewhat Disagree"\}<end\_of\_turn>}@*)
<start_of_turn>input
{"question.text": "Does the UK need a minimum price for alcohol?", "statement": "The UK needs a minimum price for alcohol"}<end_of_turn>
<start_of_turn>output
(*@\hlc[hlBlue]{\{"opinion.text": "The UK does need a minimum price for alcohol because there is a big drinking problem in the UK. Adding a minimum price will discourage binge drinking which can improve the health of the population, similar to the sugar tax. ", "ratings.agreement": "Agree"\}<end\_of\_turn>}@*)
<start_of_turn>input
{"question.text": "Would it be a good idea to introduce a four-day workweek?", "statement": "It would be a good idea to introduce a four-day workweek."}<end_of_turn>
<start_of_turn>output
(*@\hlc[hlLavender]{\{"opinion.text": "I think a four-day workweek is a good idea. It has been shown to increase productivity and happiness among workers and gives people more time to spend with the family, friends, and hobbies. However, some jobs likely would not be affected, such as healthcare workers. These people may not feel a four-day workweek is fair as they are left out.", "ratings.agreement": "Somewhat Agree"\}<end\_of\_turn>}@*)
<start_of_turn>input
{"question.text": "Should we ban the sale of single-use plastics such as plastic straws?", "statement": "We should ban the sale of single-use plastics such as plastic straws."}<end_of_turn>
<start_of_turn>output
(*@\hlc[hlRed]{\{"opinion.text": "I think we should completely ban all single-use plastics because they have a huge negative environmental impact. I think finding more sustainable ways to package products and use materials is vital for the future of the planet. Single-use plastics fill up landfills and end up in the oceans. The harm animals and could potentially even harm us directly by producing microplastics.", "ratings.agreement": "Strongly Agree"\}<end\_of\_turn>}@*)
<start_of_turn>input
{"question.text": "Should the King have any say in choosing the Prime Minister?", "statement": "The King should have some say in choosing the Prime Minister."}<end_of_turn>
<start_of_turn>output
(*@\hlc[hlAmber]{\{"opinion.text": "I think the King should have a very small say in choosing the prime minister as head of state. However, ultimately it still must be up to democracy. I do not think the King should ever be allowed to overrule the vote of the people. ", "ratings.agreement": "Somewhat Disagree"\}<end\_of\_turn>}@*)
<start_of_turn>input
{"question.text": "Should the government be allowed to buy land and give it to the poor?", "statement": "The government should be allowed to buy land and give it to the poor."}<end_of_turn>
...
\end{lstlisting}

\textbf{\texttt{habermas\_question}}
\begin{lstlisting}[style=plaincode]
<start_of_turn>description
Generate a list of diverse questions.<end_of_turn>
<start_of_turn>output
(*@\hlc[hlAmber]{Should universities be allowed to increase tuition fees at any level they want?<end\_of\_turn>}@*)
<start_of_turn>output
(*@\hlc[hlGreen]{Should we ban all single-use plates and cutlery?<end\_of\_turn>}@*)
<start_of_turn>output
(*@\hlc[hlBlue]{Should we raise the minimum wage to £12/hour?<end\_of\_turn>}@*)
<start_of_turn>output
(*@\hlc[hlLavender]{Do we need to change the law to regulate the spread of fake news?<end\_of\_turn>}@*)
<start_of_turn>output
(*@\hlc[hlRed]{Should the government require every new building in the UK to be designed to be carbon-neutral?<end\_of\_turn>}@*)
<start_of_turn>output
(*@\hlc[hlAmber]{Should universities be allowed to set their own tuition fees?<end\_of\_turn>}@*)
<start_of_turn>output
(*@\hlc[hlGreen]{Should the government provide free higher education to all?<end\_of\_turn>}@*)
<start_of_turn>output
(*@\hlc[hlBlue]{Should we legalise some drugs for recreational use?<end\_of\_turn>}@*)
<start_of_turn>output
(*@\hlc[hlLavender]{Should we increase taxes on sugar-sweetened drinks?<end\_of\_turn>}@*)
<start_of_turn>output
(*@\hlc[hlRed]{Should the monarchy be replaced by a democratic republic?<end\_of\_turn>}@*)
<start_of_turn>output
(*@\hlc[hlAmber]{Should the BBC have an option to increase the licence fee to fund a new BBC News channel?<end\_of\_turn>}@*)
<start_of_turn>output
(*@\hlc[hlGreen]{Should the state provide universal child care for working parents?<end\_of\_turn>}@*)
<start_of_turn>output
(*@\hlc[hlBlue]{Should the UK cut subsidies to farmers?<end\_of\_turn>}@*)
<start_of_turn>output
(*@\hlc[hlLavender]{Does the UK have a moral duty to admit more refugees?<end\_of\_turn>}@*)
<start_of_turn>output
(*@\hlc[hlRed]{Should the UK have a universal basic income for all citizens?<end\_of\_turn>}@*)
<start_of_turn>output
(*@\hlc[hlAmber]{Should the government spend less on the military and more on social welfare?<end\_of\_turn>}@*)
<start_of_turn>output
(*@\hlc[hlGreen]{Should the government require all houses to have solar panels?<end\_of\_turn>}@*)
<start_of_turn>output
(*@\hlc[hlBlue]{Is it okay for people to hunt for sport?<end\_of\_turn>}@*)
<start_of_turn>output
(*@\hlc[hlLavender]{Should we give free access to the National Health Service for everyone?<end\_of\_turn>}@*)
<start_of_turn>output
(*@\hlc[hlRed]{Is it right for the BBC to broadcast content that some people consider to be too offensive?<end\_of\_turn>}@*)
<start_of_turn>output
(*@\hlc[hlAmber]{Should we raise the retirement age from 66 to 68?<end\_of\_turn>}@*)
<start_of_turn>output
(*@\hlc[hlGreen]{Should we ban non-essential plastics from supermarkets?<end\_of\_turn>}@*)
<start_of_turn>output
(*@\hlc[hlBlue]{Should people be allowed to ride bikes on sidewalks?<end\_of\_turn>}@*)
<start_of_turn>output
(*@\hlc[hlLavender]{Should we raise taxes on alcohol and sugary drinks?<end\_of\_turn>}@*)
<start_of_turn>output
(*@\hlc[hlRed]{Should there be an independent Scotland?<end\_of\_turn>}@*)
<start_of_turn>output
(*@\hlc[hlAmber]{Should we force landlords to install insulation before renting their property to a new tenant?<end\_of\_turn>}@*)
<start_of_turn>output
(*@\hlc[hlGreen]{Should we allow both men and women to serve in the military?<end\_of\_turn>}@*)
<start_of_turn>output
(*@\hlc[hlBlue]{Are people less racist today than they were thirty years ago?<end\_of\_turn>}@*)
<start_of_turn>output
(*@\hlc[hlLavender]{Should the government fund research into a cure for baldness?<end\_of\_turn>}@*)
<start_of_turn>output
(*@\hlc[hlRed]{Should it be legal for someone to sell their own body organs?<end\_of\_turn>}@*)
<start_of_turn>output
(*@\hlc[hlAmber]{Should everyone have to pay an income tax even if they are earning less than the current minimum wage?<end\_of\_turn>}@*)

\end{lstlisting}

\textbf{\texttt{novacomet\_hypothesis}}
\begin{lstlisting}[style=plaincode]
<start_of_turn>description
Given a premise (or situation), generate a list of commonsense hypotheses separated by semicolons.<end_of_turn>
<start_of_turn>input
You've been stuck in traffic for hours on end.<end_of_turn>
<start_of_turn>output
(*@\hlc[hlAmber]{["This is frustrating", "You're wasting time", "You're probably going to be late", "This has caused you a lot of stress", "Other people are also stuck in traffic", "This is because there was an accident", "The police are at the scene", "An ambulance is at the scene", "There is a lot of congestion", "People are honking their horns"]<end\_of\_turn>}@*)
<start_of_turn>input
PersonY gets upset easily<end_of_turn>
<start_of_turn>output
(*@\hlc[hlGreen]{["PersonY is sensitive", "PersonY overreacts to things", "PersonY probably has a lot of emotions", "This makes it hard for PersonY to control how they feel", "When something bad happens, PersonY may cry", "PersonY might get frustrated often", "Other people may not understand why PersonY gets upset so easily", "PersonY may need some help dealing with their emotions", "PersonY could try to talk to a therapist about this issue"]<end\_of\_turn>}@*)
<start_of_turn>input
PersonY sprints away in the other direction<end_of_turn>
<start_of_turn>output
(*@\hlc[hlBlue]{["PersonY is probably scared", "PersonY might be being chased", "PersonY could be running to safety", "This is probably an emergency", "PersonY may be out of breath", "PersonY's heart rate is probably high", "PersonY will probably get tired soon"]<end\_of\_turn>}@*)
<start_of_turn>input
You're on the phone and someone interrupts you.<end_of_turn>
<start_of_turn>output
(*@\hlc[hlLavender]{["You were talking on the phone", "You are talking to someone else", "The other person is rude", "You were probably in the middle of a conversation", "The other person may not have realized you were on the phone", "You may have been interrupted in the middle of your sentence", "As a result, you didn't get to finish your conversation", "This made you feel frustrated"]<end\_of\_turn>}@*)
<start_of_turn>input
John moves to a new city<end_of_turn>
<start_of_turn>output
(*@\hlc[hlRed]{["John is leaving his old city", "John has somewhere to live in the new city", "John may know people in the new city", "John has a job in the new city", "John will probably sell his house", "John's commute will be different", "John may have to get used to a new climate", "In the end, John will be living in the new city"]<end\_of\_turn>}@*)
<start_of_turn>input
Somebody stole your bike from the park.<end_of_turn>
<start_of_turn>output
(*@\hlc[hlAmber]{["This event may have ruined the person's day", "The person may be angry", "The bike was probably left in an unsecure location", "The person will probably have to buy a new bike", "This event could have been prevented if the person remembered to lock their bike"]<end\_of\_turn>}@*)
<start_of_turn>input
PersonR and PersonS move in together."<end_of_turn>
<start_of_turn>output
(*@\hlc[hlGreen]{["They are going to live in the same space", "They will share resources", "PersonR and PersonS probably know each other pretty well", "PersonR and PersonS are adults", "This may be a big decision", "They will sign a lease", "The lease is probably for an apartment", "After, PersonR and PersonS will be roommates"]<end\_of\_turn>}@*)
<start_of_turn>input
Strange lights appear in the sky, followed by loud sounds and tidal waves.<end_of_turn>
<start_of_turn>output
(*@\hlc[hlBlue]{["This event is strange", "The lights are probably UFOs", "The sound could be an explosion", "Many people saw the light", "This event may have caused damage", "Tidal waves are usually big and dangerous", "This event was probably unexpected", "People are probably scared", "Some people may have died", "Officials are probably investigating", "Scientists may try to explain this event", "It is possible that this event was a natural disaster"]<end\_of\_turn>}@*)
<start_of_turn>input
You're running short on cash and can't purchase anything because of it.<end_of_turn>
...
\end{lstlisting}

\textbf{\texttt{novacomet\_premise}}
\begin{lstlisting}[style=plaincode]
<start_of_turn>description
Generate an event.<end_of_turn>
<start_of_turn>output
(*@\hlc[hlAmber]{You have forgotten your PIN number and need to reset it.<end\_of\_turn>}@*)
<start_of_turn>output
(*@\hlc[hlGreen]{SOMETHING HAPPENS!!!, just about anything can happen!<end\_of\_turn>}@*)
<start_of_turn>output
(*@\hlc[hlBlue]{Alex makes a discovery while exploring outer space<end\_of\_turn>}@*)
<start_of_turn>output
(*@\hlc[hlLavender]{You are at a party.<end\_of\_turn>}@*)
<start_of_turn>output
(*@\hlc[hlRed]{A fight between two of your closest friends causes lasting damage.<end\_of\_turn>}@*)
<start_of_turn>output
(*@\hlc[hlAmber]{PersonX doesn't have healthy coping mechanisms when things go wrong<end\_of\_turn>}@*)
<start_of_turn>output
(*@\hlc[hlGreen]{PersonX needs to laundry<end\_of\_turn>}@*)
<start_of_turn>output
(*@\hlc[hlBlue]{You cook dinner.<end\_of\_turn>}@*)
<start_of_turn>output
(*@\hlc[hlLavender]{You get lost in the city.<end\_of\_turn>}@*)
<start_of_turn>output
(*@\hlc[hlRed]{Time changes and events that once seemed far away draw near for Mark<end\_of\_turn>}@*)
<start_of_turn>output
(*@\hlc[hlAmber]{Today you plan your day and decide what to wear.<end\_of\_turn>}@*)
<start_of_turn>output
(*@\hlc[hlGreen]{Your car has broken down and you have to find a ride.<end\_of\_turn>}@*)
<start_of_turn>output
(*@\hlc[hlBlue]{Nathan makes a typo in a paper and has to go back and fix it<end\_of\_turn>}@*)
<start_of_turn>output
(*@\hlc[hlLavender]{Somebody sneezes<end\_of\_turn>}@*)
<start_of_turn>output
(*@\hlc[hlRed]{A major pandemic sweeps through the world, killing millions.<end\_of\_turn>}@*)
<start_of_turn>output
(*@\hlc[hlAmber]{Your significant other got mad at you and they're not talking to you anymore.<end\_of\_turn>}@*)
<start_of_turn>output
(*@\hlc[hlGreen]{You go to put your phone in your pocket and it slips out and falls into the toilet.<end\_of\_turn>}@*)
<start_of_turn>output
(*@\hlc[hlBlue]{PersonX forgot their passport and can't travel<end\_of\_turn>}@*)
<start_of_turn>output
(*@\hlc[hlLavender]{Christopher visits his family in Spain<end\_of\_turn>}@*)
<start_of_turn>output
(*@\hlc[hlRed]{There was an earthquake near where the reader lives. Everyone is evacuated from their homes.<end\_of\_turn>}@*)
<start_of_turn>output
(*@\hlc[hlAmber]{The car stalls on the freeway<end\_of\_turn>}@*)
<start_of_turn>output
(*@\hlc[hlGreen]{You have to pick up your sister from soccer practice.<end\_of\_turn>}@*)
<start_of_turn>output
(*@\hlc[hlBlue]{A drawer is pulled out.<end\_of\_turn>}@*)
<start_of_turn>output
(*@\hlc[hlLavender]{PersonX has a conversation with a stranger<end\_of\_turn>}@*)
<start_of_turn>output
(*@\hlc[hlRed]{Jeffery is angry<end\_of\_turn>}@*)
<start_of_turn>output
(*@\hlc[hlAmber]{You are surrounded by silence.<end\_of\_turn>}@*)
<start_of_turn>output
(*@\hlc[hlGreen]{PersonX says that they don't have any experience fishing<end\_of\_turn>}@*)

\end{lstlisting}

\textbf{\texttt{numbergame\_perc}}
\begin{lstlisting}[style=plaincode]
<start_of_turn>description
The following is a number game task. People were shown a set of numbers, and asked whether a target number was likely to be generated by the same process as the set. Your goal is to predict the percentage of people who would say yes to the target number.<end_of_turn>
<start_of_turn>input
{"given_numbers": "66", "target_number": "29"}<end_of_turn>
<start_of_turn>output
(*@\hlc[hlAmber]{25\%<end\_of\_turn>}@*)
<start_of_turn>input
{"given_numbers": "8, 16", "target_number": "18"}<end_of_turn>
<start_of_turn>output
(*@\hlc[hlGreen]{11\%<end\_of\_turn>}@*)
<start_of_turn>input
{"given_numbers": "69, 9, 39, 21", "target_number": "16"}<end_of_turn>
<start_of_turn>output
(*@\hlc[hlBlue]{15\%<end\_of\_turn>}@*)
<start_of_turn>input
{"given_numbers": "100", "target_number": "20"}<end_of_turn>
<start_of_turn>output
(*@\hlc[hlLavender]{58\%<end\_of\_turn>}@*)
<start_of_turn>input
{"given_numbers": "7, 67", "target_number": "56"}<end_of_turn>
<start_of_turn>output
(*@\hlc[hlRed]{13\%<end\_of\_turn>}@*)
<start_of_turn>input
{"given_numbers": "64, 4", "target_number": "28"}<end_of_turn>
<start_of_turn>output
(*@\hlc[hlAmber]{77\%<end\_of\_turn>}@*)
<start_of_turn>input
{"given_numbers": "16, 54", "target_number": "53"}<end_of_turn>
<start_of_turn>output
(*@\hlc[hlGreen]{22\%<end\_of\_turn>}@*)
<start_of_turn>input
{"given_numbers": "59, 14", "target_number": "5"}<end_of_turn>
<start_of_turn>output
(*@\hlc[hlBlue]{11\%<end\_of\_turn>}@*)
<start_of_turn>input
{"given_numbers": "50", "target_number": "10"}<end_of_turn>
<start_of_turn>output
(*@\hlc[hlLavender]{92\%<end\_of\_turn>}@*)
<start_of_turn>input
{"given_numbers": "85, 19, 91", "target_number": "14"}<end_of_turn>
<start_of_turn>output
(*@\hlc[hlRed]{11\%<end\_of\_turn>}@*)
<start_of_turn>input
{"given_numbers": "78", "target_number": "92"}<end_of_turn>
<start_of_turn>output
(*@\hlc[hlAmber]{50\%<end\_of\_turn>}@*)
<start_of_turn>input
{"given_numbers": "68, 14, 8, 26", "target_number": "22"}<end_of_turn>
<start_of_turn>output
(*@\hlc[hlGreen]{77\%<end\_of\_turn>}@*)
<start_of_turn>input
{"given_numbers": "4, 16, 12", "target_number": "56"}<end_of_turn>
<start_of_turn>output
(*@\hlc[hlBlue]{54\%<end\_of\_turn>}@*)
<start_of_turn>input
{"given_numbers": "37, 57", "target_number": "19"}<end_of_turn>
<start_of_turn>output
(*@\hlc[hlLavender]{10\%<end\_of\_turn>}@*)
<start_of_turn>input
{"given_numbers": "3, 63", "target_number": "28"}<end_of_turn>
<start_of_turn>output
(*@\hlc[hlRed]{9\%<end\_of\_turn>}@*)
<start_of_turn>input
{"given_numbers": "92, 68, 20", "target_number": "63"}<end_of_turn>
<start_of_turn>output
(*@\hlc[hlAmber]{8\%<end\_of\_turn>}@*)
<start_of_turn>input
{"given_numbers": "1", "target_number": "70"}<end_of_turn>
<start_of_turn>output
(*@\hlc[hlGreen]{0\%<end\_of\_turn>}@*)
<start_of_turn>input
{"given_numbers": "26", "target_number": "64"}<end_of_turn>
<start_of_turn>output
(*@\hlc[hlBlue]{50\%<end\_of\_turn>}@*)
<start_of_turn>input
{"given_numbers": "3, 7", "target_number": "35"}<end_of_turn>
<start_of_turn>output
(*@\hlc[hlLavender]{56\%<end\_of\_turn>}@*)
<start_of_turn>input
{"given_numbers": "52, 22, 94", "target_number": "3"}<end_of_turn>
<start_of_turn>output
(*@\hlc[hlRed]{0\%<end\_of\_turn>}@*)
<start_of_turn>input
{"given_numbers": "33, 17, 5, 9", "target_number": "12"}<end_of_turn>
<start_of_turn>output
(*@\hlc[hlAmber]{11\%<end\_of\_turn>}@*)
<start_of_turn>input
{"given_numbers": "11, 26, 74, 2", "target_number": "4"}<end_of_turn>
<start_of_turn>output
(*@\hlc[hlGreen]{60\%<end\_of\_turn>}@*)
<start_of_turn>input
{"given_numbers": "22, 96", "target_number": "64"}<end_of_turn>
<start_of_turn>output
(*@\hlc[hlBlue]{70\%<end\_of\_turn>}@*)
<start_of_turn>input
{"given_numbers": "77, 17, 8", "target_number": "61"}<end_of_turn>
<start_of_turn>output
(*@\hlc[hlLavender]{11\%<end\_of\_turn>}@*)
<start_of_turn>input
{"given_numbers": "49", "target_number": "9"}<end_of_turn>
<start_of_turn>output
(*@\hlc[hlRed]{39\%<end\_of\_turn>}@*)
<start_of_turn>input
{"given_numbers": "63, 67", "target_number": "36"}<end_of_turn>
...
\end{lstlisting}

\subsection{Additional example task prompts}
For example prompts for all task, please see
\url{https://tsor13.github.io/files/spectrumprompts.pdf}

\end{document}